\documentclass[acmtog]{acmart}

\copyrightyear{2024}
\acmYear{2024}
\setcopyright{rightsretained}
\acmConference[SA Conference Papers '24]{SIGGRAPH Asia 2024 Conference Papers}{December 3--6, 2024}{Tokyo, Japan}
\acmBooktitle{SIGGRAPH Asia 2024 Conference Papers (SA Conference Papers '24), December 3--6, 2024, Tokyo, Japan}
\acmDOI{10.1145/3680528.3687590}
\acmISBN{979-8-4007-1131-2/24/12}

\acmSubmissionID{308}

\usepackage{cleveref}
\usepackage{algorithmic}
\usepackage[export]{adjustbox}
\usepackage{pgfplots}
\usepackage[normalem]{ulem}
\usepackage{colortbl}

\newcommand{\ignorethis}[1]{}

\newcommand{\eg         }     {{e.g.}}

\newcommand{\ie         }     {{i.e.}}

\newcommand{\Reals      }     {{\textrm{I\kern-0.18em R}}}

\newcommand{\change     } [1] {\mbox{{\footnotesize $\Delta$} \kern-3pt}#1}

\definecolor{darkred}{rgb}{0.7,0.1,0.1}
\definecolor{darkgreen}{rgb}{0.1,0.6,0.1}
\definecolor{cyan}{rgb}{0.7,0.0,0.7}
\definecolor{otherblue}{rgb}{0.1,0.4,0.8}
\definecolor{maroon}{rgb}{0.76,.13,.28}
\definecolor{burntorange}{rgb}{0.81,.33,0}

\newif\ifdraft
\drafttrue

\ifdraft
  \newcommand{\todo}[1]{{\color{cyan}[\textbf{TODO:} #1]}}
  \newcommand{\omri}[1]{{\color{burntorange}[\textbf{Omri:} #1]}}
  \newcommand{\weili}[1]{{\color{darkgreen}[\textbf{Weili:} #1]}}
  \newcommand{\arash}[1]{{\color{maroon}[\textbf{Arash:} #1]}}
  \newcommand{\rinon}[1]{{\color{darkred}[\textbf{Rinon:} #1]}}
  \newcommand{\gal}[1]{{\color{red}[\textbf{Gal:} #1]}}
  \newcommand{\ohad}[1]{{\color{magenta}[\textbf{Ohad:} #1]}}
  \newcommand{\danix}[1]{{\color{blue}[\textbf{Danix:} #1]}}

\else
  \newcommand{\omri}[1]{}
  \newcommand{\weili}[1]{}
  \newcommand{\arash}[1]{}
  \newcommand{\rinon}[1]{}
  \newcommand{\gal}[1]{}
  \newcommand{\ohad}[1]{}
  \newcommand{\danix}[1]{}

  \newcommand {\note}[1]{}
  \newcommand {\todo}[1]{}
\fi

\definecolor{blobacolor}{RGB}{242,78,112}
\newcommand {\bloba}[1]{{\color{blobacolor}{\textit{``#1''}}}}
\definecolor{blobbcolor}{RGB}{55, 146, 55}
\newcommand {\blobb}[1]{{\color{blobbcolor}\textit{``#1''}}}
\definecolor{blobccolor}{RGB}{41,160,177}

\definecolor{blobdcolor}{RGB}{255,176,1}

\newcommand\todosilent[1]{}

\providecommand{\keywords}[1]
{
  \textbf{\textit{Keywords---}} #1
}

\newlength{\ww}

\definecolor{colora}{RGB}{242,78,112}

\definecolor{colorb}{RGB}{255,176,1}

\definecolor{colorc}{RGB}{41,160,177}

\definecolor{colord}{RGB}{55, 146, 55}

\newcommand\blfootnote[1]{%
  \begingroup
  \renewcommand\thefootnote{}\footnote{#1}%
  \addtocounter{footnote}{-1}%
  \endgroup
}

\begin{document}

\title{DiffUHaul: A Training-Free Method for Object Dragging in Images}

\makeatletter
\let\@authorsaddresses\@empty
\makeatother

\author{Omri Avrahami}
\orcid{0000-0002-7628-7525}
\affiliation{
 \institution{The Hebrew University of Jerusalem}
 \city{Jerusalem}
 \country{Israel}
}
\affiliation{
 \institution{NVIDIA Research}
 \city{Santa Clara}
 \country{United States of America}
}

\author{Rinon Gal}
\orcid{0000-0003-4875-965X}
\affiliation{
 \institution{NVIDIA Research}
 \city{Tel Aviv}
 \country{Israel}
}

\author{Gal Chechik}
\orcid{0000-0001-9164-5303}
\affiliation{
 \institution{NVIDIA Research}
 \city{Tel Aviv}
 \country{Israel}
}

\author{Ohad Fried}
\orcid{0000-0001-7109-4006}
\affiliation{
 \institution{Reichman University}
 \city{Herzliya}
 \country{Israel}
}

\author{Dani Lischinski}
\orcid{0000-0002-6191-0361}
\affiliation{
 \institution{The Hebrew University of Jerusalem}
 \city{Jerusalem}
 \country{Israel}
}

\author{Arash Vahdat}
\orcid{0009-0005-9476-1306}
\affiliation{
 \institution{NVIDIA Research}
 \city{Santa Clara}
 \country{United States of America}
}
\authornote{Indicates Equal Advising}

\author{Weili Nie}
\orcid{0000-0002-0030-3189}
\affiliation{
 \institution{NVIDIA Research}
 \city{Santa Clara}
 \country{United States of America}
}
\authornotemark[1]

\renewcommand\shortauthors{Avrahami et al.}

\begin{abstract}
    Text-to-image diffusion models have proven effective for solving many image editing tasks.
    However, the seemingly straightforward task of seamlessly relocating objects within a scene remains surprisingly challenging. Existing methods addressing this problem often struggle to function reliably in real-world scenarios due to lacking spatial reasoning. 
    In this work, we propose a training-free method, dubbed \emph{DiffUHaul}, that harnesses the spatial understanding of a \emph{localized} text-to-image model, for the object dragging task.
    Blindly manipulating layout inputs of the localized model tends to cause low editing performance due to the intrinsic entanglement of object representation in the model. To this end, we first apply attention masking in each denoising step to make the generation more disentangled across different objects and adopt the self-attention sharing mechanism to preserve the high-level object appearance. Furthermore, we propose a new diffusion anchoring technique: in the early denoising steps, we interpolate the attention features between source and target images to smoothly fuse new layouts with the original appearance; in the later denoising steps, we pass the localized features from the source images to the interpolated images to retain fine-grained object details. To adapt DiffUHaul to real-image editing, we apply a DDPM self-attention bucketing that can better reconstruct real images with the localized model.
    Finally, we introduce an automated evaluation pipeline for this task and  showcase the efficacy of our method. Our results are reinforced through a user preference study.
\end{abstract}




\begin{teaserfigure}
    \centering
    \includegraphics[width=\linewidth]{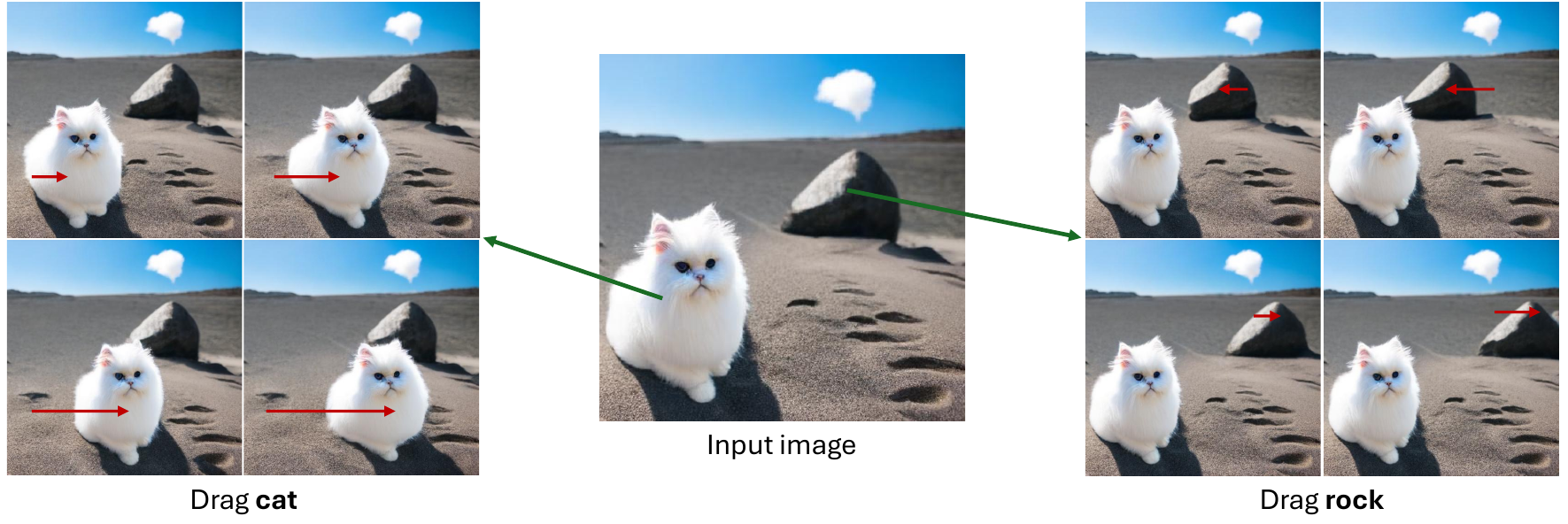} 
    \caption{\textbf{DiffUHaul:} Given a real image with multiple objects (e.g., a cat and a rock), our method is able to seamlessly drag each of the objects to an arbitrary location within the image while preserving the foreground and background appearance.}
    \label{fig:teaser}
\end{teaserfigure}

\maketitle

\blfootnote{Project page is available at: \textcolor{red}{\href{https://omriavrahami.com/diffuhaul/}{https://omriavrahami.com/diffuhaul/}}}

\section{Introduction}
\label{sec:introduction}

Think about a digital artist who recently employed an advanced generative model to craft an image featuring a Persian cat alongside a rock, as in \Cref{fig:teaser}. All that is needed for their creation to achieve perfection is for the cat (or the rock) to be moved slightly. Despite the conceptual simplicity of such a task, seamlessly dragging objects in an image is surprisingly challenging for current generative image editing methods \cite{brooks2022instructpix2pix,hertz2022prompt}. In this work, we propose a novel training-free solution for this scenario.

\begin{figure}[tp]
    \centering
    \setlength{\tabcolsep}{0.6pt}
    \renewcommand{\arraystretch}{1}
    \setlength{\ww}{0.235\columnwidth}
    \begin{tabular}{cc @{\hspace{7\tabcolsep}} ccc}
        \rotatebox[origin=c]{90}{\footnotesize{DiffEdit}} &
        {\includegraphics[valign=c, width=\ww]{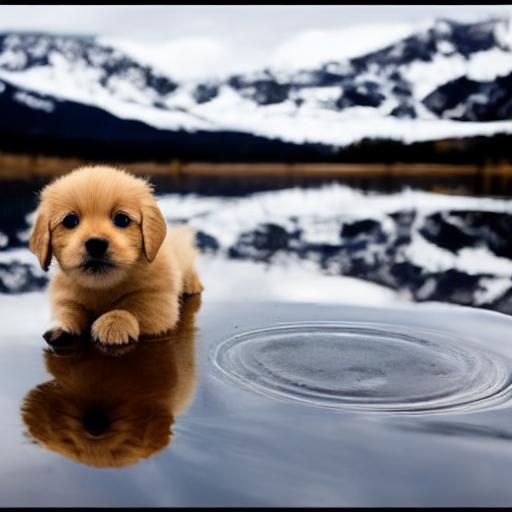}} &
        {\includegraphics[valign=c, width=\ww]{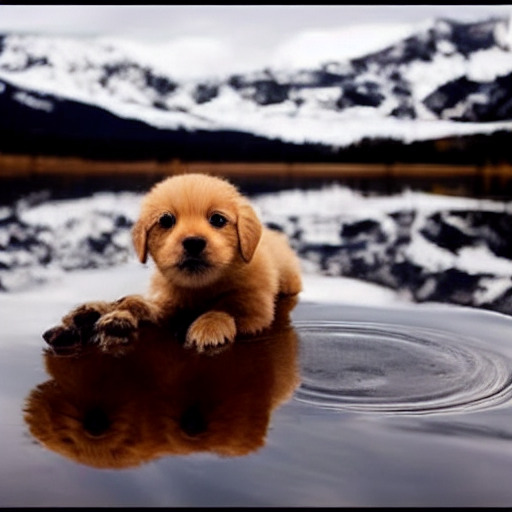}} &
        {\includegraphics[valign=c, width=\ww]{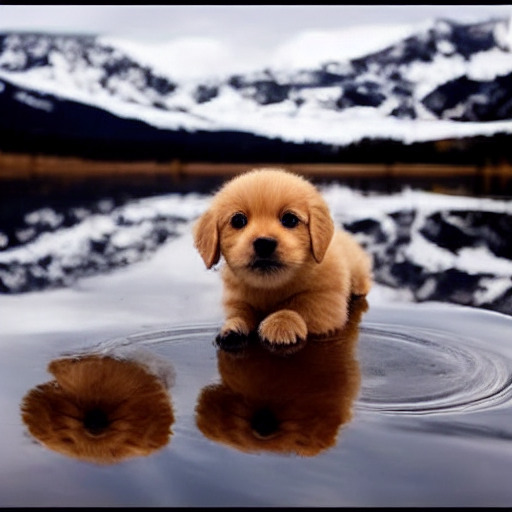}} &
        {\includegraphics[valign=c, width=\ww]{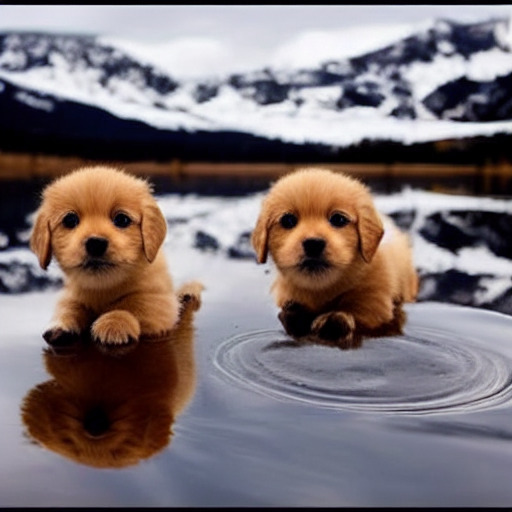}}
        \vspace{5px}
        \\

        \rotatebox[origin=c]{90}{\footnotesize{DiffUHaul (ours)}} &
        {\includegraphics[valign=c, width=\ww]{figures/motivation/assets/dog/ours/res0.jpg}} &
        {\includegraphics[valign=c, width=\ww]{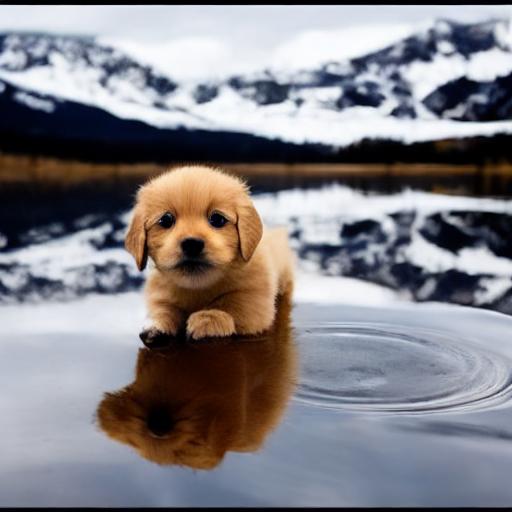}} &
        {\includegraphics[valign=c, width=\ww]{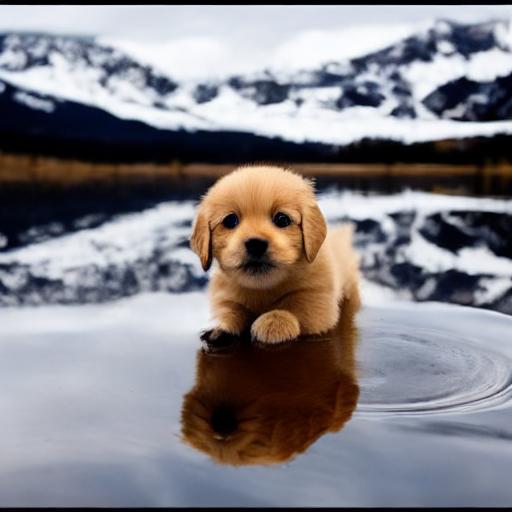}} &
        {\includegraphics[valign=c, width=\ww]{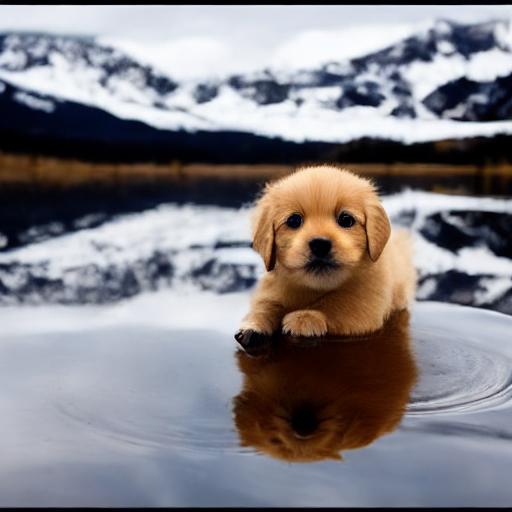}}
        \vspace{0.1px}
        \\

        &
        \footnotesize{Input} &
        \multicolumn{3}{c}{\footnotesize{Dragging the puppy from left to right ($\rightarrow$)}}
        \\

    \end{tabular}
    \caption{\textbf{Object Dragging Robustness.} When dragging a puppy in a complex environment (particularly with its reflection in the water and ripples nearby) to different locations along from left to right, previous method DiffEdit~\citep{Mou2024DiffEditorBA} struggles with the editing traces left in its original location, while our method demonstrates a more robust behavior.}
    \label{fig:motivation}
\end{figure}

Current methods that tackle this problem rely on time-consuming LoRA training per image \cite{Shi2023DragDiffusionHD}, training a designated model on a large dataset \cite{Chen2023AnyDoorZO,Yang2022PaintBE} or utilizing classifier-free guidance (CFG) with specific objectives \cite{Mou2023DragonDiffusionED,Mou2024DiffEditorBA,self_guidance}. However, these methods are not robust and struggle to operate reliably in a real-world setting. For example, as can be seen in \Cref{fig:motivation}, DiffEdit \cite{Mou2024DiffEditorBA} suffers from artifacts of traces of the puppy in its original location, while our method demonstrates a more robust behavior.

Recently, several \emph{localized} text-to-image models were developed by the community that add spatial controllability to the task of text-to-image generation~\cite{Li2023GLIGENOG,spatext_2023_CVPR,yang2023reco, zheng2023layoutdiffusion, zhang2023controlnet, nie2024compositional}. A natural question is then whether the localized understanding of the 2D pixel world in such models can be harnessed for the task of object dragging. Hence, we examine the disentanglement properties of such models, and propose a series of modifications that allow them to serve as a backbone for drag-and-drop movement of objects within an image. Specifically, we use the recently introduced BlobGEN~\cite{nie2024compositional} model, and demonstrate that its spatial understanding can enable significantly more robust object dragging without requiring fine-tuning or training. 

In pursuit of our solution, we begin by revealing an entanglement problem in the localized text-to-image models, through which the prompt-based localized controls of different image regions interfere with each other. We trace the root cause to the commonly used Gated Self-Attention layers~\citep{Li2023GLIGENOG}, 
where each individual layout embedding are free to attend to all the visual features. 
We propose an inference-time masking-based solution, named \textit{gated self-attention masking}, and show that improving the model disentanglement leads to better object dragging performance.

Next, specially for the object dragging task, we first adopt the commonly-used self-attention sharing mechanism~\citep{cao2023masactrl} to preserve the high-level object appearance. To better transfer the fine-grained object details from source images to target images and better harness spatial understanding of the model, we propose a novel soft anchoring mechanism:
in early denoising steps, which control the object shape and scene layout in an image, we interpolate the self-attention features of the source image and those of the target image with a coefficient relative to the diffusion time step. This process promotes a smooth fusion between the target layout and source appearance. Then, in later denoising steps, which control the fine-grained visual appearance in an image, we update the interpolated attention features from the corresponding features in the source image via the nearest-neighbor copying.

To adapt our method to real-image editing, we further require an inversion solution that is compatible with the localized method. We find that the standard DDIM inversion \cite{song2020denoising} struggles to reconstruct the image faithfully, even when not using classifier-free guidance \cite{Ho2022ClassifierFreeDG}. Hence, we propose a simple DDPM self-attention bucketing technique that adds noise to the reference image \emph{independently} in each diffusion step, and uses the noisy images to extract the self-attention outputs as the source attention features. This DDPM bucketing does not accumulate reconstruction errors along the denoising process and preserves details for real images.

Finally, we offer automatic metrics for our problem to assess different aspects of the editing operations, and use them for an extensive comparison that demonstrates the effectiveness of our method over the baselines. In addition, we conduct a user study and show that our method is also preferred by human evaluators.

In summary, our contributions are: (1) we show that the spatial understanding of a localized text-to-image model can be effectively harnessed to tackle the object dragging task, (2) we reveal an entanglement problem in the gated self-attention layers and offer an inference-time solution, (3) we introduce a novel soft anchoring mechanism that fuses the source object appearances and the target scene layouts during the denoising process, (4) we show that DDPM self-attention bucketing suffices for real image editing, and finally (5) we develop automatic metrics to the task of object dragging and use them to evaluate our method quantitatively, in addition to a user study, to demonstrate its effectiveness.

\section{Related Work}
\label{sec:related_work}

\paragraph{\textbf{Localized text-to-image models.}} Recently, text-to-image diffusion models~\cite{sohl2015deep,song2019generative,ho2020denoising,song2020denoising, ramesh2022hierarchical,Rombach2021HighResolutionIS, yu2022scaling} became a foundational tool for creative tasks \cite{Avrahami2023TheCO, richardson2023conceptlab, Molad2023DreamixVD, Frenkel2024ImplicitSS}. To add spatial control to existing text-to-image models, some works suggested training a designated localization component to take in visual layouts~\cite{Li2023GLIGENOG,spatext_2023_CVPR,yang2023reco, zheng2023layoutdiffusion, zhang2023controlnet, nie2024compositional} while others offer training-free methods that incorporate the spatial conditioning into the diffusion sampling process~\cite{feng2022training, Chefer2023AttendandExciteAS, chen2023trainingfree, phung2023grounded,BarTal2023MultiDiffusionFD}. In this work, we utilize BlobGEN \cite{nie2024compositional} as our base model since it has shown better spatial understanding and generation quality.

\paragraph{\textbf{Text-to-image editing.}} Soon after the emergence of text-to-image diffusion models, a plethora of methods were offered for various image editing tasks~\cite{meng2021sdedit, blended_2022_CVPR, avrahami2023blendedlatent, mokady2022null, pnpDiffusion2022, hertz2023delta, Kawar2022ImagicTR, cao2023masactrl, Patashnik2023LocalizingOS, Sheynin2023EmuEP}. However, most of these editing methods are spatially preserving (i.e., changing the object attributes and categories), and suffer from the editing tasks requiring spatial reasoning, such as object dragging~\cite{brooks2022instructpix2pix,hertz2022prompt}. 
Localized text-to-image models, such as GLIGEN~\citep{Li2023GLIGENOG} and BlobGEN~\citep{nie2024compositional}, has the potential to solve the object dragging task, but their performance is far from satisfactory without specialized designs. Concurrently, Diffusion Handles~\cite{pandey2023diffusion} offers 3D
object edits using a depth-to-image diffusion model and by performing manipulations on the diffusion activations in 3D. In addition, Magic Fixup~\cite{alzayer2024magic} offers a model that given a coarsely edited image, synthesizes a photorealistic version of it, by leveraging a video dataset, this way they manage to offer a way to edit an image coarsely, and then harmonize the result.

\paragraph{\textbf{Keypoint dragging.}} A similar task is keypoint dragging, where users provide source and target keypoints in the image, and move the source keypoints to the target ones. For example, UserControllableLT~\cite{Endo2022UserControllableLT}, GANWarping~\cite{Wang2022RewritingGR} and DragGAN~\cite{Pan2023DragYG} employ StyleGAN~\cite{karras2019style, karras2020analyzing, karras2021aliasfree} for editing generated images. But they work only on the narrow domain the GAN~\cite{goodfellow2014generative} was trained on (\eg, human faces, churches~\cite{Yu2015LSUNCO}). 
DragDiffusion~\cite{Shi2023DragDiffusionHD} propose a LoRA-based~\cite{lora} method that finetunes a diffusion model given a test image and optimizes the latent noises at inference time. In contrast, our method is training-free. Concurrently, EasyDrag~\cite{hou2024easydrag} improves DragDiffusion~\cite{Shi2023DragDiffusionHD} by replacing the LoRA training with reference guidance.

\paragraph{\textbf{Object dragging.}} Different from keypoints dragging that warps the image to match the target keypoints, object dragging moves the entire object seamlessly to a new position. Object dragging was initially introduced by ~\cite{Epstein2022BlobGANSD, Wang2021ImprovingGE} for single-domain images generated by GANs. Diffusion self-guidance~\cite{self_guidance} proposed to use the guidance from internal representations of a diffusion model for various editing tasks, including object dragging. DragonDiffusion~\cite{Mou2023DragonDiffusionED} and DiffEditor~\cite{Mou2024DiffEditorBA} developed a new classifier guidance~\cite{Dhariwal2021DiffusionMB} specifically designed for object dragging. Most of them use a general diffusion model as the base model, but our method harnesses the spatial understanding of a \emph{localized} diffusion model to better tackle the object dragging task.

\paragraph{\textbf{Object insertion.}}
Many works use multiple images \cite{Ruiz2022DreamBoothFT,Gal2022AnII,arar2024Palp,Alaluf2023ANS,Voynov2023PET} or a single image \cite{Avrahami2023BreakASceneEM, gal2023encoder, arar2023domain} of the same object for image personalization. They are also effective in tackling the task of referenced-based object insertion, in which a reference object is being inserted to a target image. AnyDoor~\cite{Chen2023AnyDoorZO} and PaintByExample~\cite{Yang2022PaintBE} train a designated encoder for this task, which can be used for object dragging by utilizing an inpainting method, as explained in \Cref{sec:experiments}. The concurrent work ObjectDrop~\cite{Winter2024ObjectDropBC} collected a high-quality tailored dataset to train a model for object removal, insertion, and dragging. Our method, however, is training-free with a pre-trained \emph{localized} diffusion model.

\section{Preliminaries}
\label{sec:preliminaries}

\begin{figure}[t]
    \centering
    \includegraphics[width=1\linewidth]{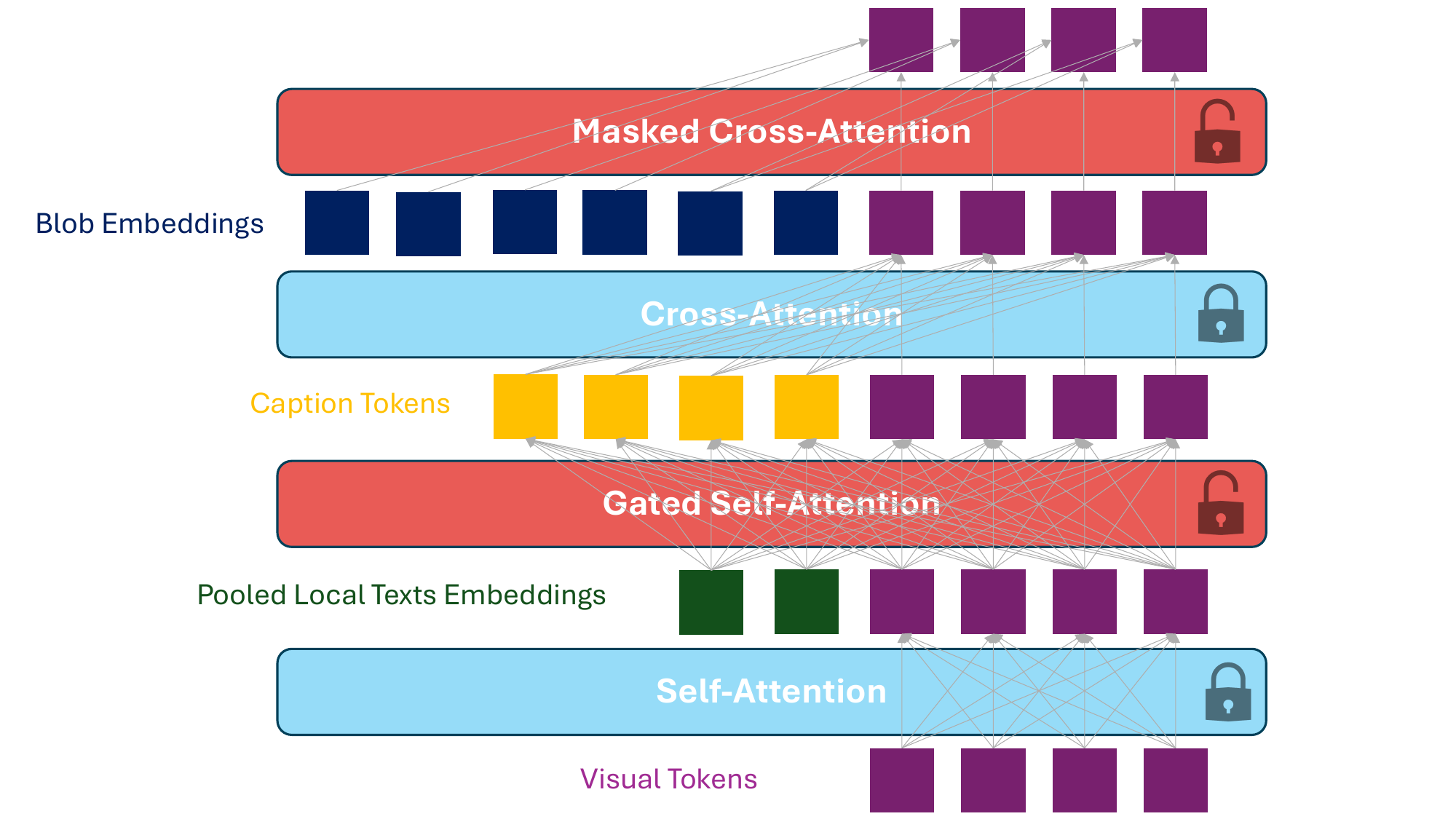}
    \caption{\textbf{BlobGEN Architecture.} BlobGEN incorporates the additional blob information into the Stable Diffusion model by adding two new layers in each attention block: masked cross-attention and gated cross-attention.}
    \label{fig:blobgen_architecture}
\end{figure}

Existing large text-to-image diffusion models suffer from the prompt-following issue, making it challenging to control the visual layouts of their generation via complex prompts only. Thus, incorporating the visual layout information into these large text-to-image diffusion models can enable better object-level controllability~\cite{Li2023GLIGENOG,yang2023reco,Avrahami2023BreakASceneEM}. Among them, visual layouts are usually represented by bounding boxes (along with object categories). %

More recently, BlobGEN~\cite{nie2024compositional} has introduced a new type of visual layouts called blob representations to guide the image synthesis, which shows more fine-grained controllability than all previous approaches. Specifically, the blob representations denote the object-level visual primitives in a scene, each of which consists of two components: blob parameters $\tau$ and blob description $S$. A blob parameter depicts a tilted ellipse using a vector of five variables $\tau = [c_x, c_y, a, b, \theta]$ to specify the object’s position, size and orientation, where ($c_x$, $c_y$) is the center point of the ellipse, $a$ and $b$ are the radii of its semi-major and semi-minor axes, and $\theta \in (-\pi, \pi]$ is the orientation angle of the ellipse.
A blob description $S$ captures the object's visual appearance using a region-level synthetic caption extracted by an image captioning model. Compared with bounding boxes and object categories, the blob representations can retain more detailed spatial and appearance information about the objects in a complex scene.
 
To incorporate blob representations into the existing Stable Diffusion model, BlobGEN adopts a similar architecture design idea to GLIGEN~\citep{Li2023GLIGENOG} that introduces new attention layers in a gated way. To retain the prior knowledge of pre-trained models for synthesizing high-quality images, it freezes the weights of the pre-trained diffusion model and only trains the newly added layers. 
As demonstrated in \Cref{fig:blobgen_architecture}, BlobGEN keeps the gated self-attention module originally developed by GLIGEN while also introducing a new masked cross-attention module in each attention block. 
These two new layers fuse blob inputs into the model differently: In the gated self-attention layer, the blob embeddings are first passed to a pooling layer and then concatenated with the visual features, while, in the masked cross-attention layer, each blob embedding only attends to visual features in its local region as the feature maps are masked by the (rescaled) blob ellipses. 

With this masking design, each blob representation and its local visual feature are trained to align with each other, and thus the model becomes more modular and disentangled. BlobGEN has demonstrated more fine-grained control over its generation. Therefore, we use BlobGEN as our network backbone for solving the object dragging task.

\section{Method}
\label{sec:method}

\begin{figure*}[t]
    \centering
    \includegraphics[width=1\linewidth]{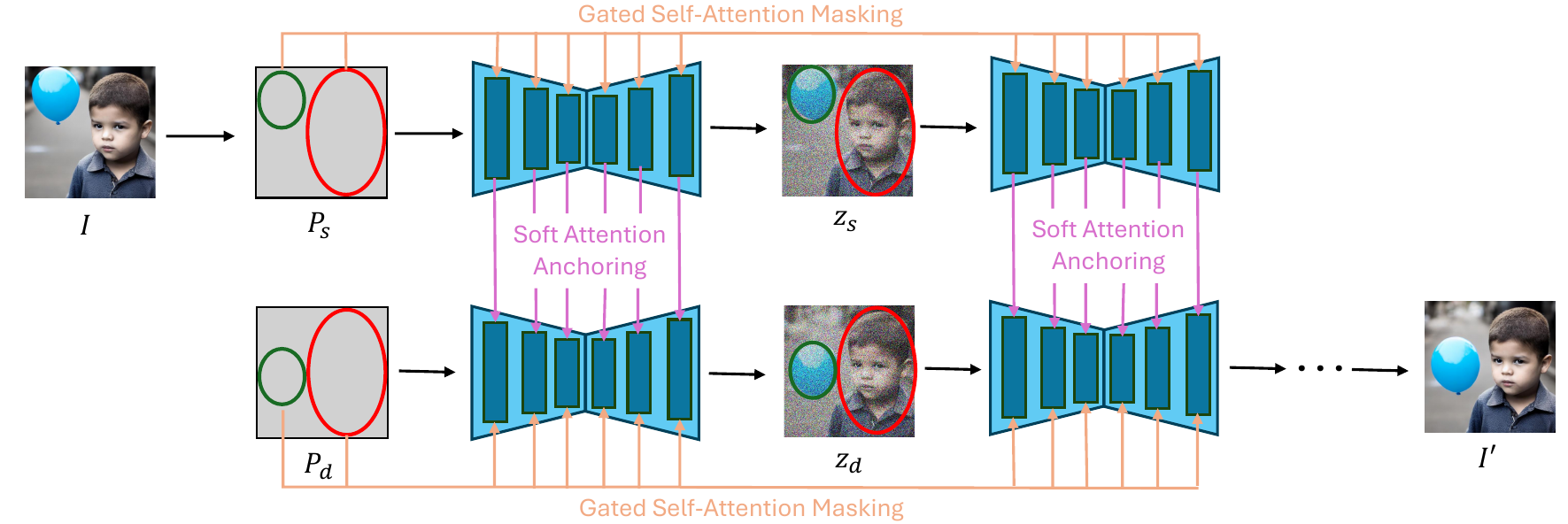}
    \caption{\textbf{Method Overview.} Given an input image $I$, we start by extracting the blob parameters $P_s$ of its layout; then, by changing its layout based on the user provided target location, we get the new blob parameters $P_d$. By conditioning the localized text-to-image model on the respective blob representations, we iteratively denoise the source and target images ($z_s$ and $z_d$) while incorporating gated self-attention masking (\Cref{sec:blobgen_entanglement}) and soft attention anchoring (\Cref{sec:object_moving_generated}) in each self-attention block until we get the desired editing result $I'$.}
    \label{fig:method}
\end{figure*}

Our goal is to offer a solution to the problem of object dragging. To this end, we propose to leverage the spatial knowledge of blob-based text-to-image model BlobGEN~\cite{nie2024compositional}. In \Cref{sec:blobgen_entanglement} we start by investigating the disentanglement offered by this model. We discover significant lingering entanglement, and trace it to the gated-self attention of GLIGEN-style models. Hence, we offer an inference-time mask-based solution to this problem. In \Cref{sec:object_moving_generated} we present our solution for object dragging in generated images: (1) we first utilize self-attention sharing~\cite{cao2023masactrl, Wu2022TuneAVideoOT, geyer2023tokenflow, Tewel2024TrainingFreeCT} to increase the consistency of the dragged object, and (2) we propose a soft anchoring technique to improve the consistency of results. Finally, in \Cref{sec:object_moving_real} we extend our solution to real images by relying on the proposed DDPM self-attention bucketing instead of standard DDIM inversion. Our method is summarized in \Cref{fig:method}.

Formally, given an input image $I$ with an object located in $(c_x, c_y)$ that the user wants to drag,
and a desired target location $(c'_x, c'_y)$, the task of object dragging aims at moving the object to the target location while the rest of the image is left intact, up to desired environment changes (\eg, reflections) in the edited image $I'$.

\subsection{Gated Self-Attention Entanglement}
\label{sec:blobgen_entanglement}

\begin{figure}[tp]
    \centering
    \includegraphics[width=1\linewidth]{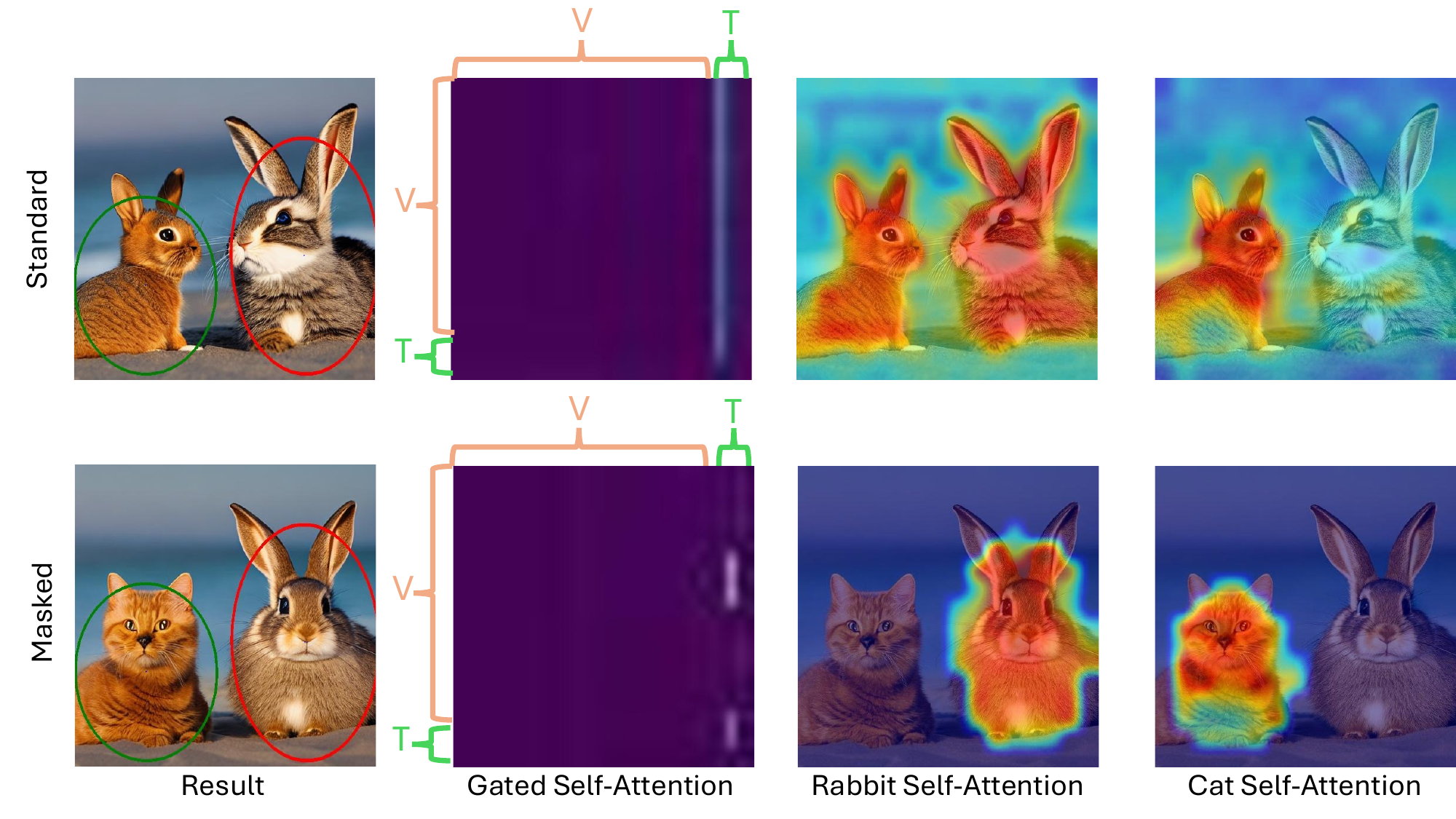}

    \caption{\textbf{Gated Self-Attention Leakage.} Given scene descriptions of two blobs: \bloba{a photo of a rabbit} and \blobb{a photo of a cat}, we can see that the standard BlobGEN model (the first column in the first row) generates two rabbits instead of a cat and a rabbit, we then visualize the gated self-attention layers, as explained in \Cref{sec:blobgen_entanglement}. As can be seen, the standard BlobGEN model (first row) leaks the rabbit information also to the cat blob (the first row third column), while our masked version of the gated self-attention (second row) is able to disentangle the blobs (the second row third column). In addition, we can see that the gated self-attention (second column) behaves de facto as a cross-attention layer, as the vast majority of the attention is between the text tokens $T$ and the visual tokens $V$.}
    \label{fig:gated_self_attention_leakage}
\end{figure}

As explained in \Cref{sec:preliminaries}, BlobGEN was trained to take a set of input blobs ${B_1, ... B_n}$ with corresponding text descriptions ${S_1, ... S_n}$ and blob parameters ${\tau_1, ... \tau_n}$, and generate a scene. This scene is expected to be created in a \emph{disentangled} manner, \ie, the text description $S_i$ should correspond only to the local region depicted by $\tau_i$. To this end, the authors introduced a masked cross-attention layer. However, a simple investigation reveals that the generated result is not fully disentangled in practice. For example, as can be seen in \Cref{fig:gated_self_attention_leakage} (first row), the rabbit text description from one blob spills over to the spatial region of the cat blob.

We hypothesize that the gated self-attention modules that BlobGEN derives from GLIGEN is the root cause of entanglement.
In gated self-attention, a projection layer first converts the CLIP \cite{Radford2021LearningTV} text embeddings of the text description $S_i$ to the text tokens $T = \{t_1, ... t_n\}$. They are then merged with the visual tokens $V = \{v_1, ... v_k\}$
into a unified set $V \cup T = \{v_1, ... v_k, t_1, ... t_n\}$, which altogether are used to calculate the self-attention features, using the standard self-attention mechanism (plus a gated skip connection).

This design choice adds no constraint over the attended areas, \ie, the projected text tokens $T$ can attend to themselves and all the visual tokens $V$. To visualize this phenomenon, we average the gated self-attention maps over the diffusion process. An example is shown in \Cref{fig:gated_self_attention_leakage} (the second column of the first row). This visualization reveals an interesting aspect: the vast majority of attention weights is between the projected text tokens $T$ and the visual tokens $V$, and not within these sets themselves. It means that the gated self-attention layer behaves as a de facto cross-attention layer. 

We examine the attention between the projected text token $t_i$ and all the visual tokens $V$. This is a $K$-dimensional vector, which we first reshape into two dimensions $\sqrt{K}\times \sqrt{K}$, and then resize to a canonical size. We term these maps ``reshaped self-attention'', which are averaged over all the denoising steps. This visualization, as shown in \Cref{fig:gated_self_attention_leakage} (last two columns of the first row), reveals that text tokens indeed attend to undesired areas: the ``rabbit'' text token 
attends to the visual features in both the ``rabbit'' and ``cat'' blob regions, leading to an \emph{entangled} generation. For more details about the visualizations, please refer to the supplementary material.

To this end, we suggest an inference-time solution to the entanglement problem: given $n$ different input blobs with the corresponding parameters ${\tau_1, ... \tau_n}$ we first convert them into $n$ masks ${M_1, ... M_n}$ of $512 \times 512$ resolution. Then, during the diffusion process, for each self-attention layer and for each projected text token $t_i$, we reshape the mask $M_i$ to the corresponding spatial size of the layer, and use it to mask the area of the gated self-attention between the projected text token $t_i$ and the visual tokens $V$. 
This way, we can prevent the token $t_i$ from attending to undesired areas at the inference time.

\subsection{Consistent Object Dragging for Generated Images}
\label{sec:object_moving_generated}

\begin{figure}[t]
    \centering
    \includegraphics[width=1\linewidth]{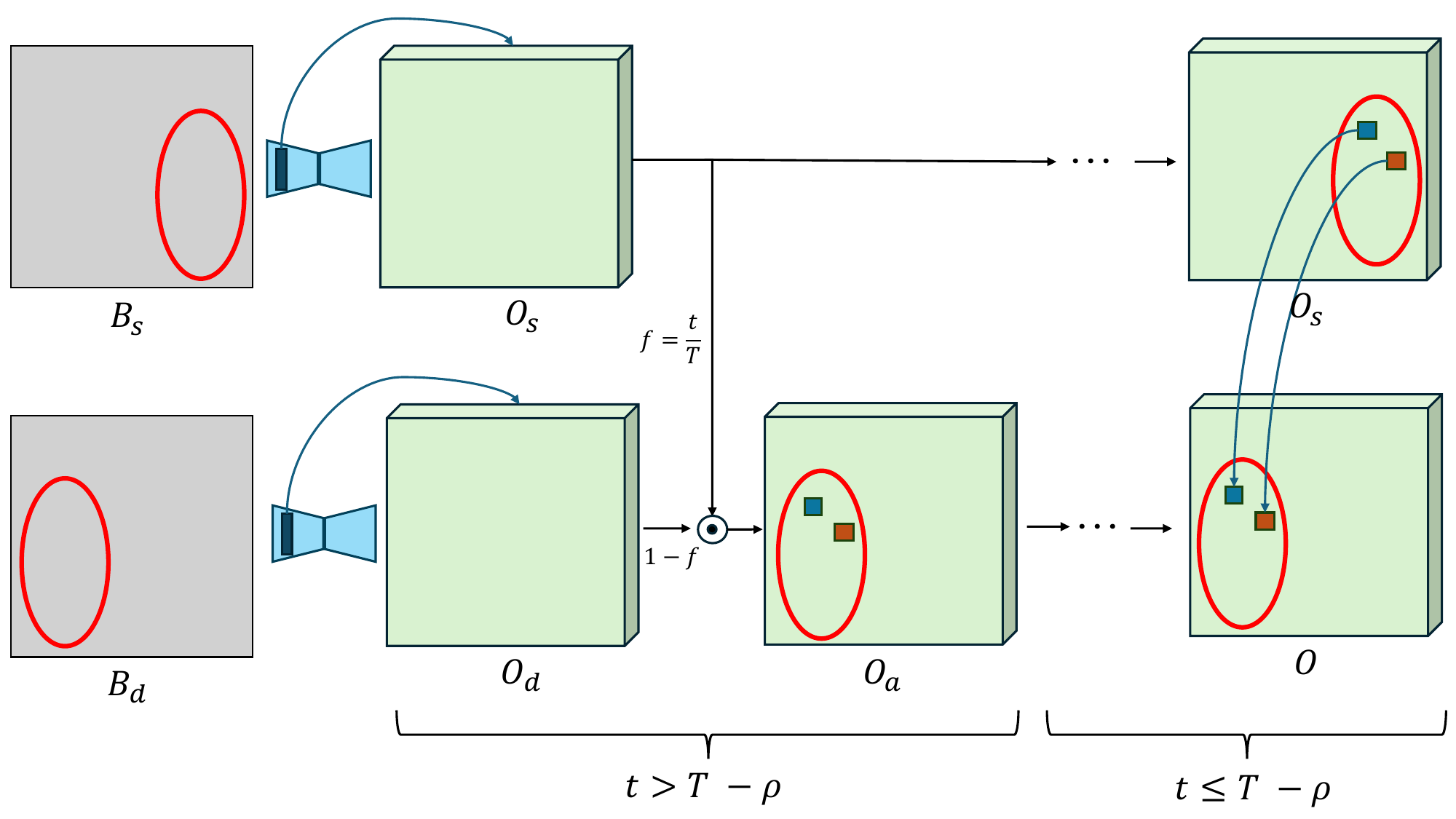}
    \caption{\textbf{Self-Attention Soft Anchoring.} Given the source blob $B_s$ and target blob $B_d$, we start by extracting the self-attention outputs $O_s$ and $O_d$ correspondingly, then, during the first $\rho$ iterations, we blend these maps according to the timestep ratio $f = \frac{t}{T}$ where $t$ is the current timestep and $T$ is the total number of timesteps. Then, after the anchor map $O_a$ is calculated, we use it for determining the position of the new blob, while taking the appearance from the corresponding $O_s$ map using nearest-neighbor copying.}
    \label{fig:self_attention_anchoring}
\end{figure}

Now, we first focus on tackling the object dragging problem for generated images from the localized model: given a scene represented by $n$ blob inputs ${B_1, ... B_n}$,
we change the parameters $\tau_s$ of one blob $B_s$ to $\tau_d$ with a different spatial location such that the $s^{\text{th}}$ object in the generated image will be relocated to the designated location, without changing the appearance of all other objects and the background (barring direct interactions with the object, e.g., shadows).

To preserve the high-level object appearance, we adopt the self-attention sharing mechanism \cite{cao2023masactrl, Wu2022TuneAVideoOT}: we iteratively generate the source image using the source parameters $\tau_s$ in parallel to the target image with the $\tau_d$ parameters. Then, we replace the self-attention keys $K_d$ and values $V_d$ from the target image in each self-attention layer and each denoising step by the keys and values $K_s, V_s$ from the source image.

However, this mechanism alone does not fully preserve the fine-grained details of the source image, so we propose adding a novel \emph{soft anchoring mechanism}: the motivation is that the generated source image already contains the information needed for generating the target image, we can take advantage of the self-attention layers \emph{output} (\ie,~attention features) in the local region that corresponds to the source blob. The soft anchoring is designed to fuse the object appearance information represented by the attention features within the source blob and the positional information indicated by the target blob. Specifically, in the first $\rho$ steps of the denoising process, we perform an adaptive, \emph{soft blending} of the attention features of the generated target image with the features of the source image. The interpolation coefficient is time-dependent: 
we take more visual appearance from the source image in the beginning but more spatial information from the target image in the later steps,
as depicted in \Cref{fig:self_attention_anchoring}. Formally, for each denoising step $t \in [T, T-1, ..., T - \rho + 1]$ and for each self-attention layer, the interpolated self-attention output of the target image is:
$$O_a = O_s * f + O_d * (1 - f); f = \frac{t}{T}$$
where $O_s$ is the self-attention output of the generated source image, $O_d$ is the self-attention output of the generated target image, and $T$ is the total number of denoising steps.
The length of soft blending is controlled by the hyperparameter $\rho$.

Next, during the last $T - \rho$ steps of the denoising process, we use the soft blending result $O_a$ as anchor points for the target object. In each denoising step $t\in[T-\rho, ..., 2, 1]$ and each self-attention layer, we perform the \emph{nearest-neighbor copying}: each entry from the anchor attention features $O_a$ within the target blob $B_d$ is replaced by its nearest-neighbor entry from the source attention features $O_s$ within the source blob $B_s$.
The nearest-neighbor entry is obtained by measuring the normalized cosine similarity.
Formally, 
$$(O_a)_{(j, k) \in B_d} = (O_s)_{\textit{NN}(j, k) \in B_s}$$
where ${(j, k) \in B_d}$ represents the set of coordinates for each entry from $O_a$ within the target blob $B_d$ and $\textit{NN}(j, k) \in B_s$ denotes the set of coordinates for each nearest-neighbor entry from $O_s$ within the source blob $B_s$.

\subsection{Extension for Real Images}
\label{sec:object_moving_real}

In order to extend our method for dragging objects in real images, we first extract the blobs parameters as we explain later, then, we need to invert the image. However, we found that directly applying theg DDIM inversion~\cite{song2020denoising} in a localized model is not able to preserve the details of the input image, even without classifier-free guidance \cite{Ho2022ClassifierFreeDG}. Using more advanced inversion methods~\cite{huberman2023edit, qi2023fatezero} will not work in our case, as they preserve the general structure of the scene, whereas we are interested in changing the scene layout significantly.

Recall that when dealing with generated images, the input signal from the source image is fed in our pipeline through its self-attention outputs. Hence, we only need to extract the self-attention features in different attention layers and different denoising steps from the real image, rather than an actual inversion that searches for the optimal latent noises. 
To this end, we propose the \emph{DDPM self-attention bucketing}: we first add \emph{independent} noises with various scales to the real image, where the noise scale corresponds to a time step in the DDPM forward process. The noisy images at every time step, along with the above extracted blobs, are then passed to the localized model to get self-attention outputs in every attention layer, as needed. Note that the DDPM self-attention bucketing is specifically designed for the object dragging task, where we aim to preserve the visual details of the real image. It may not be suitable for other image editing tasks that change the object appearance or category.

For extracting the blobs representations from real images, we utilize ODISE~\cite{Xu2023OpenVocabularyPS} to get instance segmentation maps, then we use an ellipse fitting optimization with the goal of maximizing the Intersection Over Union (IOU) between the ellipse and the generated mask. Finally, we crop a local region around each blob and use  LLaVA-1.5~\cite{Liu2023ImprovedBW} for the local captioning.

Finally, in order to better preserve the background, we incorporated the Blended Latent Diffusion \cite{blended_2022_CVPR, avrahami2023blendedlatent} method into our process in which the background pixels are being integrated into the diffusion process in order to seamlessly blend the generated result in the original scene. For more details, please refer to the supplementary material.

\section{Experiments}
\label{sec:experiments}

In \Cref{sec:comparisons}, we compare our method against several baselines, both qualitatively and quantitatively. Next, in \Cref{sec:user_study} we describe the user study on various methods and present the outcome. Lastly, in \Cref{sec:ablation_study}, we show the ablation study results to highlight the importance of each component.

\subsection{Qualitative and Quantitative Comparison}
\label{sec:comparisons}

\begin{table}
    \centering
    \caption{\textbf{Quantitative Comparison.} We compare our method against the baselines in terms of foreground similarity (higher is better), object traces (lower is better) and realism (lower is better). As can be seen, DiffEditor~\cite{Mou2024DiffEditorBA} and DragonDiffusion~\cite{Mou2023DragonDiffusionED} struggle with object traces as they suffer from the object traces issue. PBE~\cite{Yang2022PaintBE}, Anydoor~\cite{Chen2023AnyDoorZO}, DragDiffusion~\cite{Shi2023DragDiffusionHD} and Diffusion SG~\cite{self_guidance} struggle with foreground similarity as they tend not to drag the object. In contrast, our method significantly outperforms all the baselines in terms of object traces and also achieves higher foreground similarity with comparable image realism.}
    \begin{adjustbox}{width=1\columnwidth}
        \begin{tabular}{>{\columncolor[gray]{0.95}}lccc}
            \toprule

            \textbf{Method} & 
            Foreground $(\uparrow)$ &
            Traces $(\downarrow)$ &
            Realism $(\downarrow)$
            \\
            
            \midrule

            PBE \cite{Yang2022PaintBE} &
            0.614 &
            0.446 &
            0.0029
            \\

            DiffusionSG \cite{self_guidance} &
            0.515 &
            0.459 &
            0.0096
            \\

            Anydoor \cite{Chen2023AnyDoorZO} &
            0.684 &
            0.454 &
            0.0035
            \\

            DragDiffusion \cite{Shi2023DragDiffusionHD} &
            0.738 &
            0.603 &
            0.0011
            \\

            DragonDiffusion \cite{Mou2023DragonDiffusionED} &
            0.818 &
            0.773 &
            0.0009
            \\

            DiffEditor \cite{Mou2024DiffEditorBA} &
            0.826 &
            0.604 &
            \textbf{0.0008}
            \\

            \midrule

            DiffUHaul (ours) &
            \textbf{0.835} &
            \textbf{0.32} &
            \textbf{0.0008}
            \\
            
            \bottomrule
        \end{tabular}
    \end{adjustbox}
    \label{tab:quantitative_comparison}
\end{table}

\begin{table}
    \centering
    \caption{\textbf{Ablation Study.} We ablate the following components of our method: (1) w/o gated self-attention (GSA) masking, (2) w/o self-attention (SA) sharing, (3) w/o soft attention anchoring and (4) w/o DDPM noising. As can be seen, removing the (1) GSA masking harms the foreground similarity, as leakages from neighboring blobs can interfere. Removing the (2) SA sharing or the (3) soft attention anchoring harms the foreground similarity as well, as it reduces the similarity the input image. Removing the DDPM SA bucketing slightly improves the object traces but significantly harms the foreground similarity, as the details of source images are not well preserved.}
    \begin{adjustbox}{width=1\columnwidth}
        \begin{tabular}{>{\columncolor[gray]{0.95}}lccc}
            \toprule

            \textbf{Method} & 
            Foreground $(\uparrow)$ &
            Traces $(\downarrow)$ &
            Realism $(\downarrow)$
            \\
            \midrule
            
            DiffUHaul (ours) &
            \textbf{0.835} &
            0.320 &
            \textbf{0.0008}
            \\
            
            \midrule

            w/o GSA masking &
            0.823 &
            0.320 &
            \textbf{0.0008}
            \\

            w/o SA sharing &
            0.755 &
            0.314 &
            0.0014
            \\

            w/o soft attention anchoring &
            0.780 &
            0.322 &
            0.0008
            \\

            w/o DDPM SA bucketing &
            0.675 &
            \textbf{0.298} &
            0.0029
            \\
            
            \bottomrule
        \end{tabular}
    \end{adjustbox}
    \label{tab:ablation_study}
\end{table}

We compared our method against the most relevant available object dragging baselines. Paint-By-Example (PBE) \cite{Yang2022PaintBE} and AnyDoor \cite{Chen2023AnyDoorZO} present a way of adding an object to an image. To use them for object dragging, we crop the object from the source image, apply the image inpainting in the cropped region, and then add the object in the new location.
Diffusion Self-Guidance (Diffusion SG) \cite{self_guidance} tackles the general image editing tasks via attention guidance, which can be tailored to object dragging.
DragDiffusion \cite{Shi2023DragDiffusionHD} is designed for the task of keypoint-based dragging, which we can convert into object dragging by selecting multiple points on the source object. Finally, DragonDiffusion \cite{Mou2023DragonDiffusionED} and DiffEditor \cite{Mou2024DiffEditorBA} directly tackle the problem of object dragging. For more details, please see the supplementary material.

As can be seen in \Cref{fig:qualitative_comparison}, PBE \cite{Yang2022PaintBE}, Anydoor \cite{Chen2023AnyDoorZO} and DiffusionSG \cite{self_guidance} can hardly preserve the appearance of the edited object and always have undesirable objects or artifacts left in the source location, indicating that existing general-purpose image editing methods tend to completely fail in the object dragging task. DragDiffusion \cite{Shi2023DragDiffusionHD} struggles with moving the object to the target location, while DragonDiffusion \cite{Mou2023DragonDiffusionED} and DiffEditor \cite{Mou2024DiffEditorBA} often suffers from the object traces issue, where the object appears in both source and target locations. In contrast, our method strikes the best balance between effectively dragging the object to the right position and preserving its visual appearance.

\begin{figure*}[t]
    \centering
    \setlength{\tabcolsep}{5pt}
    \renewcommand{\arraystretch}{0.9}
    \setlength{\ww}{0.28\columnwidth}

    \begin{tabular}{cccccc}
        \rotatebox[origin=c]{90}{{Input}}
        {\includegraphics[valign=c, width=\ww]{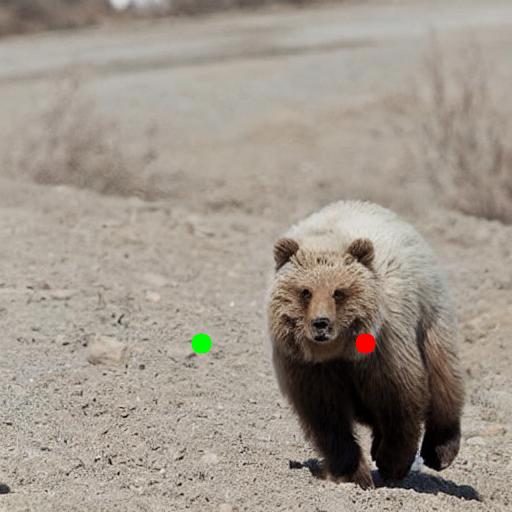}} &
        {\includegraphics[valign=c, width=\ww]{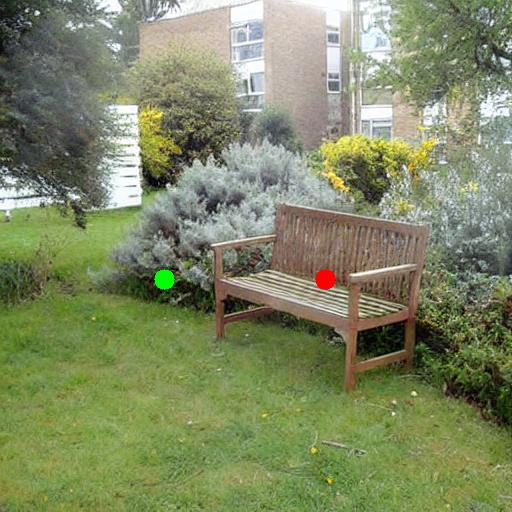}} &
        {\includegraphics[valign=c, width=\ww]{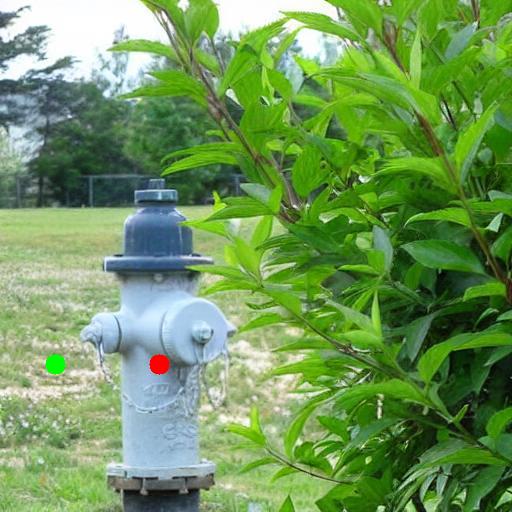}} &
        {\includegraphics[valign=c, width=\ww]{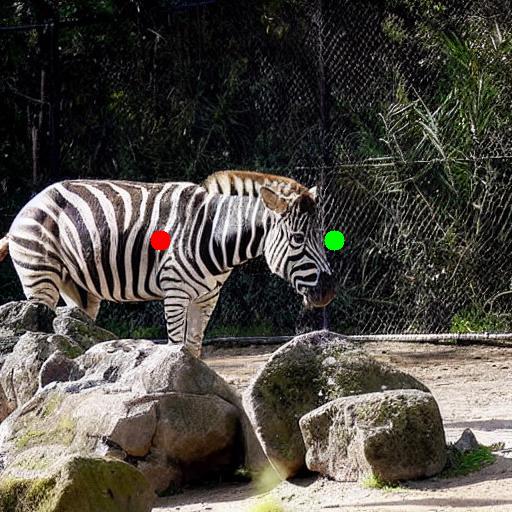}} &
        {\includegraphics[valign=c, width=\ww]{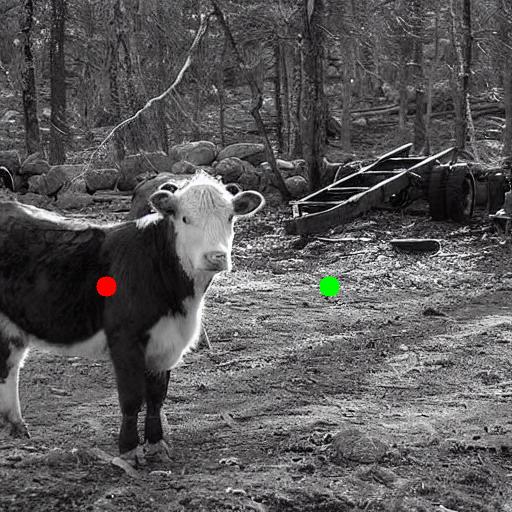}} &
        {\includegraphics[valign=c, width=\ww]{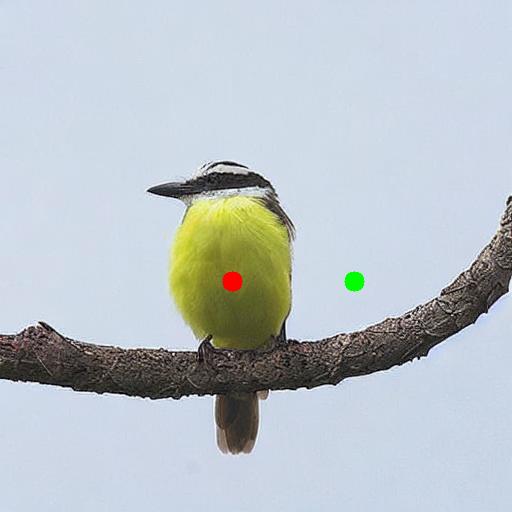}}
        \vspace{1px}
        \\

        \midrule

        \rotatebox[origin=c]{90}{{PBE}}
        {\includegraphics[valign=c, width=\ww]{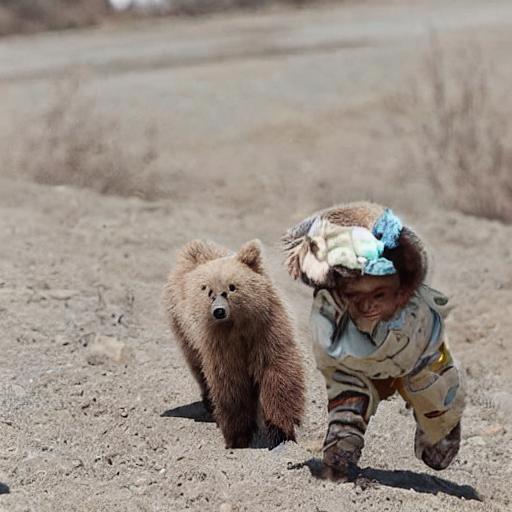}} &
        {\includegraphics[valign=c, width=\ww]{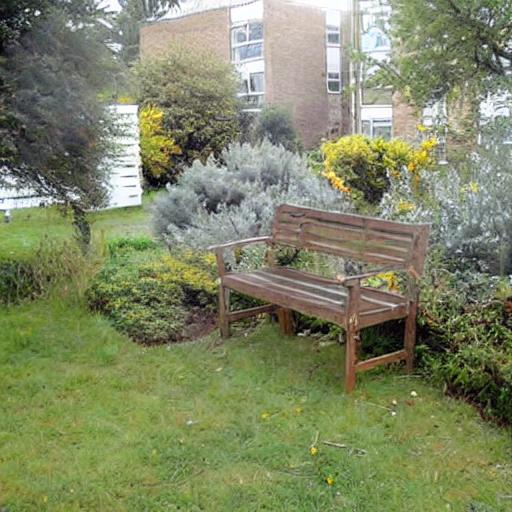}} &
        {\includegraphics[valign=c, width=\ww]{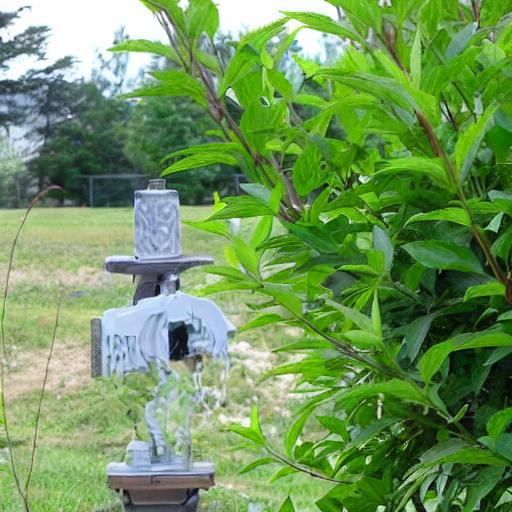}} &
        {\includegraphics[valign=c, width=\ww]{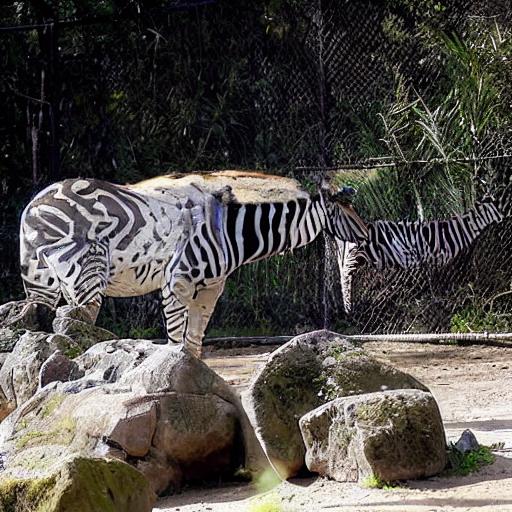}} &
        {\includegraphics[valign=c, width=\ww]{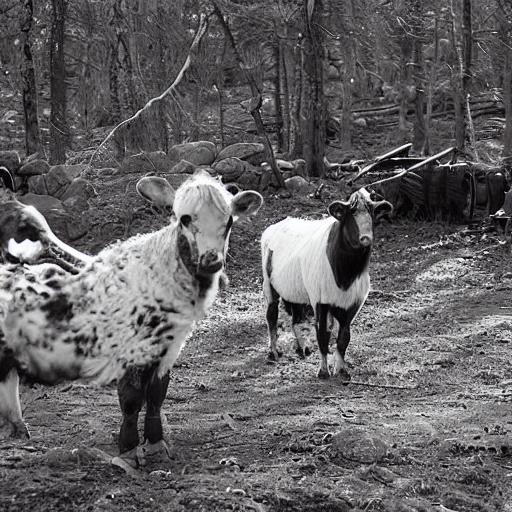}} &
        {\includegraphics[valign=c, width=\ww]{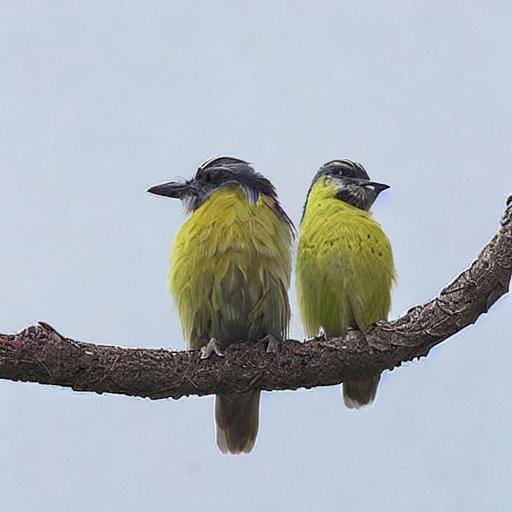}}
        \vspace{1px}
        \\

        \rotatebox[origin=c]{90}{{Diffusion SG}}
        {\includegraphics[valign=c, width=\ww]{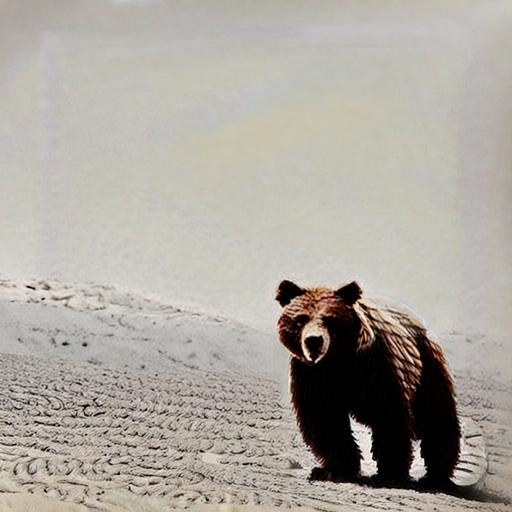}} &
        {\includegraphics[valign=c, width=\ww]{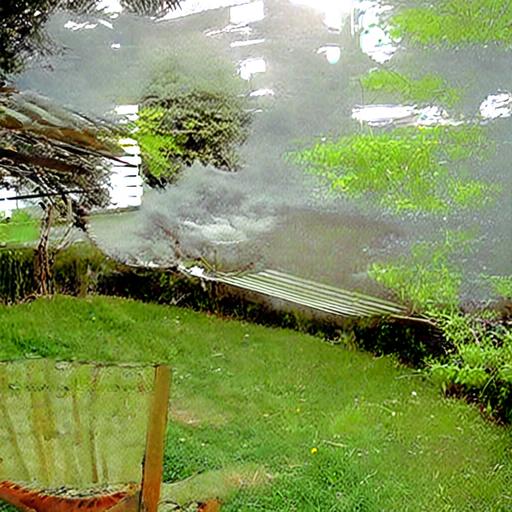}} &
        {\includegraphics[valign=c, width=\ww]{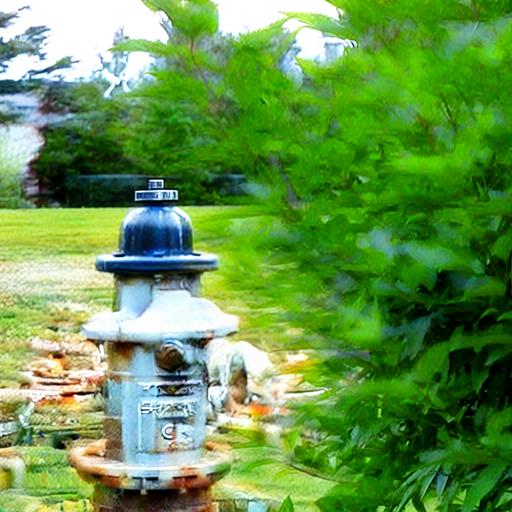}} &
        {\includegraphics[valign=c, width=\ww]{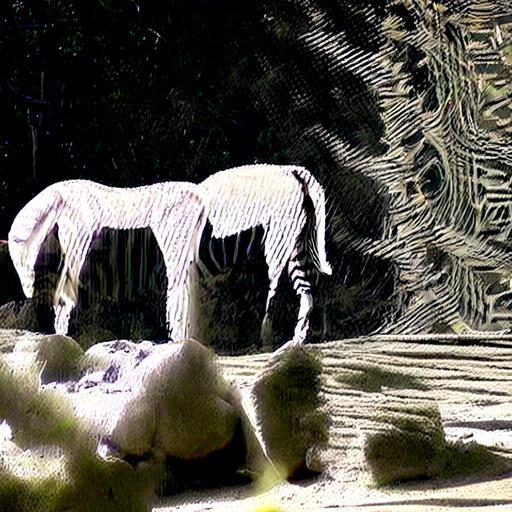}} &
        {\includegraphics[valign=c, width=\ww]{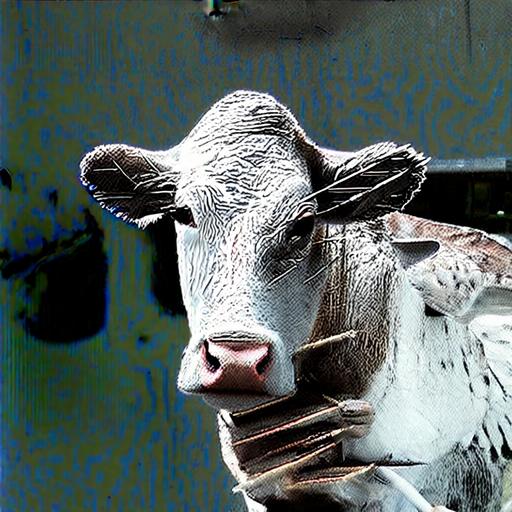}} &
        {\includegraphics[valign=c, width=\ww]{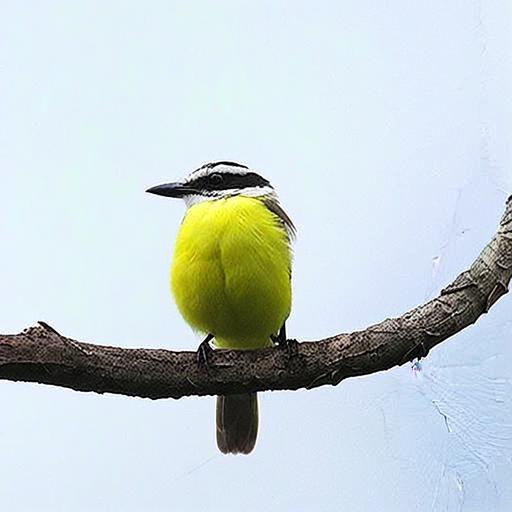}}
        \vspace{1px}
        \\

        \rotatebox[origin=c]{90}{{AnyDoor}}
        {\includegraphics[valign=c, width=\ww]{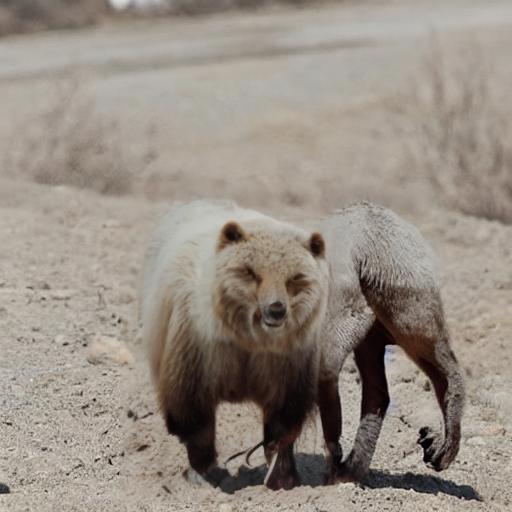}} &
        {\includegraphics[valign=c, width=\ww]{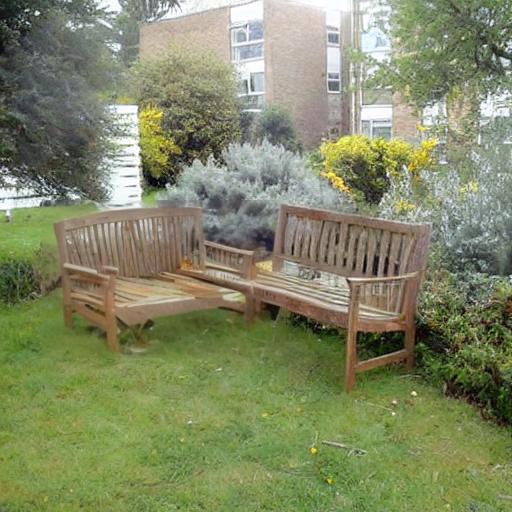}} &
        {\includegraphics[valign=c, width=\ww]{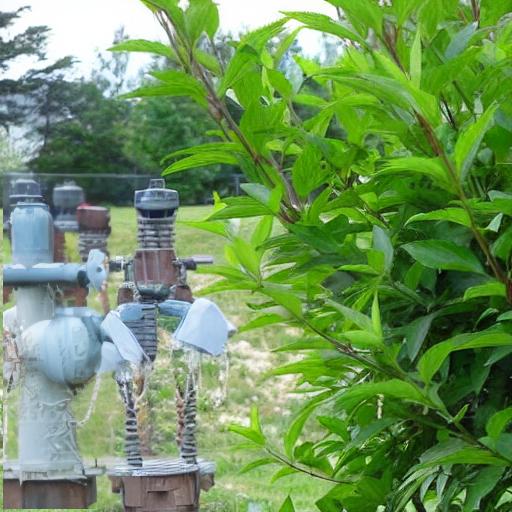}} &
        {\includegraphics[valign=c, width=\ww]{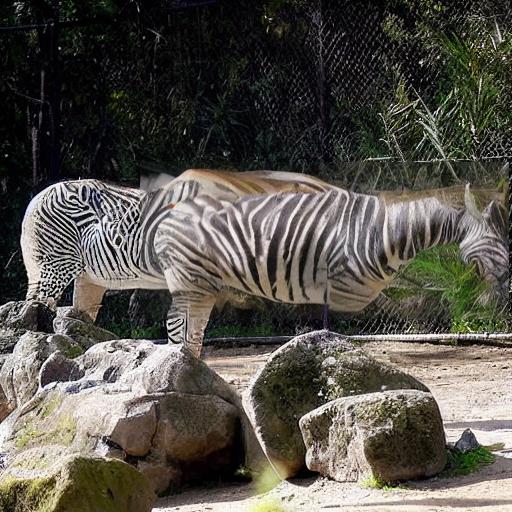}} &
        {\includegraphics[valign=c, width=\ww]{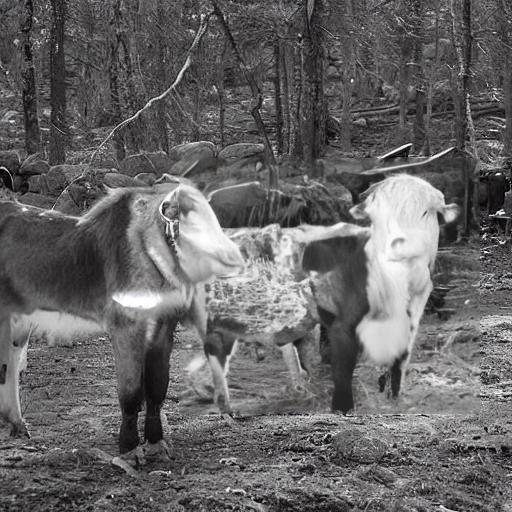}} &
        {\includegraphics[valign=c, width=\ww]{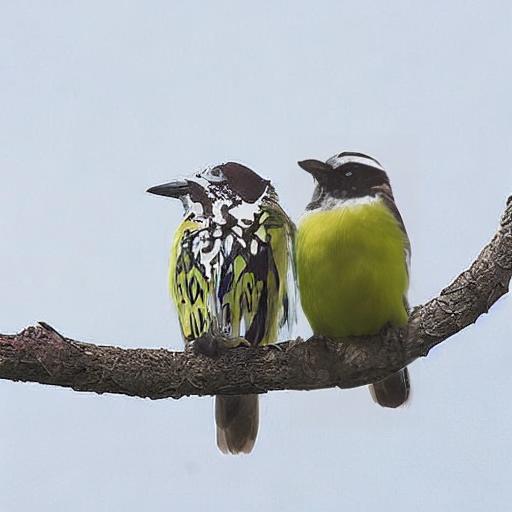}}
        \vspace{1px}
        \\

        \rotatebox[origin=c]{90}{{DragDiffusion}}
        {\includegraphics[valign=c, width=\ww]{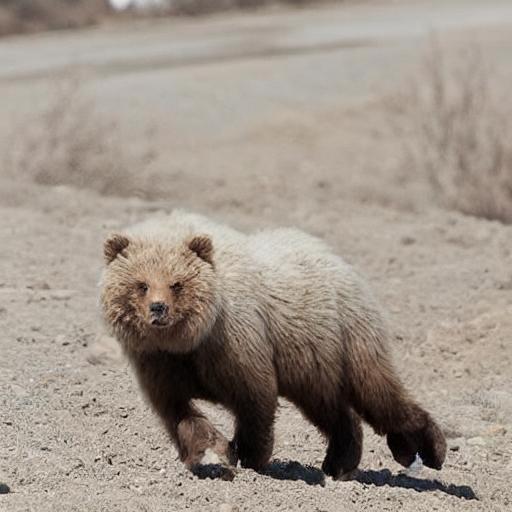}} &
        {\includegraphics[valign=c, width=\ww]{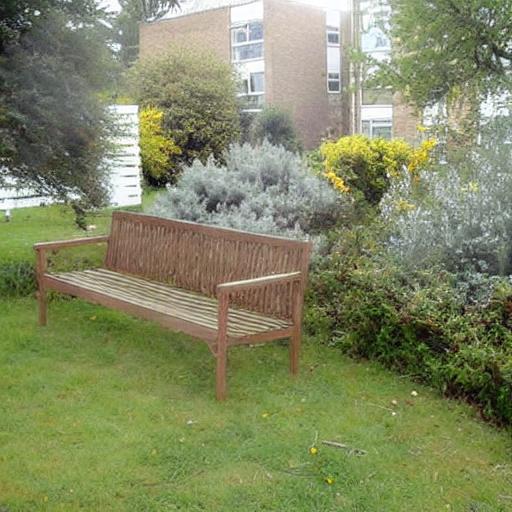}} &
        {\includegraphics[valign=c, width=\ww]{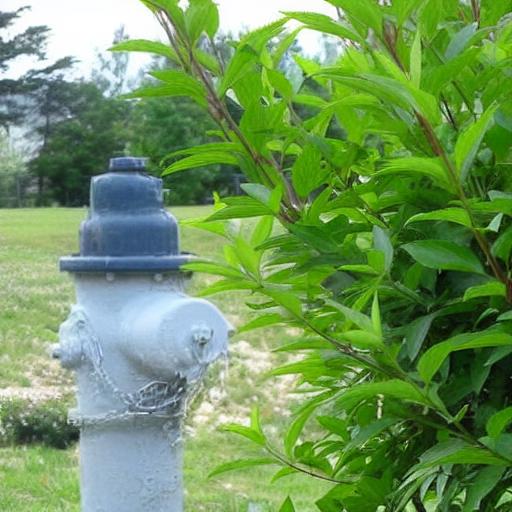}} &
        {\includegraphics[valign=c, width=\ww]{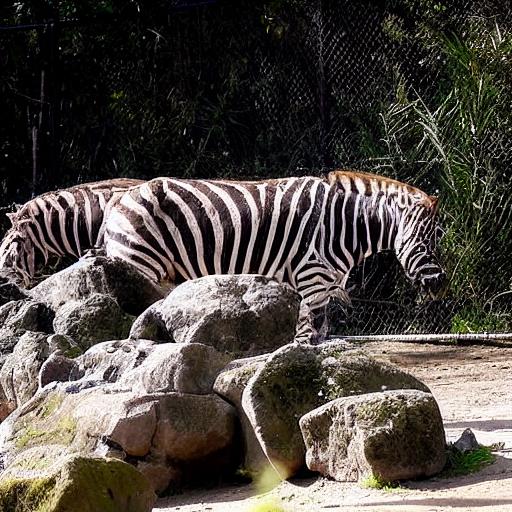}} &
        {\includegraphics[valign=c, width=\ww]{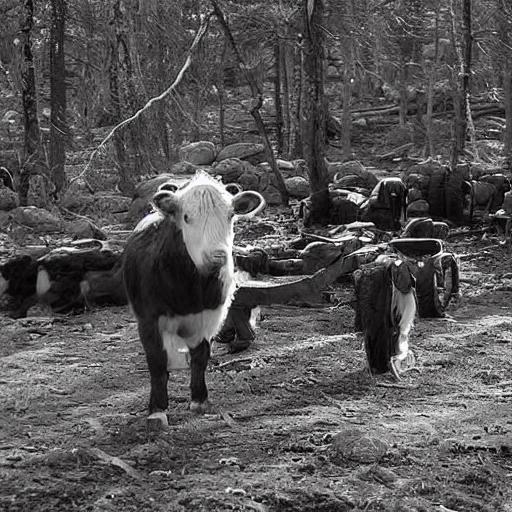}} &
        {\includegraphics[valign=c, width=\ww]{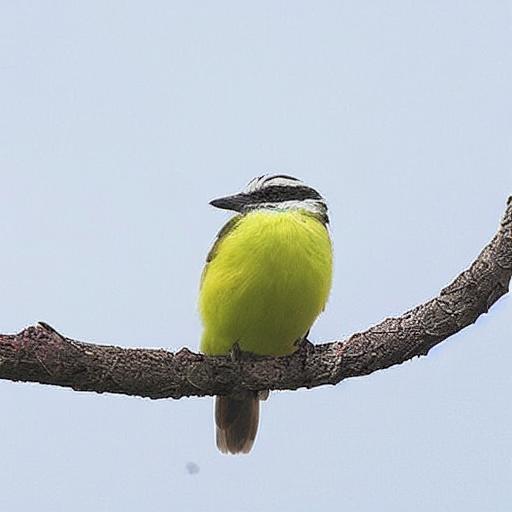}}
        \vspace{1px}
        \\

        \rotatebox[origin=c]{90}{{DragonDiffusion}}
        {\includegraphics[valign=c, width=\ww]{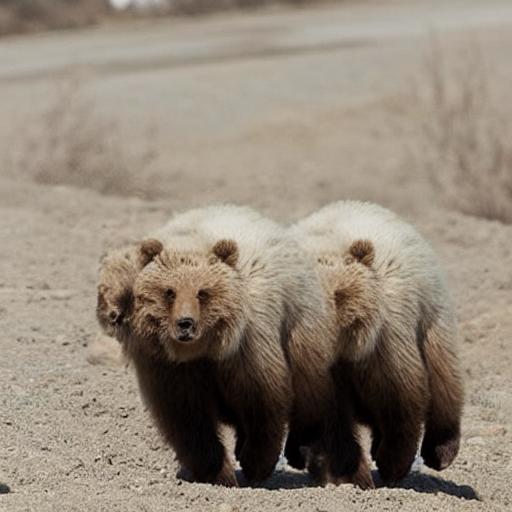}} &
        {\includegraphics[valign=c, width=\ww]{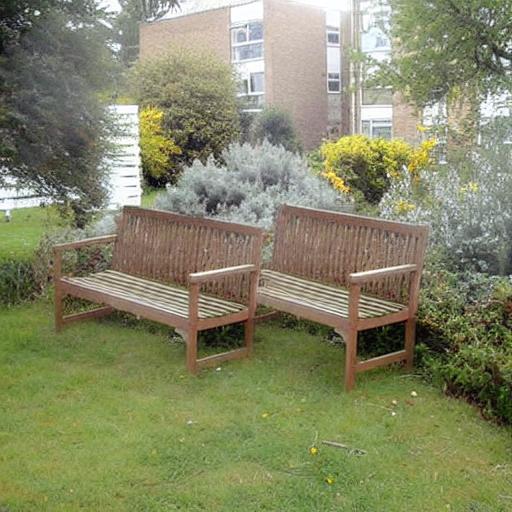}} &
        {\includegraphics[valign=c, width=\ww]{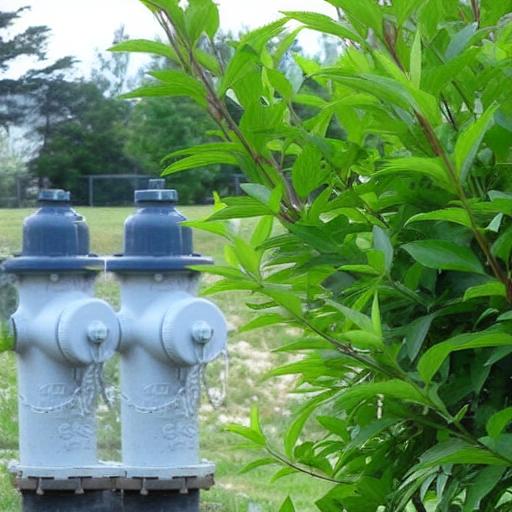}} &
        {\includegraphics[valign=c, width=\ww]{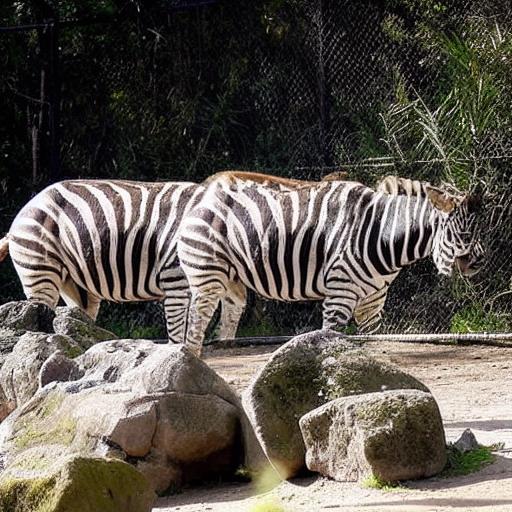}} &
        {\includegraphics[valign=c, width=\ww]{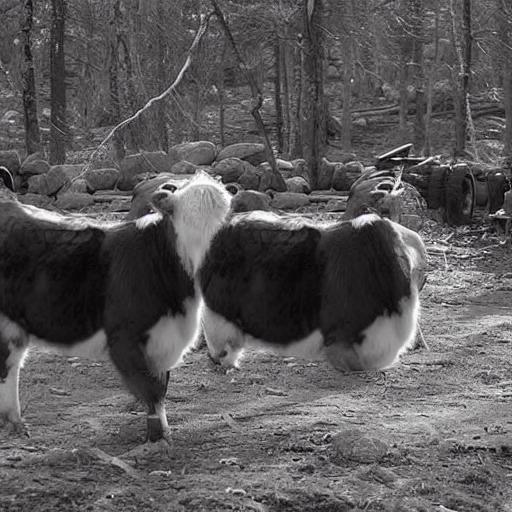}} &
        {\includegraphics[valign=c, width=\ww]{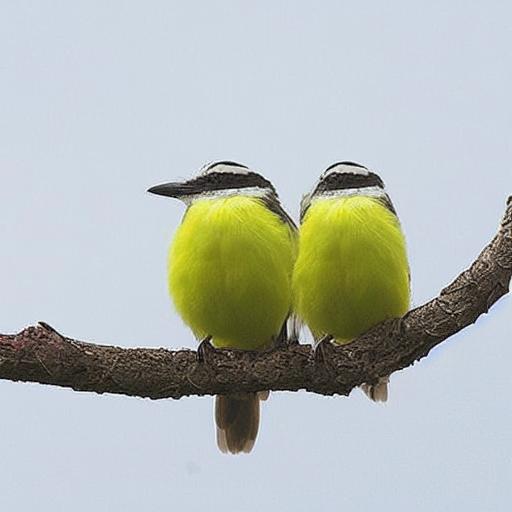}}
        \vspace{1px}
        \\

        \rotatebox[origin=c]{90}{{DiffEditor}}
        {\includegraphics[valign=c, width=\ww]{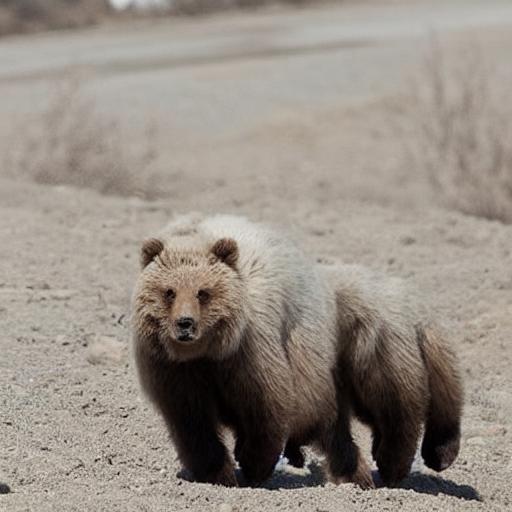}} &
        {\includegraphics[valign=c, width=\ww]{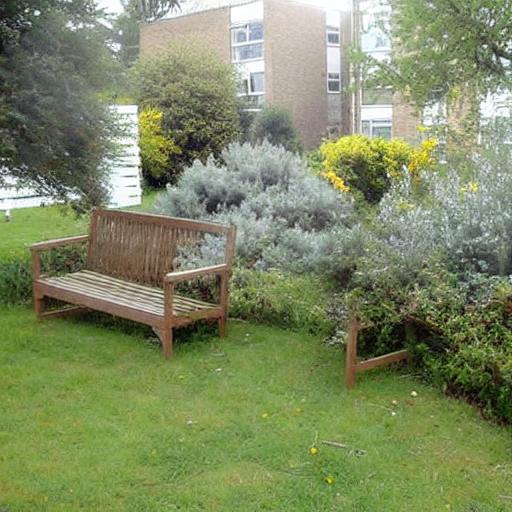}} &
        {\includegraphics[valign=c, width=\ww]{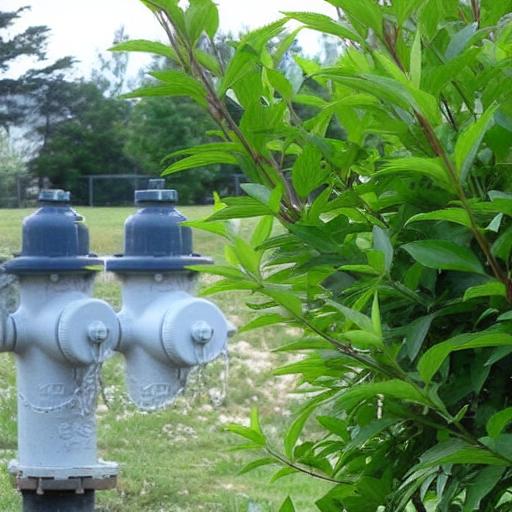}} &
        {\includegraphics[valign=c, width=\ww]{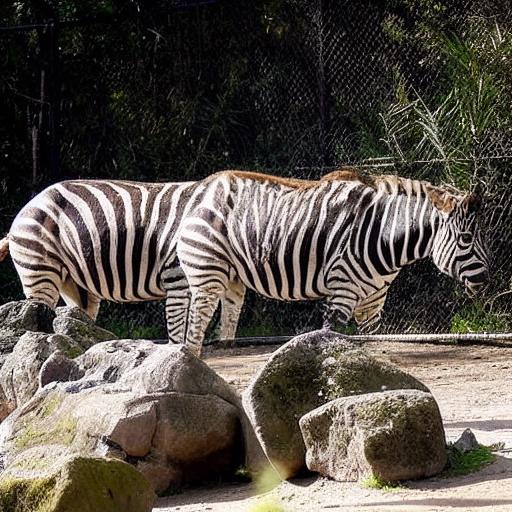}} &
        {\includegraphics[valign=c, width=\ww]{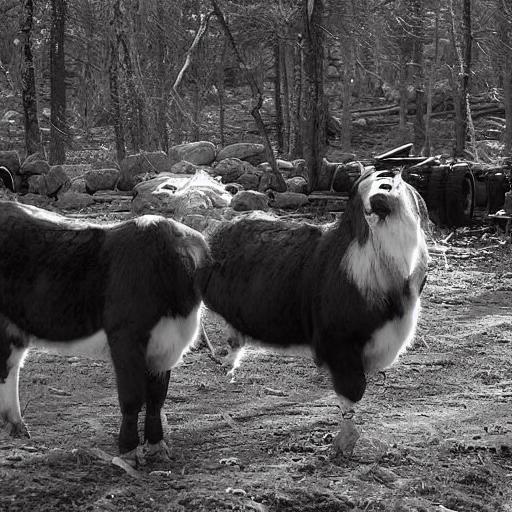}} &
        {\includegraphics[valign=c, width=\ww]{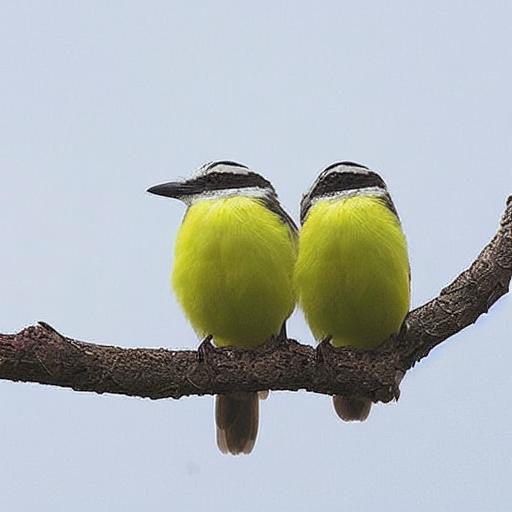}}
        \vspace{1px}
        \\

        \rotatebox[origin=c]{90}{{Ours}}
        {\includegraphics[valign=c, width=\ww]{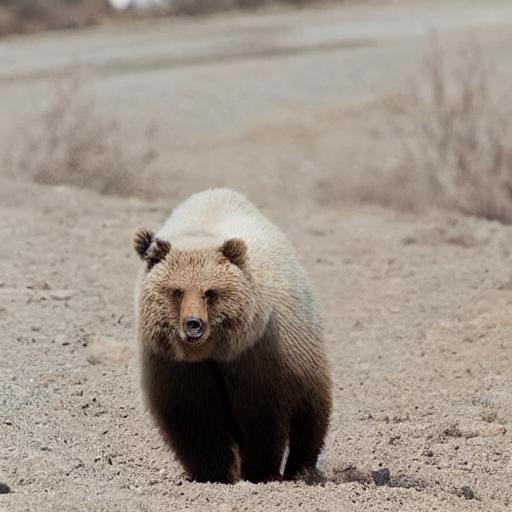}} &
        {\includegraphics[valign=c, width=\ww]{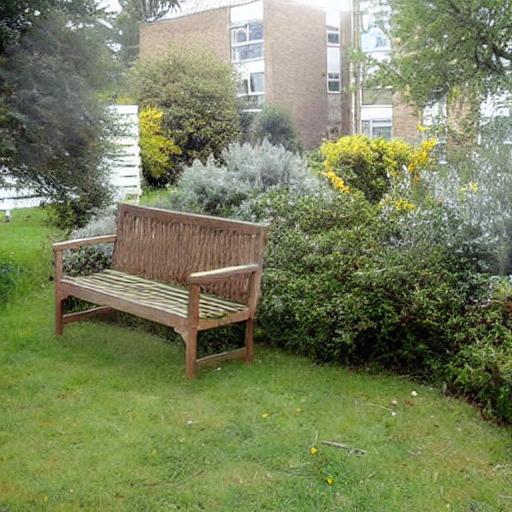}} &
        {\includegraphics[valign=c, width=\ww]{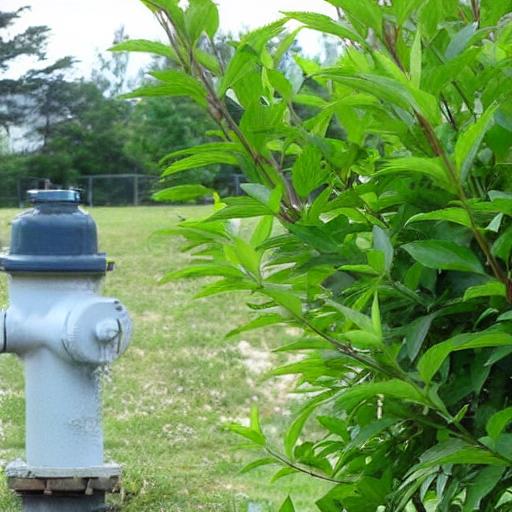}} &
        {\includegraphics[valign=c, width=\ww]{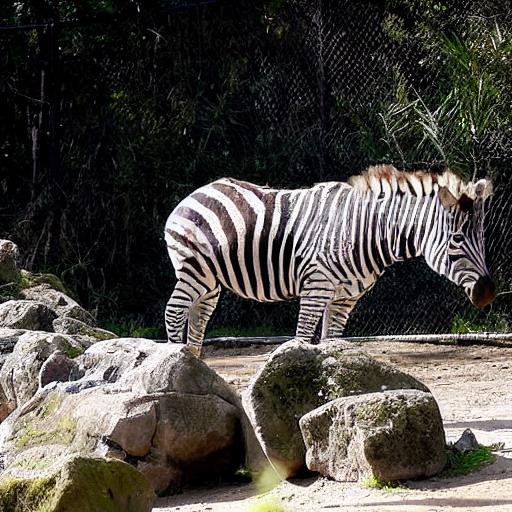}} &
        {\includegraphics[valign=c, width=\ww]{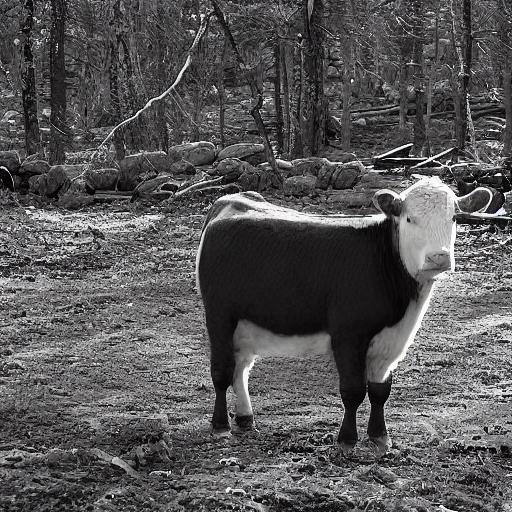}} &
        {\includegraphics[valign=c, width=\ww]{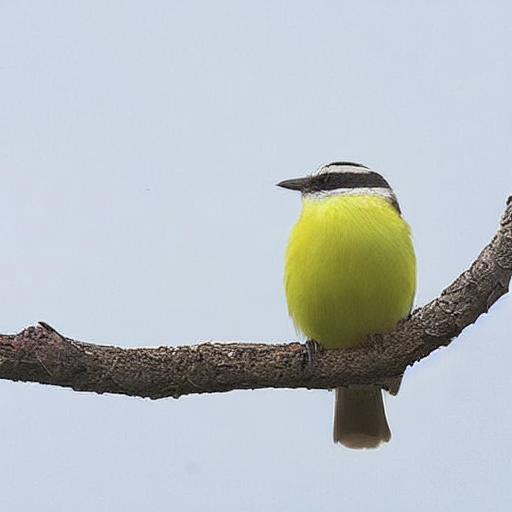}}
        \\

    \end{tabular}
    
    \caption{\textbf{Qualitative Automatic Comparison.} As explained in \Cref{sec:comparisons}, we used a filtered version of COCO validation set \cite{Lin2014MicrosoftCC}. The source and target locations are denoted by red and green points, respectively. As can be seen, PBE \cite{Yang2022PaintBE}, DiffusionSG \cite{self_guidance} and Anydoor \cite{Chen2023AnyDoorZO} mainly suffer from a bad preservation of the foreground object. DragDiffusion \cite{Shi2023DragDiffusionHD} struggles with dragging the object, while DragonDiffusion \cite{Mou2023DragonDiffusionED} and DiffEditor \cite{Mou2024DiffEditorBA} suffers from object traces. Our method, on the other hand, strikes the balance between dragging the object and preserving its identity.}
    \label{fig:qualitative_automatic_comparison}
\end{figure*}

To quantify the performance of our method and baselines, we prepare a specialized evaluation dataset based on the COCO~\cite{Lin2014MicrosoftCC} validation set. We first filter it to contain only images that have a single ``thing'' object with a prominent size. Then, we use the same blobs extraction pipeline as explained in \Cref{sec:object_moving_real}. For the object dragging task, we randomly sample a new location in the pixel space as the center of the target blob $B_d$. For each sample, we calculate 8 different target drag locations, resulting in a total dataset of 6,048 samples. For more details, please read the supplementary material. In \Cref{fig:qualitative_automatic_comparison}, we provide a qualitative comparison on the automatic dataset, where we make similar observations as before.

Based on this new dataset, we propose three evaluation metrics: \textit{foreground similarity}, \textit{object traces} and \textit{realism}. Foreground similarity quantifies whether the source object indeed dragged to the target location without appearance changes. To this end, we crop a tight box area around the source blob $B_s$ in the source image $I_s$ and around the target blob $B_d$ in the target image $I'$, respectively, and pass the crops to DINOv2 \cite{Oquab2023DINOv2LR} to measure the perceptual similarity after aligning them to a canonical position and masking the background.
We strive to \emph{maximize} this metric.
To measure the object traces phenomenon, we crop a tight box area around the source blob $B_s$ in the source image $I_s$ and around the source blob $B_s$ in the target image $I'$. 
Next, we mask the target blob $B_d$ area in the target image $I'$. 
Similarly, we utilize DINOv2 \cite{Oquab2023DINOv2LR} to measure the perceptual similarity between the crops. We strive to \emph{minimize} this metric. Lastly, to measure the realism of the edited image, we utilize KID score \cite{Binkowski2018DemystifyingMG} of sets of 672 real and generated images. For more details, please read the supplementary material.

As can be seen in \Cref{tab:quantitative_comparison}, DiffEditor~\cite{Mou2024DiffEditorBA} and DragonDiffusion~\cite{Mou2023DragonDiffusionED} rank high in object traces as they suffer from object traces problem. PBE~\cite{Yang2022PaintBE}, Anydoor~\cite{Chen2023AnyDoorZO}, DragDiffusion~\cite{Shi2023DragDiffusionHD} and Diffusion SG~\cite{self_guidance} struggle with foreground similarity as they tend not to drag the object. On the other hand, our method significantly outperforms all the baselines in terms of object traces, which demonstrates the robustness of our method. In addition, it achieves higher foreground similarity and is on par in terms of image realism. These results are supported by the qualitative comparison.

\subsection{User Study}
\label{sec:user_study}

\begin{table}
    \centering
    \caption{\textbf{User Study.} We compare our method against the baselines using the standard two-alternative forced-choice format. Users were asked to rate which editing result is better (Ours vs. the baseline) in terms of: (1) dragging the object to the desired location (2) leaving no traces of the original object, (3) realism and (4) overall edit quality. The number represents the win rate of our method over each of the baselines. As we can see, our method wins the baselines in all terms more than the random win rate of 50\%.}
    \begin{adjustbox}{width=1\columnwidth}
        \begin{tabular}{>{\columncolor[gray]{0.95}}lcccc}
            \toprule

            \textbf{Ours vs} & 
            Dragging $(\uparrow)$ &
            No traces $(\uparrow)$ &
            Realism $(\uparrow)$ &
            Overall $(\uparrow)$
            \\
            
            \midrule

            PBE \cite{Yang2022PaintBE} &
            82.14\% &
            78.57\% &
            82.58\% &
            79.68\%
            \\

            DiffusionSG \cite{self_guidance} &
            79.46\% &
            76.78\% &
            77.45\% &
            77.45\%
            \\

            Anydoor \cite{Chen2023AnyDoorZO} &
            81.02\% &
            76.78\% &
            81.91\% &
            81.25\%
            \\

            DragDiffusion \cite{Shi2023DragDiffusionHD} &
            64.73\% &
            58.92\% &
            59.82\% &
            61.16\%
            \\

            DragonDiffusion \cite{Mou2023DragonDiffusionED} &
            77.00\% &
            75.22\% &
            81.02\% &
            76.56\%
            \\

            DiffEditor \cite{Mou2024DiffEditorBA} &
            73.43\% &
            70.08\% &
            70.08\% &
            76.33\%
            \\
            
            \bottomrule
        \end{tabular}
    \end{adjustbox}
    \label{tab:user_study}
\end{table}

We conduct an extensive user study using the Amazon Mechanical Turk (AMT) platform \cite{amt}, where the test examples are also sampled from the automatically extracted dataset as explained in \Cref{sec:comparisons}. We compare all the baselines using the standard two-alternative forced-choice format. Users were given the source image, the edit instructions and two edited images: one from our method and another one from a baseline. For each comparison, users were asked to rate which edited image is better in terms of: (1) dragging the object to the desired location (2) leaving no traces of the original object, (3) realism and (4) overall edit quality (i.e., taking all the aspects into account). As can be seen in \Cref{tab:user_study}, our method is preferred over all the baselines in terms of the overall edit quality and different individual perspectives. This observation aligns well with our automatic metrics. The user study suggests that DragDiffusion is the second-strongest baseline, it may be due to the fact that it also results with realistic images, as it also avoids leaving traces of the dragged object, which the automatically calculated KID do not take into account. For more details and statistical significance analysis, please read the supplementary material.

\subsection{Ablation Study}
\label{sec:ablation_study}

We perform the ablation study for the following components of our method: (1) \emph{Without gated self-attention masking} --- we remove the gated self-attention masking that is described in \Cref{sec:blobgen_entanglement}. (2) \emph{Without self-attention sharing} --- we remove the self-attention sharing component. (3) \emph{Without soft attention anchoring} --- we remove the soft attention anchoring that is described in \Cref{sec:object_moving_generated}. (4) \emph{Without DDPM noising} --- we replace the DDPM noising that is described in \Cref{sec:object_moving_real} with a DDIM inversion \cite{song2020denoising}.

We use the same automatic evaluation metrics as described in \Cref{sec:comparisons} to quantify the importance of each component. As can be seen in \Cref{tab:ablation_study}, removing the (1) GSA masking harms the foreground similarity, as leakages from neighboring blobs can interfere the visual appearances of the focused object. Removing the (2) SA sharing or the (3) soft attention anchoring harms the foreground similarity as well, as it reduces the similarity to the input image. Removing the DDPM noising slightly improves the object traces, but it significantly harms the foreground similarity, as the reconstructed image itself has changed significantly. For a qualitative visualization of the ablation study, please refer to the supplementary material.

\section{Limitations and Conclusions}
\label{sec:limitations_and_conclusions}

\begin{figure}[t]
    \centering
    \setlength{\tabcolsep}{1pt}
    \renewcommand{\arraystretch}{0.7}
    \setlength{\ww}{0.232\columnwidth}
    \hspace{-1pt}\begin{tabular}{ccccc}

        \rotatebox[origin=c]{90}{\footnotesize{(a) Rotation}} 
        &
        {\includegraphics[valign=c, width=\ww]{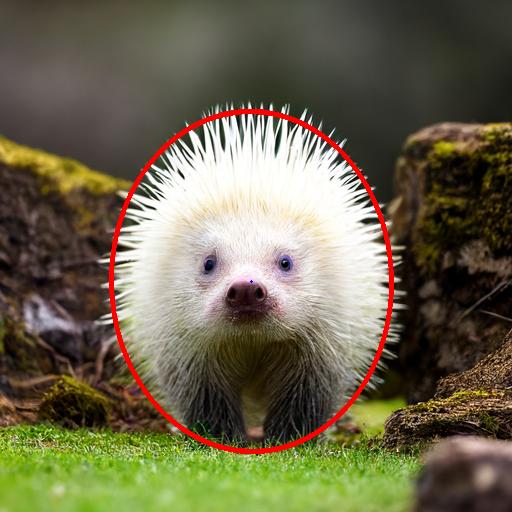}} &
        {\includegraphics[valign=c, width=\ww]{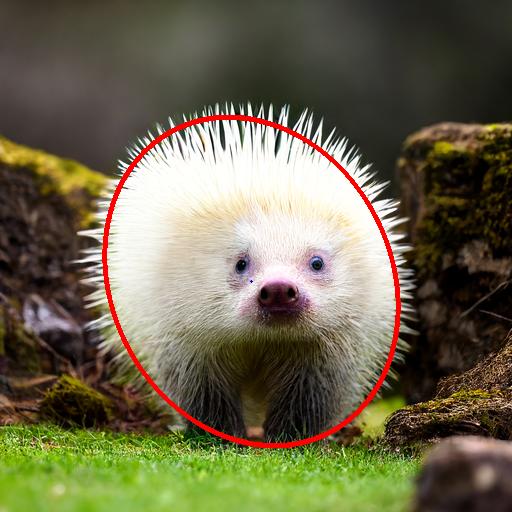}} &
        {\includegraphics[valign=c, width=\ww]{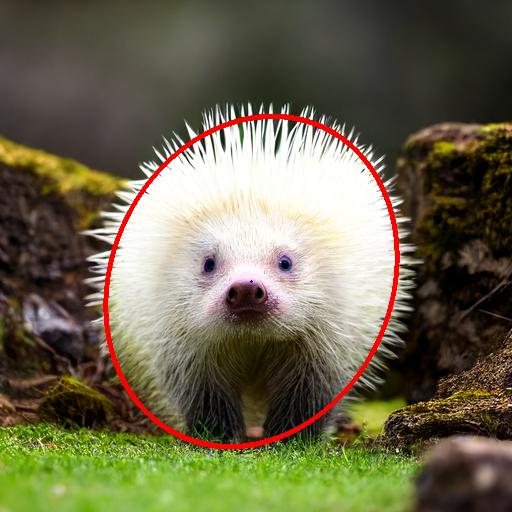}} &
        {\includegraphics[valign=c, width=\ww]{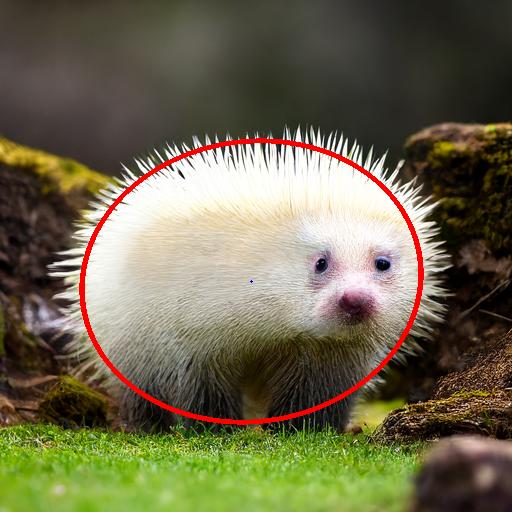}}
        \\
        
        &
        \footnotesize{Original} &
        \footnotesize{Rotation 1} &
        \footnotesize{Rotation 2} &
        \footnotesize{Rotation 3}
        \vspace{0.1cm}
        \\

        \rotatebox[origin=c]{90}{\footnotesize{(b) Resize}} 
        &
        {\includegraphics[valign=c, width=\ww]{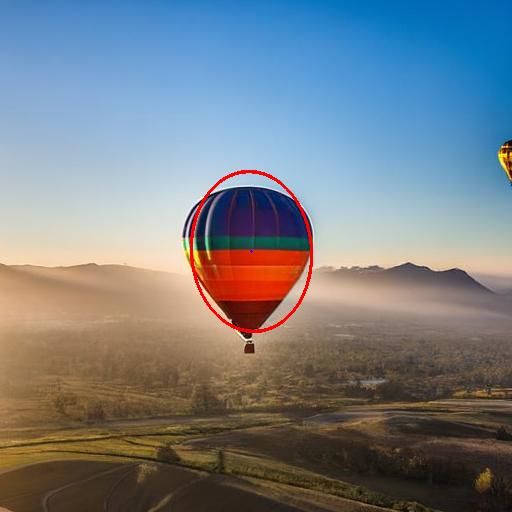}} &
        {\includegraphics[valign=c, width=\ww]{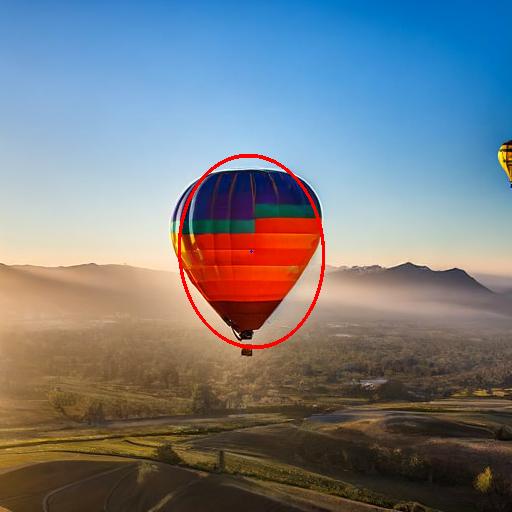}} &
        {\includegraphics[valign=c, width=\ww]{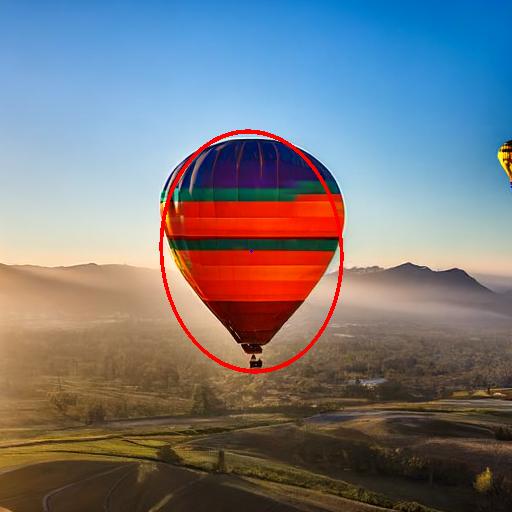}} &
        {\includegraphics[valign=c, width=\ww]{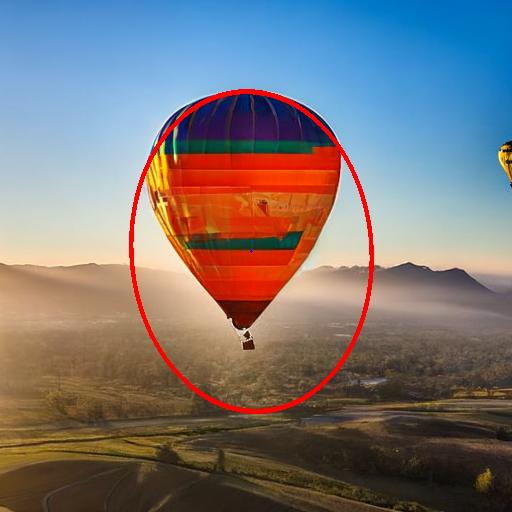}}
        \\
        
        &
        \footnotesize{Original} &
        \footnotesize{Resize 1} &
        \footnotesize{Resize 2} &
        \footnotesize{Resize 3}
        \vspace{0.1cm}
        \\

        \rotatebox[origin=c]{90}{\footnotesize{(c) Collision}} 
        &
        {\includegraphics[valign=c, width=\ww]{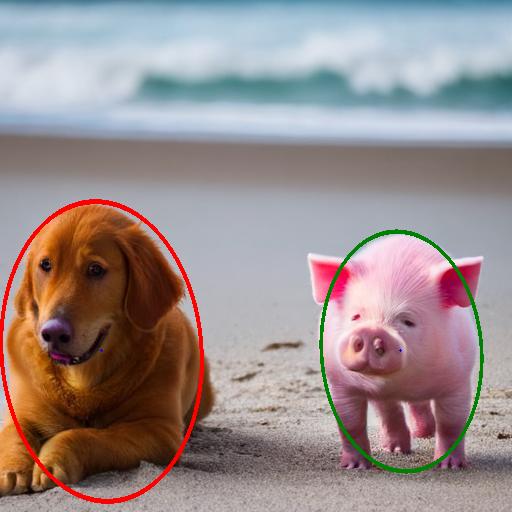}} &
        {\includegraphics[valign=c, width=\ww]{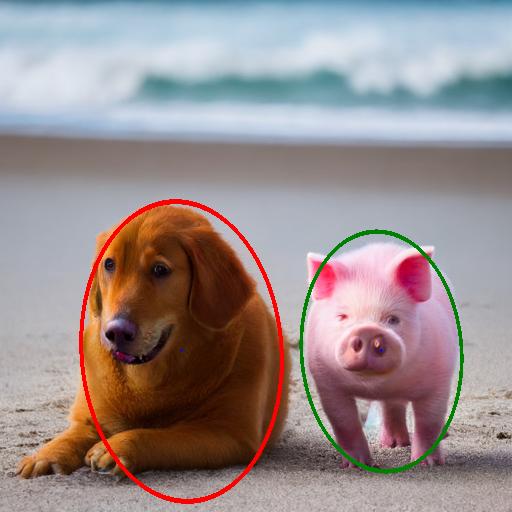}} &
        {\includegraphics[valign=c, width=\ww]{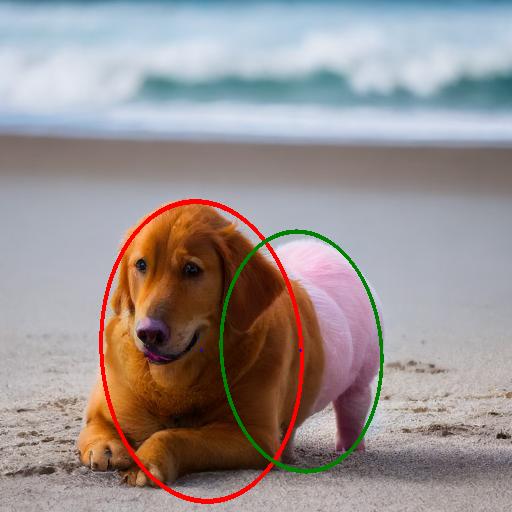}} &
        {\includegraphics[valign=c, width=\ww]{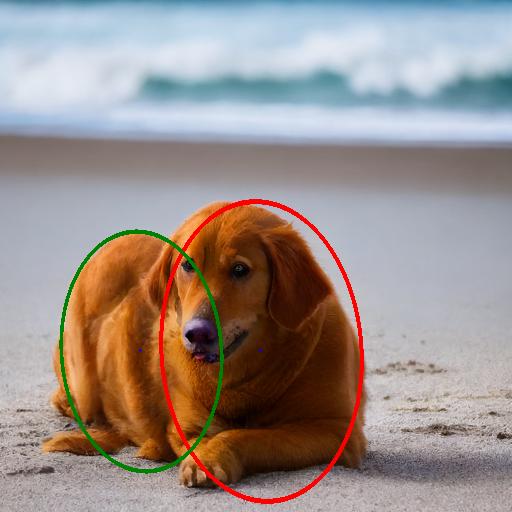}}
        \\
        
        &
        \footnotesize{Original} &
        \footnotesize{Drag 1} &
        \footnotesize{Drag 2} &
        \footnotesize{Drag 3}
        \vspace{0.1cm}
        \\

    \end{tabular}
    
    \caption{\textbf{Limitations.} Our method suffers from the following limitations: (a) We found our method to be incapable of rotating objects, and instead stretch the objects to fit the new blob shape without changing the orientation. (b) We found our method to struggle with resizing objects, especially in large resizes (\eg, Resize 3 in the second row). (c) We found our method to struggle to handle colliding objects while dragging, which may result with a hybrid between the objects (\eg, Drag 2 in the third row) or one object being merged (\eg, Drag 3 in the third row).}
    \label{fig:limitations}
\end{figure}

Our method suffers from the following limitations that are depicted in \Cref{fig:limitations}: (a) We found our diffusion anchoring technique introduced in \Cref{sec:object_moving_generated} to be incapable of rotating objects, and instead, as can be seen in \Cref{fig:limitations}(a), stretch the object to fit the new blob shape without changing the orientation, this may be caused due to the fact that rotation involves understanding the 3D structure, which is not reflected by the self-attention nearest-neighbor copying. (b) We found our method to struggle with resizing object, especially in large resizes, as can be seen in \Cref{fig:limitations}(b) Resize 3. (c) We found our method to struggle to handle colliding object while dragging, which may result with a hybrid between the objects (\Cref{fig:limitations}(c) Drag 2) or one object being merged (\Cref{fig:limitations}(c) Drag 3).

In conclusion, we presented DiffUHaul, our solution to the seemingly straightforward task of object dragging. We demonstrated that the spatial understanding of the \emph{localized} BlobGEN can be harnessed to this task, using our novel diffusion anchoring technique that manages to merge the location signal from the model with the object appearance signal from the input image. %

\begin{acks}
  This work was supported in part by the Israel Science Foundation (grants 2492/20, 3611/21 and 1574/21).
\end{acks}

\clearpage
\begin{figure*}[t]
    \centering
    \setlength{\tabcolsep}{5pt}
    \renewcommand{\arraystretch}{0.9}
    \setlength{\ww}{0.28\columnwidth}

    \begin{tabular}{cccccc}
        \rotatebox[origin=c]{90}{{Input}}
        {\includegraphics[valign=c, width=\ww]{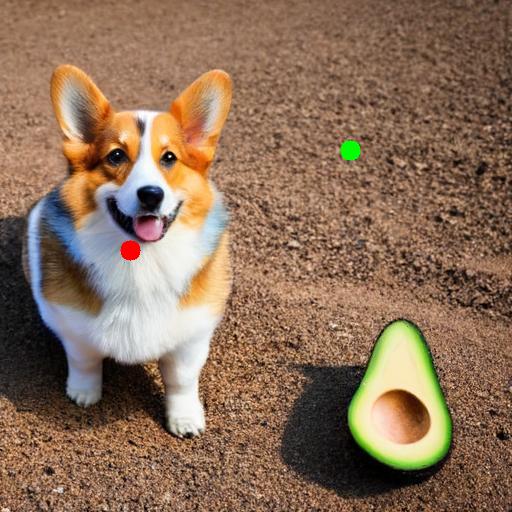}} &
        {\includegraphics[valign=c, width=\ww]{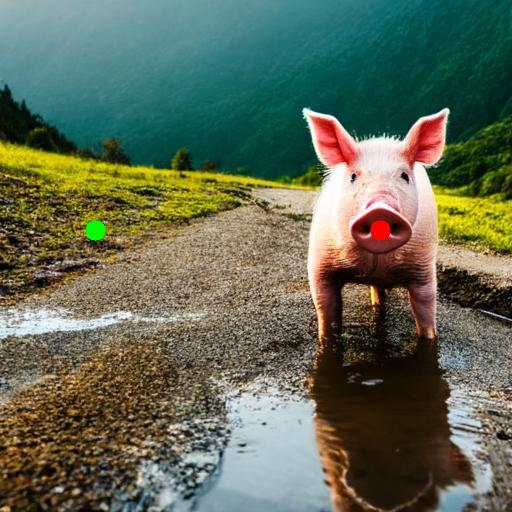}} &
        {\includegraphics[valign=c, width=\ww]{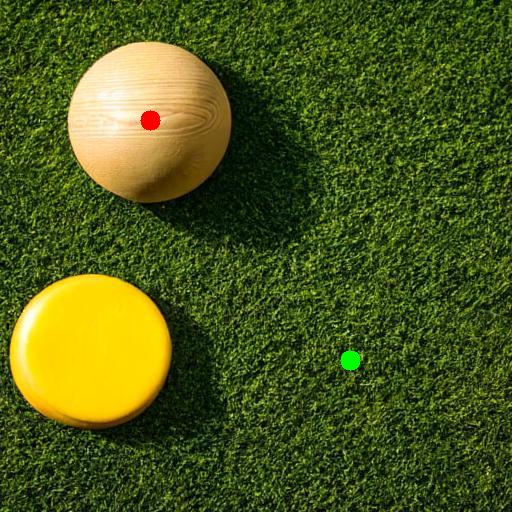}} &
        {\includegraphics[valign=c, width=\ww]{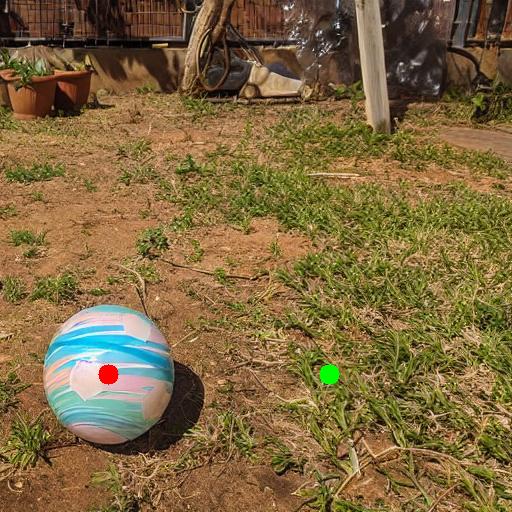}} &
        {\includegraphics[valign=c, width=\ww]{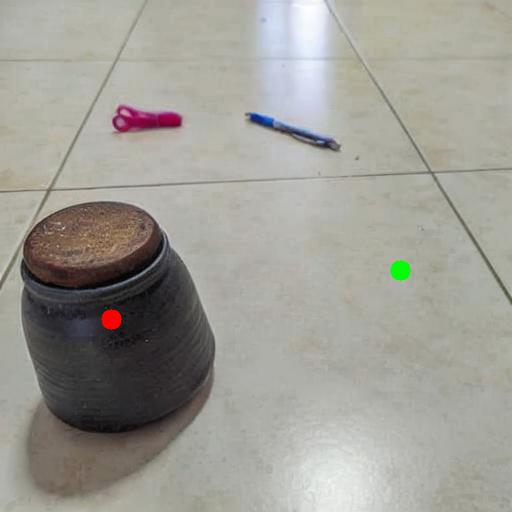}} &
        {\includegraphics[valign=c, width=\ww]{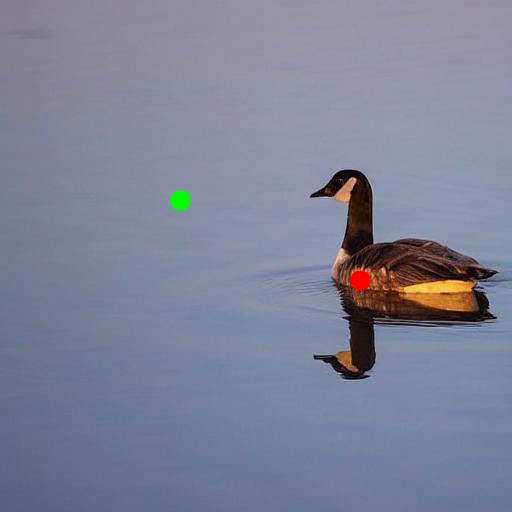}}
        \vspace{1px}
        \\

        \midrule

        \rotatebox[origin=c]{90}{{PBE}}
        {\includegraphics[valign=c, width=\ww]{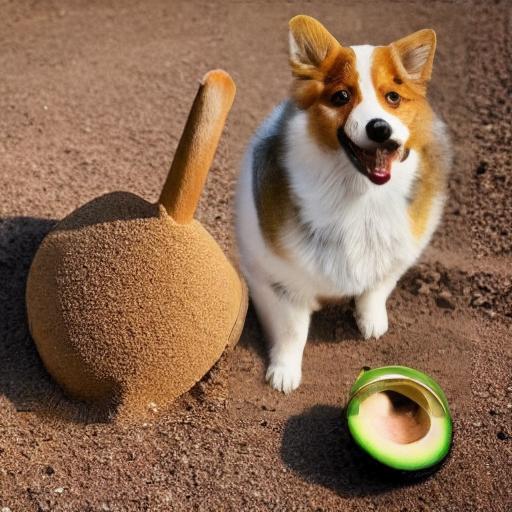}} &
        {\includegraphics[valign=c, width=\ww]{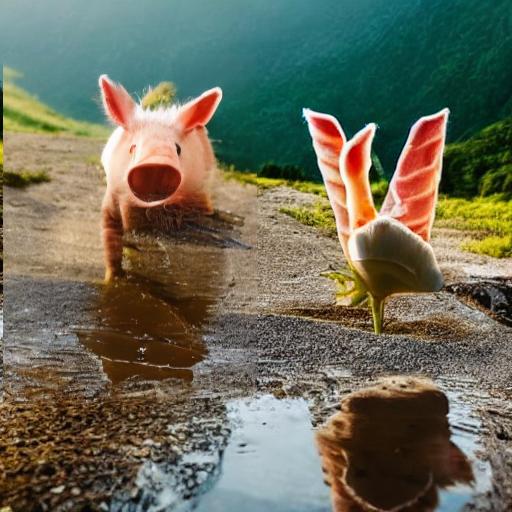}} &
        {\includegraphics[valign=c, width=\ww]{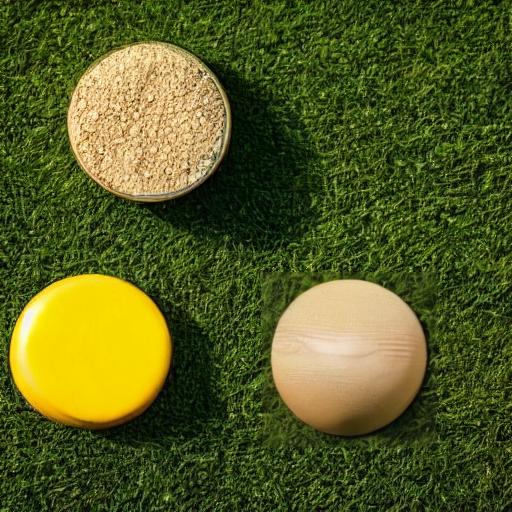}} &
        {\includegraphics[valign=c, width=\ww]{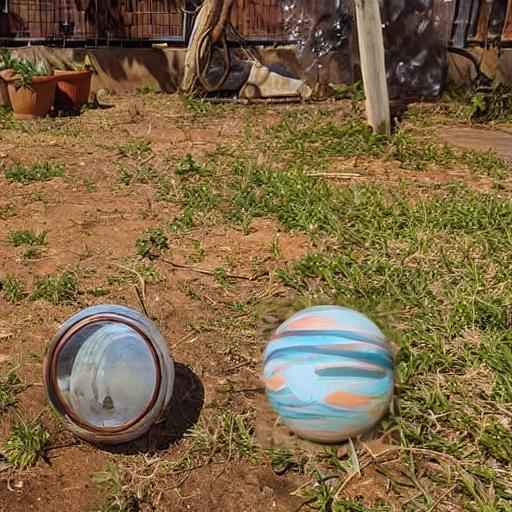}} &
        {\includegraphics[valign=c, width=\ww]{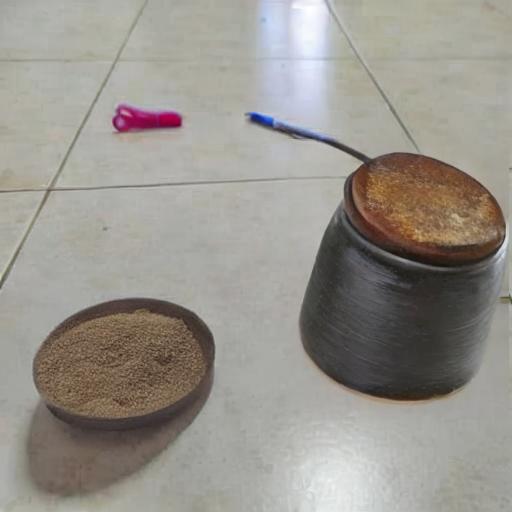}} &
        {\includegraphics[valign=c, width=\ww]{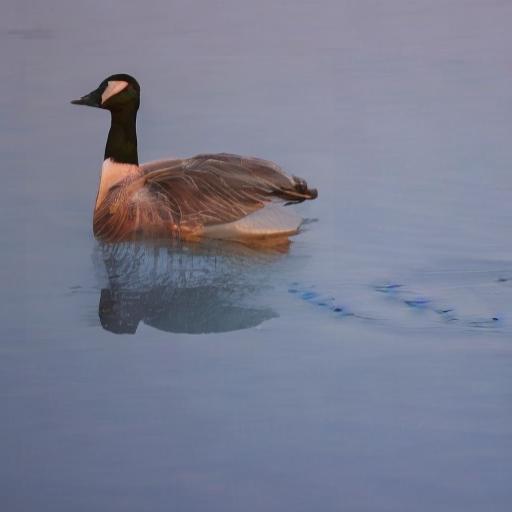}}
        \vspace{1px}
        \\

        \rotatebox[origin=c]{90}{{Diffusion SG}}
        {\includegraphics[valign=c, width=\ww]{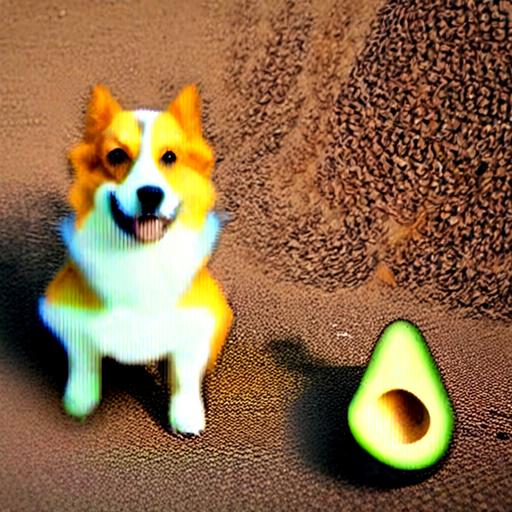}} &
        {\includegraphics[valign=c, width=\ww]{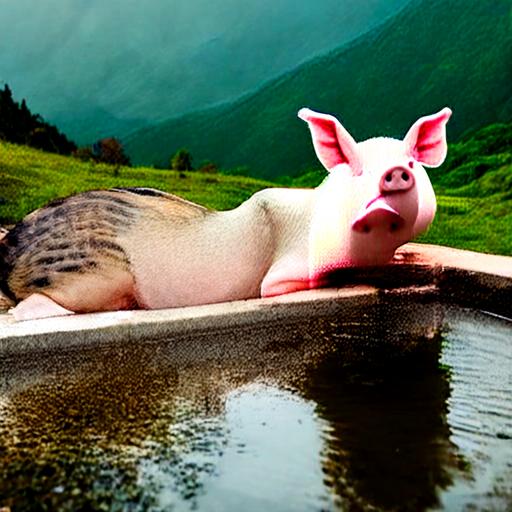}} &
        {\includegraphics[valign=c, width=\ww]{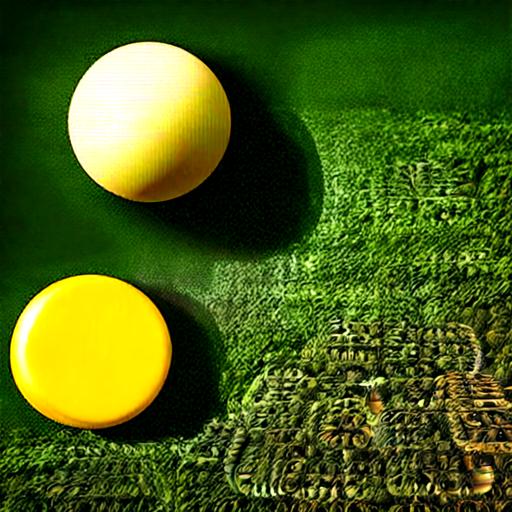}} &
        {\includegraphics[valign=c, width=\ww]{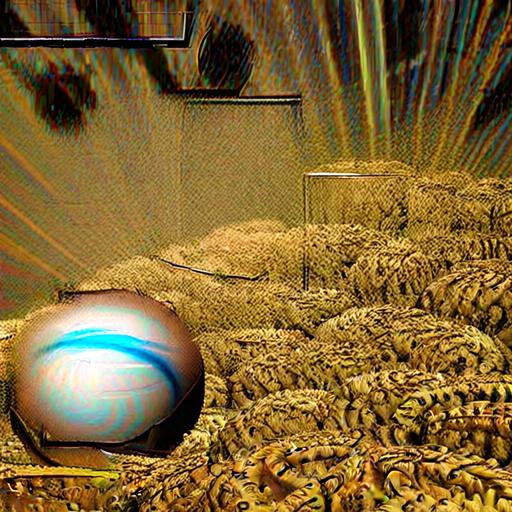}} &
        {\includegraphics[valign=c, width=\ww]{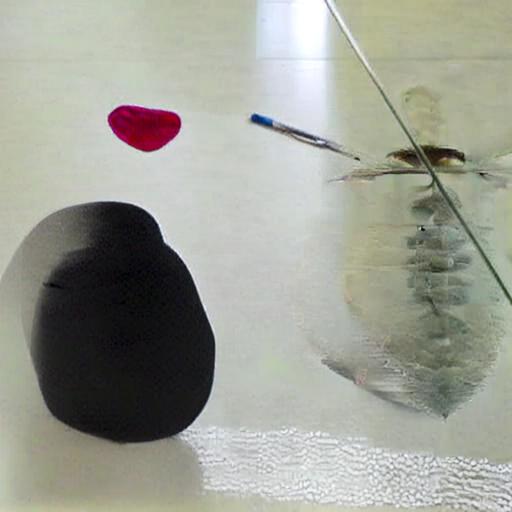}} &
        {\includegraphics[valign=c, width=\ww]{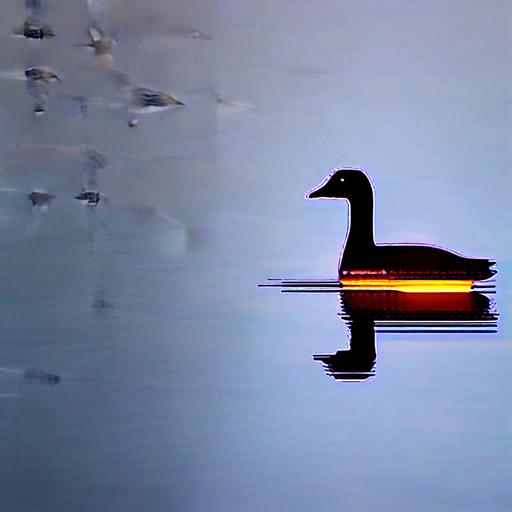}}
        \vspace{1px}
        \\

        \rotatebox[origin=c]{90}{{AnyDoor}}
        {\includegraphics[valign=c, width=\ww]{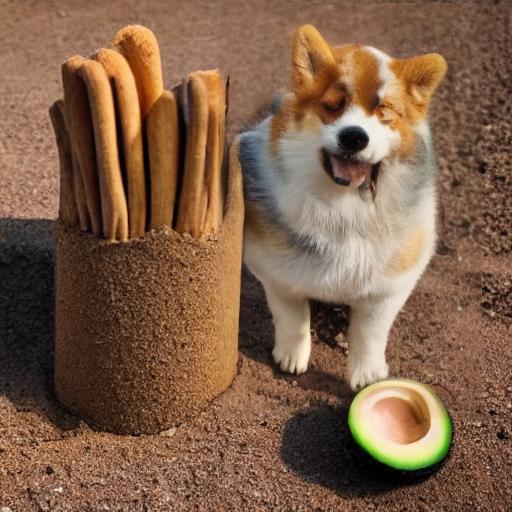}} &
        {\includegraphics[valign=c, width=\ww]{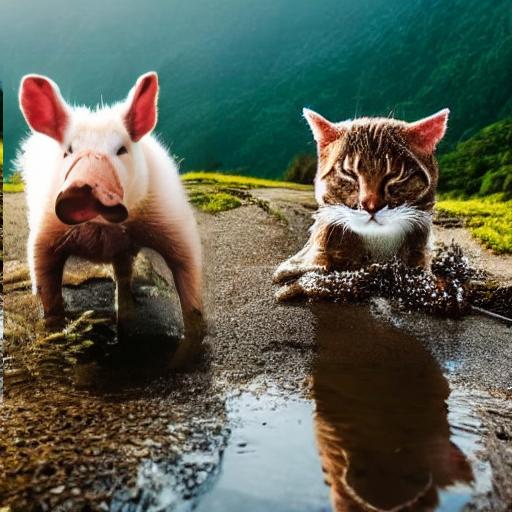}} &
        {\includegraphics[valign=c, width=\ww]{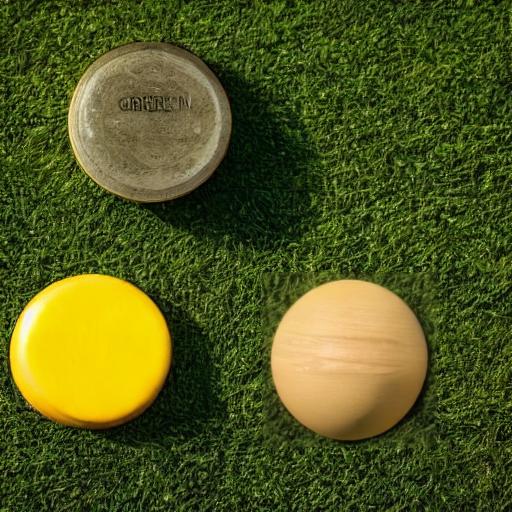}} &
        {\includegraphics[valign=c, width=\ww]{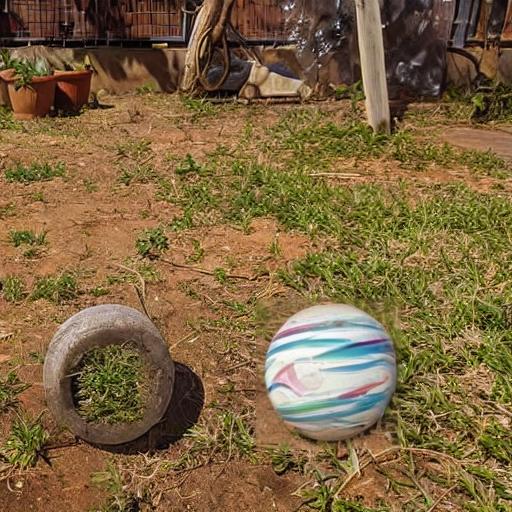}} &
        {\includegraphics[valign=c, width=\ww]{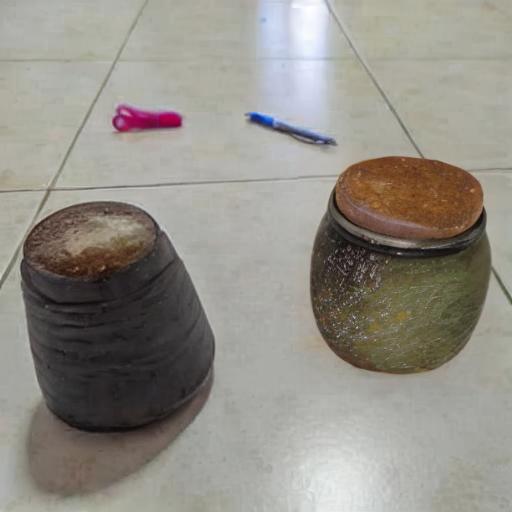}} &
        {\includegraphics[valign=c, width=\ww]{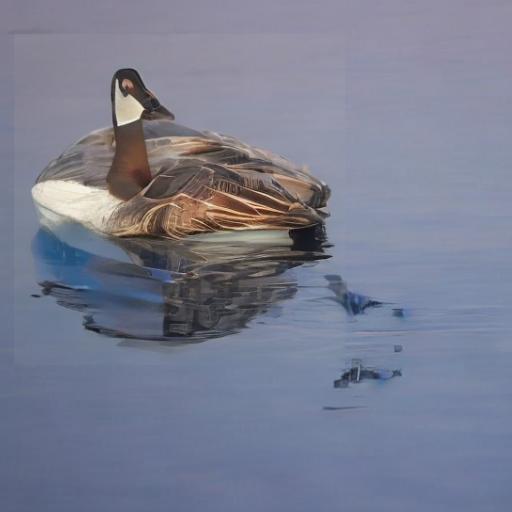}}
        \vspace{1px}
        \\

        \rotatebox[origin=c]{90}{{DragDiffusion}}
        {\includegraphics[valign=c, width=\ww]{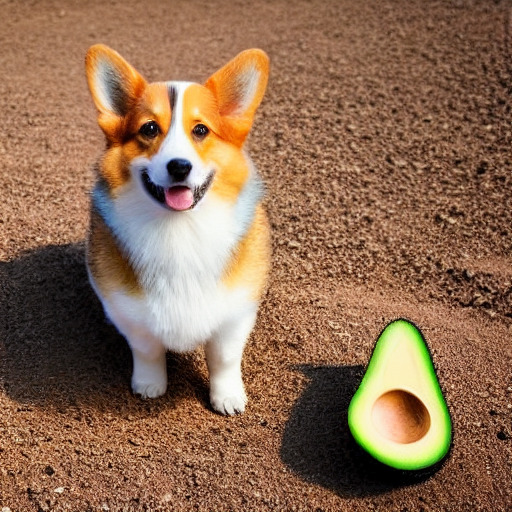}} &
        {\includegraphics[valign=c, width=\ww]{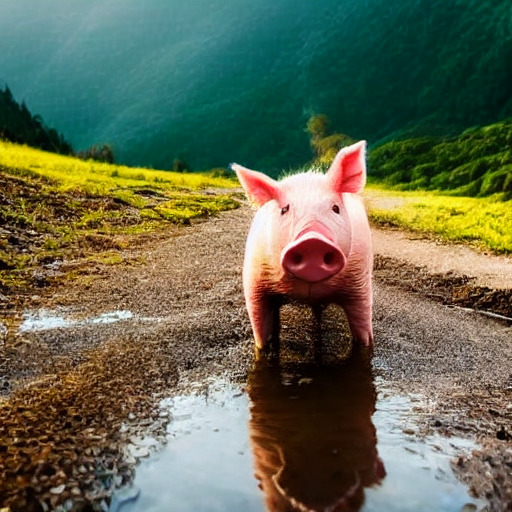}} &
        {\includegraphics[valign=c, width=\ww]{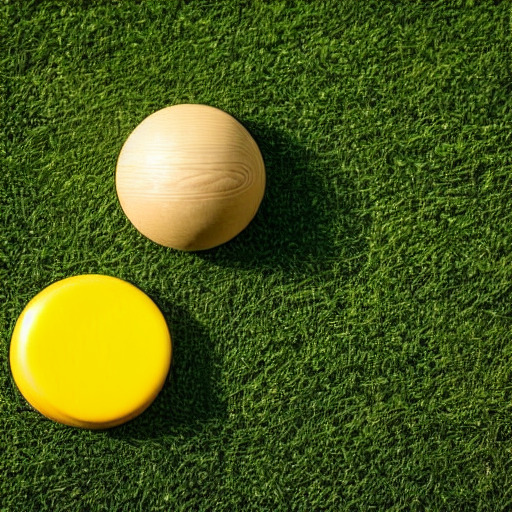}} &
        {\includegraphics[valign=c, width=\ww]{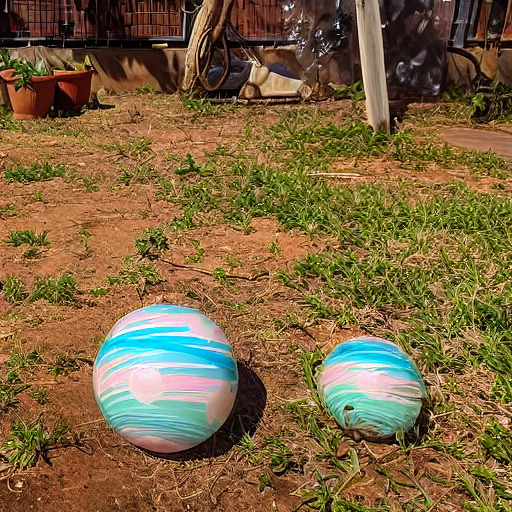}} &
        {\includegraphics[valign=c, width=\ww]{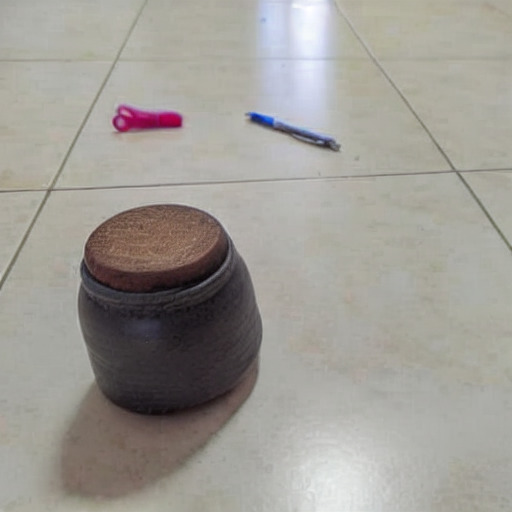}} &
        {\includegraphics[valign=c, width=\ww]{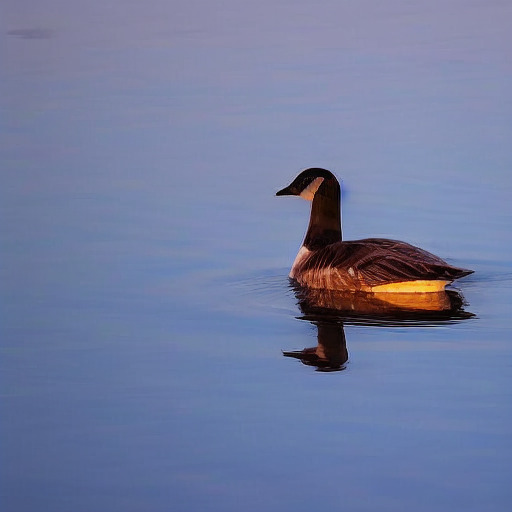}}
        \vspace{1px}
        \\

        \rotatebox[origin=c]{90}{{DragonDiffusion}}
        {\includegraphics[valign=c, width=\ww]{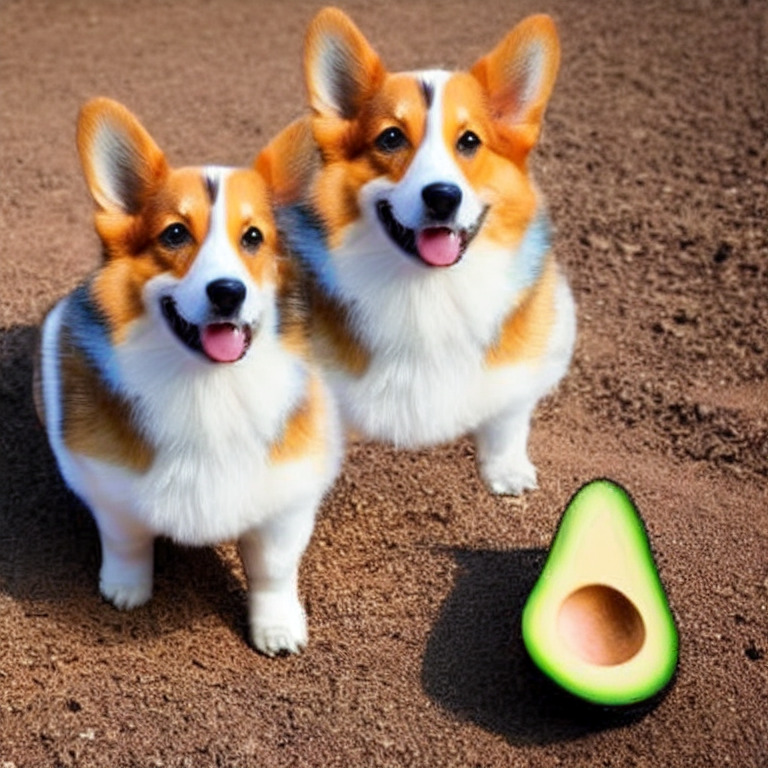}} &
        {\includegraphics[valign=c, width=\ww]{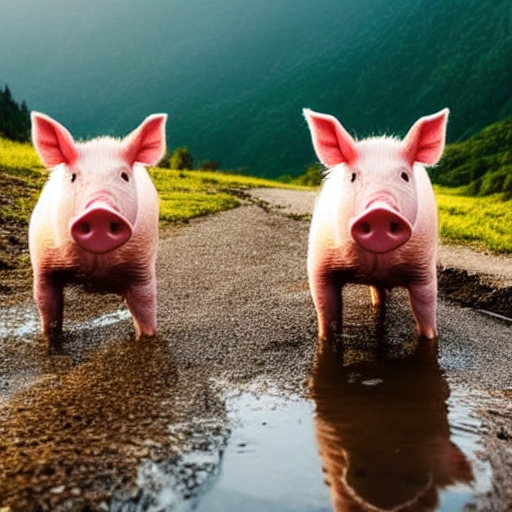}} &
        {\includegraphics[valign=c, width=\ww]{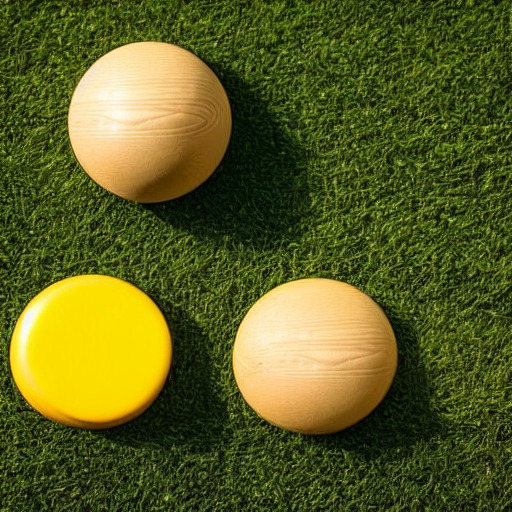}} &
        {\includegraphics[valign=c, width=\ww]{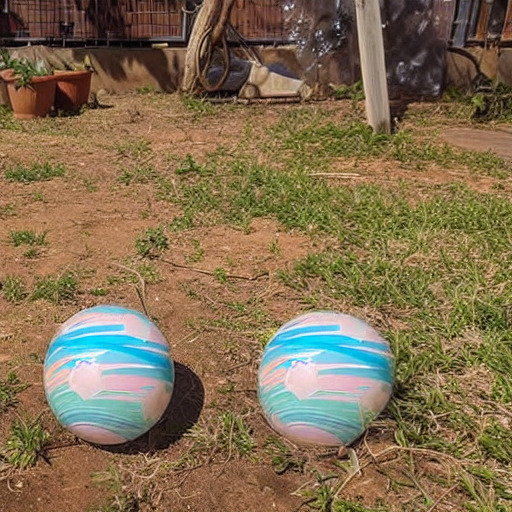}} &
        {\includegraphics[valign=c, width=\ww]{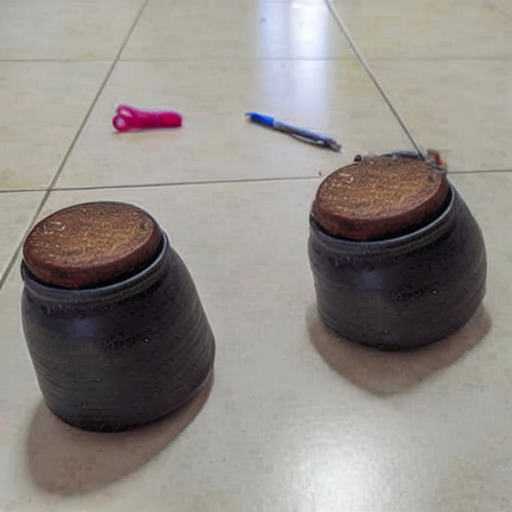}} &
        {\includegraphics[valign=c, width=\ww]{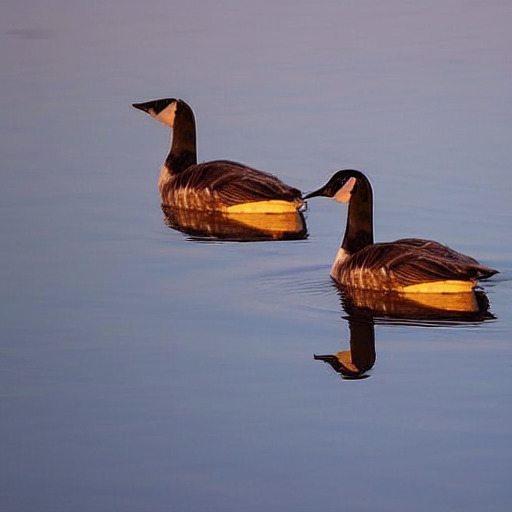}}
        \vspace{1px}
        \\

        \rotatebox[origin=c]{90}{{DiffEditor}}
        {\includegraphics[valign=c, width=\ww]{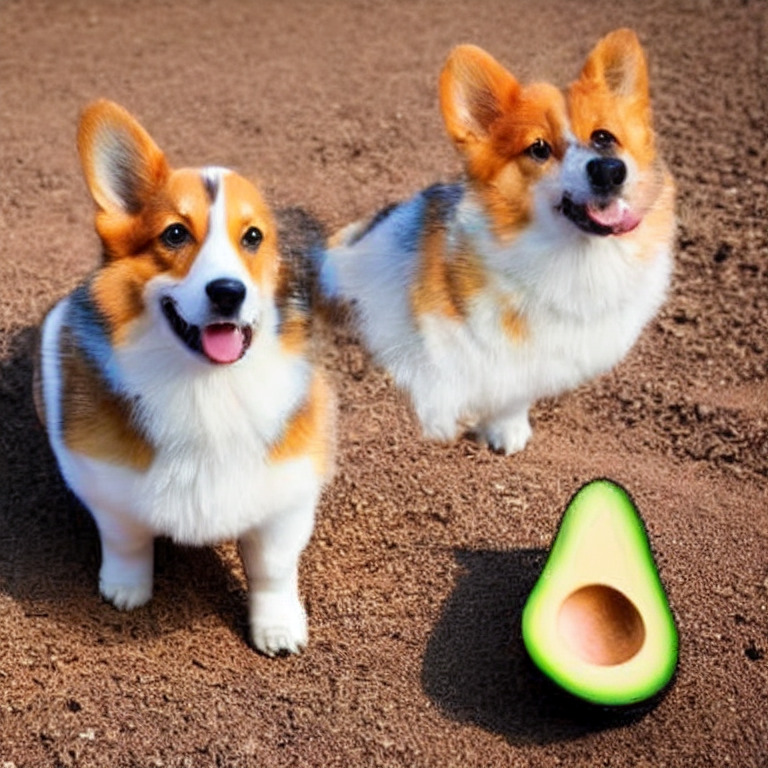}} &
        {\includegraphics[valign=c, width=\ww]{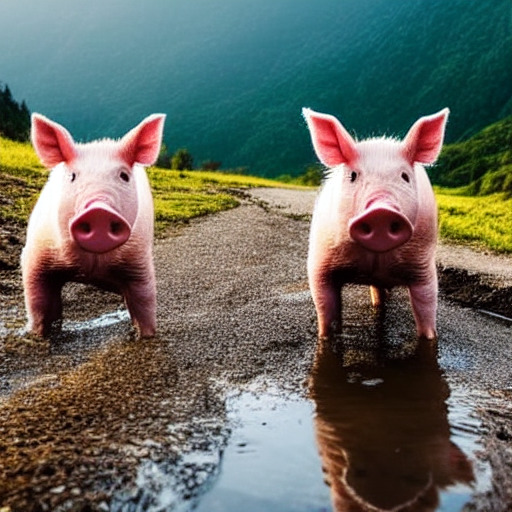}} &
        {\includegraphics[valign=c, width=\ww]{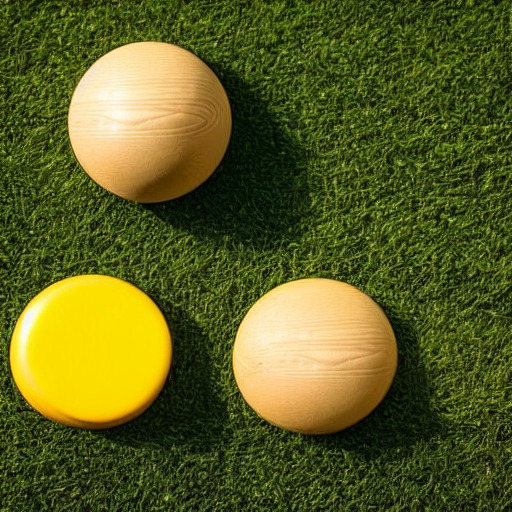}} &
        {\includegraphics[valign=c, width=\ww]{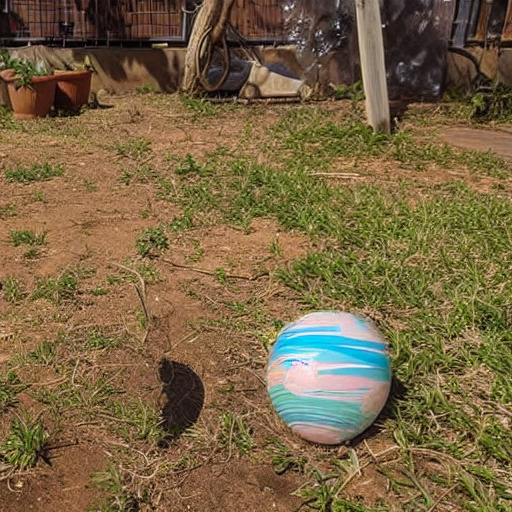}} &
        {\includegraphics[valign=c, width=\ww]{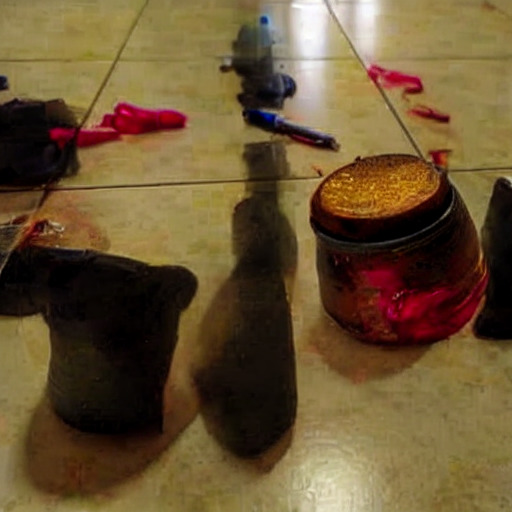}} &
        {\includegraphics[valign=c, width=\ww]{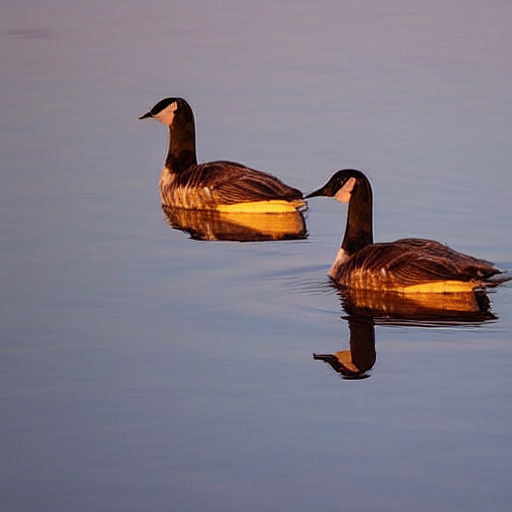}}
        \vspace{1px}
        \\

        \rotatebox[origin=c]{90}{{Ours}}
        {\includegraphics[valign=c, width=\ww]{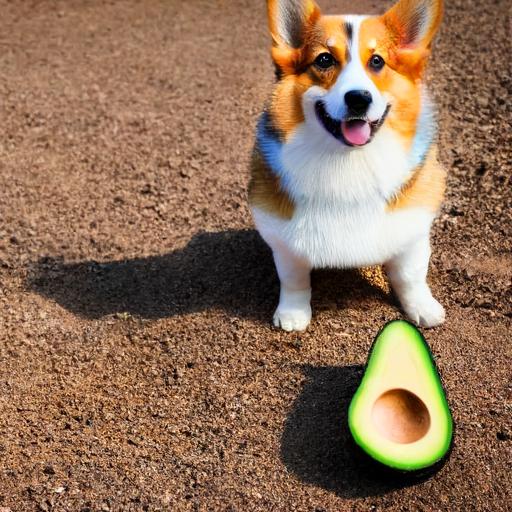}} &
        {\includegraphics[valign=c, width=\ww]{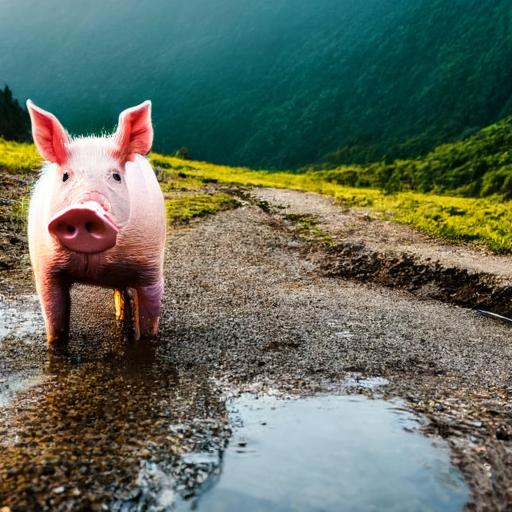}} &
        {\includegraphics[valign=c, width=\ww]{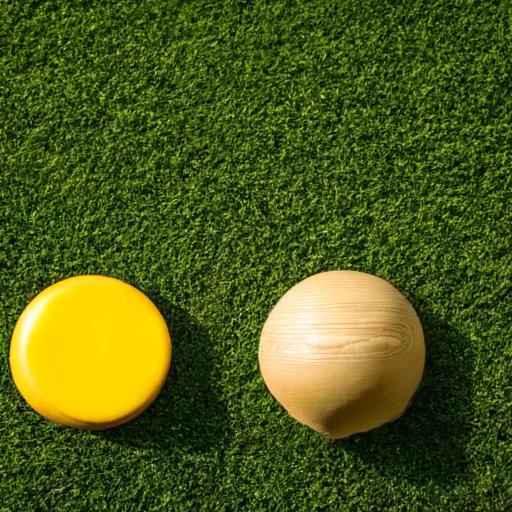}} &
        {\includegraphics[valign=c, width=\ww]{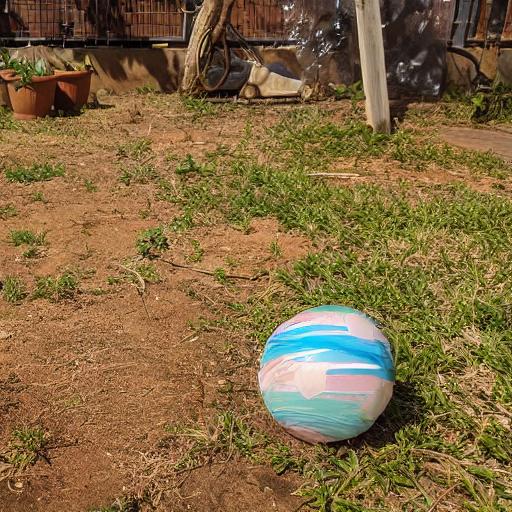}} &
        {\includegraphics[valign=c, width=\ww]{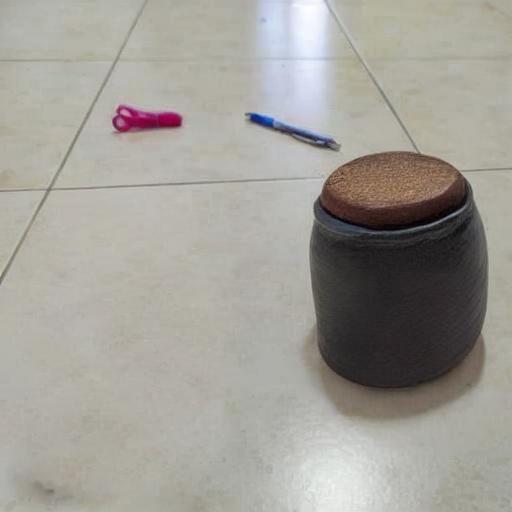}} &
        {\includegraphics[valign=c, width=\ww]{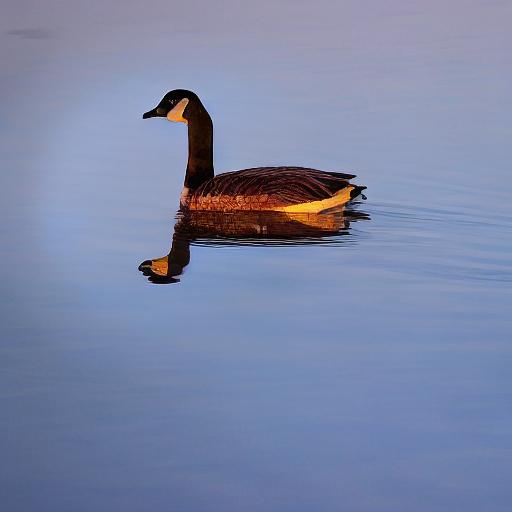}}
        \\

    \end{tabular}
    
    \caption{\textbf{Qualitative Comparison.} We compared our method against several baselines on both generated (first three columns) and real images (second three columns). The source and target locations are denoted by red and green points, respectively. As can be seen, PBE \cite{Yang2022PaintBE}, DiffusionSG \cite{self_guidance} and Anydoor \cite{Chen2023AnyDoorZO} mainly suffer from a bad preservation of the foreground object. DragDiffusion \cite{Shi2023DragDiffusionHD} struggles with dragging the object, while DragonDiffusion \cite{Mou2023DragonDiffusionED} and DiffEditor \cite{Mou2024DiffEditorBA} suffers from object traces. Our method, on the other hand, strikes the balance between dragging the object and preserving its identity.}
    \label{fig:qualitative_comparison}
\end{figure*}

\clearpage
\bibliographystyle{ACM-Reference-Format}
\bibliography{egbib}


\begin{thebibliography}{82}

%%% the Web address.

\ifx \showCODEN    \undefined \def \showCODEN     #1{\unskip}     \fi
\ifx \showDOI      \undefined \def \showDOI       #1{#1}\fi
\ifx \showISBNx    \undefined \def \showISBNx     #1{\unskip}     \fi
\ifx \showISBNxiii \undefined \def \showISBNxiii  #1{\unskip}     \fi
\ifx \showISSN     \undefined \def \showISSN      #1{\unskip}     \fi
\ifx \showLCCN     \undefined \def \showLCCN      #1{\unskip}     \fi
\ifx \shownote     \undefined \def \shownote      #1{#1}          \fi
\ifx \showarticletitle \undefined \def \showarticletitle #1{#1}   \fi
\ifx \showURL      \undefined \def \showURL       {\relax}        \fi
% invisible to TeX
\providecommand\bibfield[2]{#2}
\providecommand\bibinfo[2]{#2}
\providecommand\natexlab[1]{#1}
\providecommand\showeprint[2][]{arXiv:#2}

\bibitem[Alaluf et~al\mbox{.}(2023)]%
        {Alaluf2023ANS}
\bibfield{author}{\bibinfo{person}{Yuval Alaluf}, \bibinfo{person}{Elad Richardson}, \bibinfo{person}{Gal Metzer}, {and} \bibinfo{person}{Daniel Cohen-Or}.} \bibinfo{year}{2023}\natexlab{}.
\newblock \showarticletitle{A Neural Space-Time Representation for Text-to-Image Personalization}.
\newblock \bibinfo{journal}{\emph{ArXiv}}  \bibinfo{volume}{abs/2305.15391} (\bibinfo{year}{2023}).
\newblock
\urldef\tempurl%
\url{https://api.semanticscholar.org/CorpusID:258866047}
\showURL{%
\tempurl}


\bibitem[Alzayer et~al\mbox{.}(2024)]%
        {alzayer2024magic}
\bibfield{author}{\bibinfo{person}{Hadi Alzayer}, \bibinfo{person}{Zhihao Xia}, \bibinfo{person}{Xuaner Zhang}, \bibinfo{person}{Eli Shechtman}, \bibinfo{person}{Jia-Bin Huang}, {and} \bibinfo{person}{Michael Gharbi}.} \bibinfo{year}{2024}\natexlab{}.
\newblock \showarticletitle{Magic Fixup: Streamlining Photo Editing by Watching Dynamic Videos}.
\newblock \bibinfo{journal}{\emph{arXiv preprint arXiv:2403.13044}} (\bibinfo{year}{2024}).
\newblock


\bibitem[Amazon(2024)]%
        {amt}
\bibfield{author}{\bibinfo{person}{Amazon}.} \bibinfo{year}{2024}\natexlab{}.
\newblock \bibinfo{title}{Amazon Mechanical Turk}.
\newblock \bibinfo{howpublished}{\url{https://www.mturk.com/}}.
\newblock


\bibitem[Arar et~al\mbox{.}(2023)]%
        {arar2023domain}
\bibfield{author}{\bibinfo{person}{Moab Arar}, \bibinfo{person}{Rinon Gal}, \bibinfo{person}{Yuval Atzmon}, \bibinfo{person}{Gal Chechik}, \bibinfo{person}{Daniel Cohen-Or}, \bibinfo{person}{Ariel Shamir}, {and} \bibinfo{person}{Amit~H Bermano}.} \bibinfo{year}{2023}\natexlab{}.
\newblock \showarticletitle{Domain-agnostic tuning-encoder for fast personalization of text-to-image models}.
\newblock \bibinfo{journal}{\emph{arXiv preprint arXiv:2307.06925}} (\bibinfo{year}{2023}).
\newblock


\bibitem[Arar et~al\mbox{.}(2024)]%
        {arar2024Palp}
\bibfield{author}{\bibinfo{person}{Moab Arar}, \bibinfo{person}{Andrey Voynov}, \bibinfo{person}{Amir Hertz}, \bibinfo{person}{Omri Avrahami}, \bibinfo{person}{Shlomi Fruchter}, \bibinfo{person}{Yael Pritch}, \bibinfo{person}{Daniel Cohen-Or}, {and} \bibinfo{person}{Ariel Shamir}.} \bibinfo{year}{2024}\natexlab{}.
\newblock \showarticletitle{PALP: Prompt Aligned Personalization of Text-to-Image Models}.
\newblock  (\bibinfo{year}{2024}).
\newblock


\bibitem[Avrahami et~al\mbox{.}(2023a)]%
        {Avrahami2023BreakASceneEM}
\bibfield{author}{\bibinfo{person}{Omri Avrahami}, \bibinfo{person}{Kfir Aberman}, \bibinfo{person}{Ohad Fried}, \bibinfo{person}{Daniel Cohen-Or}, {and} \bibinfo{person}{Dani Lischinski}.} \bibinfo{year}{2023}\natexlab{a}.
\newblock \showarticletitle{Break-A-Scene: Extracting Multiple Concepts from a Single Image}.
\newblock \bibinfo{journal}{\emph{ArXiv}}  \bibinfo{volume}{abs/2305.16311} (\bibinfo{year}{2023}).
\newblock
\urldef\tempurl%
\url{https://api.semanticscholar.org/CorpusID:258888228}
\showURL{%
\tempurl}


\bibitem[Avrahami et~al\mbox{.}(2023b)]%
        {avrahami2023blendedlatent}
\bibfield{author}{\bibinfo{person}{Omri Avrahami}, \bibinfo{person}{Ohad Fried}, {and} \bibinfo{person}{Dani Lischinski}.} \bibinfo{year}{2023}\natexlab{b}.
\newblock \showarticletitle{Blended Latent Diffusion}.
\newblock \bibinfo{journal}{\emph{ACM Trans. Graph.}} \bibinfo{volume}{42}, \bibinfo{number}{4}, Article \bibinfo{articleno}{149} (\bibinfo{date}{jul} \bibinfo{year}{2023}), \bibinfo{numpages}{11}~pages.
\newblock
\showISSN{0730-0301}
\urldef\tempurl%
\url{https://doi.org/10.1145/3592450}
\showDOI{\tempurl}


\bibitem[Avrahami et~al\mbox{.}(2023c)]%
        {spatext_2023_CVPR}
\bibfield{author}{\bibinfo{person}{Omri Avrahami}, \bibinfo{person}{Thomas Hayes}, \bibinfo{person}{Oran Gafni}, \bibinfo{person}{Sonal Gupta}, \bibinfo{person}{Yaniv Taigman}, \bibinfo{person}{Devi Parikh}, \bibinfo{person}{Dani Lischinski}, \bibinfo{person}{Ohad Fried}, {and} \bibinfo{person}{Xi Yin}.} \bibinfo{year}{2023}\natexlab{c}.
\newblock \showarticletitle{SpaText: Spatio-Textual Representation for Controllable Image Generation}. In \bibinfo{booktitle}{\emph{Proceedings of the IEEE/CVF Conference on Computer Vision and Pattern Recognition (CVPR)}}. \bibinfo{pages}{18370--18380}.
\newblock


\bibitem[Avrahami et~al\mbox{.}(2023d)]%
        {Avrahami2023TheCO}
\bibfield{author}{\bibinfo{person}{Omri Avrahami}, \bibinfo{person}{Amir Hertz}, \bibinfo{person}{Yael Vinker}, \bibinfo{person}{Moab Arar}, \bibinfo{person}{Shlomi Fruchter}, \bibinfo{person}{Ohad Fried}, \bibinfo{person}{Daniel Cohen-Or}, {and} \bibinfo{person}{Dani Lischinski}.} \bibinfo{year}{2023}\natexlab{d}.
\newblock \showarticletitle{The Chosen One: Consistent Characters in Text-to-Image Diffusion Models}.
\newblock \bibinfo{journal}{\emph{ArXiv}}  \bibinfo{volume}{abs/2311.10093} (\bibinfo{year}{2023}).
\newblock
\urldef\tempurl%
\url{https://api.semanticscholar.org/CorpusID:265221238}
\showURL{%
\tempurl}


\bibitem[Avrahami et~al\mbox{.}(2022)]%
        {blended_2022_CVPR}
\bibfield{author}{\bibinfo{person}{Omri Avrahami}, \bibinfo{person}{Dani Lischinski}, {and} \bibinfo{person}{Ohad Fried}.} \bibinfo{year}{2022}\natexlab{}.
\newblock \showarticletitle{Blended Diffusion for Text-Driven Editing of Natural Images}. In \bibinfo{booktitle}{\emph{Proceedings of the IEEE/CVF Conference on Computer Vision and Pattern Recognition (CVPR)}}. \bibinfo{pages}{18208--18218}.
\newblock


\bibitem[Bar-Tal et~al\mbox{.}(2023)]%
        {BarTal2023MultiDiffusionFD}
\bibfield{author}{\bibinfo{person}{Omer Bar-Tal}, \bibinfo{person}{Lior Yariv}, \bibinfo{person}{Yaron Lipman}, {and} \bibinfo{person}{Tali Dekel}.} \bibinfo{year}{2023}\natexlab{}.
\newblock \showarticletitle{Multidiffusion: Fusing diffusion paths for controlled image generation}.
\newblock  (\bibinfo{year}{2023}).
\newblock


\bibitem[Binkowski et~al\mbox{.}(2018)]%
        {Binkowski2018DemystifyingMG}
\bibfield{author}{\bibinfo{person}{Mikolaj Binkowski}, \bibinfo{person}{Danica~J. Sutherland}, \bibinfo{person}{Michal Arbel}, {and} \bibinfo{person}{Arthur Gretton}.} \bibinfo{year}{2018}\natexlab{}.
\newblock \showarticletitle{Demystifying MMD GANs}.
\newblock \bibinfo{journal}{\emph{ArXiv}}  \bibinfo{volume}{abs/1801.01401} (\bibinfo{year}{2018}).
\newblock
\urldef\tempurl%
\url{https://api.semanticscholar.org/CorpusID:3531856}
\showURL{%
\tempurl}


\bibitem[Brooks et~al\mbox{.}(2023)]%
        {brooks2022instructpix2pix}
\bibfield{author}{\bibinfo{person}{Tim Brooks}, \bibinfo{person}{Aleksander Holynski}, {and} \bibinfo{person}{Alexei~A. Efros}.} \bibinfo{year}{2023}\natexlab{}.
\newblock \showarticletitle{InstructPix2Pix: Learning to Follow Image Editing Instructions}. In \bibinfo{booktitle}{\emph{CVPR}}.
\newblock


\bibitem[Cao et~al\mbox{.}(2023)]%
        {cao2023masactrl}
\bibfield{author}{\bibinfo{person}{Mingdeng Cao}, \bibinfo{person}{Xintao Wang}, \bibinfo{person}{Zhongang Qi}, \bibinfo{person}{Ying Shan}, \bibinfo{person}{Xiaohu Qie}, {and} \bibinfo{person}{Yinqiang Zheng}.} \bibinfo{year}{2023}\natexlab{}.
\newblock \showarticletitle{{MasaCtrl:} Tuning-Free Mutual Self-Attention Control for Consistent Image Synthesis and Editing}. In \bibinfo{booktitle}{\emph{Proceedings of the IEEE/CVF International Conference on Computer Vision (ICCV)}}. \bibinfo{pages}{22560--22570}.
\newblock


\bibitem[Chefer et~al\mbox{.}(2023)]%
        {Chefer2023AttendandExciteAS}
\bibfield{author}{\bibinfo{person}{Hila Chefer}, \bibinfo{person}{Yuval Alaluf}, \bibinfo{person}{Yael Vinker}, \bibinfo{person}{Lior Wolf}, {and} \bibinfo{person}{Daniel Cohen-Or}.} \bibinfo{year}{2023}\natexlab{}.
\newblock \showarticletitle{Attend-and-Excite: Attention-Based Semantic Guidance for Text-to-Image Diffusion Models}.
\newblock \bibinfo{journal}{\emph{ACM Transactions on Graphics (TOG)}}  \bibinfo{volume}{42} (\bibinfo{year}{2023}), \bibinfo{pages}{1 -- 10}.
\newblock
\urldef\tempurl%
\url{https://api.semanticscholar.org/CorpusID:256416326}
\showURL{%
\tempurl}


\bibitem[Chen et~al\mbox{.}(2023b)]%
        {chen2023trainingfree}
\bibfield{author}{\bibinfo{person}{Minghao Chen}, \bibinfo{person}{Iro Laina}, {and} \bibinfo{person}{Andrea Vedaldi}.} \bibinfo{year}{2023}\natexlab{b}.
\newblock \showarticletitle{Training-Free Layout Control with Cross-Attention Guidance}.
\newblock \bibinfo{journal}{\emph{arXiv preprint arXiv:2304.03373}} (\bibinfo{year}{2023}).
\newblock


\bibitem[Chen et~al\mbox{.}(2023a)]%
        {Chen2023AnyDoorZO}
\bibfield{author}{\bibinfo{person}{Xi Chen}, \bibinfo{person}{Lianghua Huang}, \bibinfo{person}{Yu Liu}, \bibinfo{person}{Yujun Shen}, \bibinfo{person}{Deli Zhao}, {and} \bibinfo{person}{Hengshuang Zhao}.} \bibinfo{year}{2023}\natexlab{a}.
\newblock \showarticletitle{AnyDoor: Zero-shot Object-level Image Customization}.
\newblock \bibinfo{journal}{\emph{ArXiv}}  \bibinfo{volume}{abs/2307.09481} (\bibinfo{year}{2023}).
\newblock
\urldef\tempurl%
\url{https://api.semanticscholar.org/CorpusID:259951373}
\showURL{%
\tempurl}


\bibitem[Dhariwal and Nichol(2021)]%
        {Dhariwal2021DiffusionMB}
\bibfield{author}{\bibinfo{person}{Prafulla Dhariwal} {and} \bibinfo{person}{Alex Nichol}.} \bibinfo{year}{2021}\natexlab{}.
\newblock \showarticletitle{Diffusion Models Beat GANs on Image Synthesis}.
\newblock \bibinfo{journal}{\emph{ArXiv}}  \bibinfo{volume}{abs/2105.05233} (\bibinfo{year}{2021}).
\newblock
\urldef\tempurl%
\url{https://api.semanticscholar.org/CorpusID:234357997}
\showURL{%
\tempurl}


\bibitem[Endo(2022)]%
        {Endo2022UserControllableLT}
\bibfield{author}{\bibinfo{person}{Yuki Endo}.} \bibinfo{year}{2022}\natexlab{}.
\newblock \showarticletitle{User‐Controllable Latent Transformer for StyleGAN Image Layout Editing}.
\newblock \bibinfo{journal}{\emph{Computer Graphics Forum}}  \bibinfo{volume}{41} (\bibinfo{year}{2022}).
\newblock
\urldef\tempurl%
\url{https://api.semanticscholar.org/CorpusID:251881740}
\showURL{%
\tempurl}


\bibitem[Epstein et~al\mbox{.}(2023)]%
        {self_guidance}
\bibfield{author}{\bibinfo{person}{Dave Epstein}, \bibinfo{person}{Allan Jabri}, \bibinfo{person}{Ben Poole}, \bibinfo{person}{Alexei Efros}, {and} \bibinfo{person}{Aleksander Holynski}.} \bibinfo{year}{2023}\natexlab{}.
\newblock \showarticletitle{Diffusion self-guidance for controllable image generation}.
\newblock \bibinfo{journal}{\emph{Advances in Neural Information Processing Systems}}  \bibinfo{volume}{36} (\bibinfo{year}{2023}), \bibinfo{pages}{16222--16239}.
\newblock


\bibitem[Epstein et~al\mbox{.}(2022)]%
        {Epstein2022BlobGANSD}
\bibfield{author}{\bibinfo{person}{Dave Epstein}, \bibinfo{person}{Taesung Park}, \bibinfo{person}{Richard Zhang}, \bibinfo{person}{Eli Shechtman}, {and} \bibinfo{person}{Alexei~A. Efros}.} \bibinfo{year}{2022}\natexlab{}.
\newblock \showarticletitle{BlobGAN: Spatially Disentangled Scene Representations}.
\newblock \bibinfo{journal}{\emph{ArXiv}}  \bibinfo{volume}{abs/2205.02837} (\bibinfo{year}{2022}).
\newblock
\urldef\tempurl%
\url{https://api.semanticscholar.org/CorpusID:248524853}
\showURL{%
\tempurl}


\bibitem[Feng et~al\mbox{.}(2022)]%
        {feng2022training}
\bibfield{author}{\bibinfo{person}{Weixi Feng}, \bibinfo{person}{Xuehai He}, \bibinfo{person}{Tsu-Jui Fu}, \bibinfo{person}{Varun Jampani}, \bibinfo{person}{Arjun~Reddy Akula}, \bibinfo{person}{Pradyumna Narayana}, \bibinfo{person}{Sugato Basu}, \bibinfo{person}{Xin~Eric Wang}, {and} \bibinfo{person}{William~Yang Wang}.} \bibinfo{year}{2022}\natexlab{}.
\newblock \showarticletitle{Training-Free Structured Diffusion Guidance for Compositional Text-to-Image Synthesis}. In \bibinfo{booktitle}{\emph{The Eleventh International Conference on Learning Representations}}.
\newblock


\bibitem[Frenkel et~al\mbox{.}(2024)]%
        {Frenkel2024ImplicitSS}
\bibfield{author}{\bibinfo{person}{Yarden Frenkel}, \bibinfo{person}{Yael Vinker}, \bibinfo{person}{Ariel Shamir}, {and} \bibinfo{person}{Daniel Cohen-Or}.} \bibinfo{year}{2024}\natexlab{}.
\newblock \showarticletitle{Implicit Style-Content Separation using B-LoRA}.
\newblock \bibinfo{journal}{\emph{ArXiv}}  \bibinfo{volume}{abs/2403.14572} (\bibinfo{year}{2024}).
\newblock
\urldef\tempurl%
\url{https://api.semanticscholar.org/CorpusID:268553753}
\showURL{%
\tempurl}


\bibitem[Gal et~al\mbox{.}(2022)]%
        {Gal2022AnII}
\bibfield{author}{\bibinfo{person}{Rinon Gal}, \bibinfo{person}{Yuval Alaluf}, \bibinfo{person}{Yuval Atzmon}, \bibinfo{person}{Or Patashnik}, \bibinfo{person}{Amit~Haim Bermano}, \bibinfo{person}{Gal Chechik}, {and} \bibinfo{person}{Daniel Cohen-or}.} \bibinfo{year}{2022}\natexlab{}.
\newblock \showarticletitle{An Image is Worth One Word: Personalizing Text-to-Image Generation using Textual Inversion}. In \bibinfo{booktitle}{\emph{The Eleventh International Conference on Learning Representations}}.
\newblock


\bibitem[Gal et~al\mbox{.}(2023)]%
        {gal2023encoder}
\bibfield{author}{\bibinfo{person}{Rinon Gal}, \bibinfo{person}{Moab Arar}, \bibinfo{person}{Yuval Atzmon}, \bibinfo{person}{Amit~H Bermano}, \bibinfo{person}{Gal Chechik}, {and} \bibinfo{person}{Daniel Cohen-Or}.} \bibinfo{year}{2023}\natexlab{}.
\newblock \showarticletitle{Encoder-based domain tuning for fast personalization of text-to-image models}.
\newblock \bibinfo{journal}{\emph{ACM Transactions on Graphics (TOG)}} \bibinfo{volume}{42}, \bibinfo{number}{4} (\bibinfo{year}{2023}), \bibinfo{pages}{1--13}.
\newblock


\bibitem[Geyer et~al\mbox{.}(2023)]%
        {geyer2023tokenflow}
\bibfield{author}{\bibinfo{person}{Michal Geyer}, \bibinfo{person}{Omer Bar-Tal}, \bibinfo{person}{Shai Bagon}, {and} \bibinfo{person}{Tali Dekel}.} \bibinfo{year}{2023}\natexlab{}.
\newblock \showarticletitle{Tokenflow: Consistent diffusion features for consistent video editing}.
\newblock \bibinfo{journal}{\emph{arXiv preprint arXiv:2307.10373}} (\bibinfo{year}{2023}).
\newblock


\bibitem[Goodfellow et~al\mbox{.}(2014)]%
        {goodfellow2014generative}
\bibfield{author}{\bibinfo{person}{Ian Goodfellow}, \bibinfo{person}{Jean Pouget-Abadie}, \bibinfo{person}{Mehdi Mirza}, \bibinfo{person}{Bing Xu}, \bibinfo{person}{David Warde-Farley}, \bibinfo{person}{Sherjil Ozair}, \bibinfo{person}{Aaron Courville}, {and} \bibinfo{person}{Yoshua Bengio}.} \bibinfo{year}{2014}\natexlab{}.
\newblock \showarticletitle{Generative adversarial nets}.
\newblock \bibinfo{journal}{\emph{Advances in neural information processing systems}}  \bibinfo{volume}{27} (\bibinfo{year}{2014}).
\newblock


\bibitem[Hertz et~al\mbox{.}(2023)]%
        {hertz2023delta}
\bibfield{author}{\bibinfo{person}{Amir Hertz}, \bibinfo{person}{Kfir Aberman}, {and} \bibinfo{person}{Daniel Cohen-Or}.} \bibinfo{year}{2023}\natexlab{}.
\newblock \showarticletitle{Delta denoising score}. In \bibinfo{booktitle}{\emph{Proceedings of the IEEE/CVF International Conference on Computer Vision}}. \bibinfo{pages}{2328--2337}.
\newblock


\bibitem[Hertz et~al\mbox{.}(2022)]%
        {hertz2022prompt}
\bibfield{author}{\bibinfo{person}{Amir Hertz}, \bibinfo{person}{Ron Mokady}, \bibinfo{person}{Jay Tenenbaum}, \bibinfo{person}{Kfir Aberman}, \bibinfo{person}{Yael Pritch}, {and} \bibinfo{person}{Daniel Cohen-Or}.} \bibinfo{year}{2022}\natexlab{}.
\newblock \showarticletitle{Prompt-to-prompt image editing with cross attention control}.
\newblock \bibinfo{journal}{\emph{arXiv preprint arXiv:2208.01626}} (\bibinfo{year}{2022}).
\newblock


\bibitem[Heusel et~al\mbox{.}(2017)]%
        {heusel2017gans}
\bibfield{author}{\bibinfo{person}{Martin Heusel}, \bibinfo{person}{Hubert Ramsauer}, \bibinfo{person}{Thomas Unterthiner}, \bibinfo{person}{Bernhard Nessler}, {and} \bibinfo{person}{Sepp Hochreiter}.} \bibinfo{year}{2017}\natexlab{}.
\newblock \showarticletitle{Gans trained by a two time-scale update rule converge to a local nash equilibrium}.
\newblock \bibinfo{journal}{\emph{Advances in neural information processing systems}}  \bibinfo{volume}{30} (\bibinfo{year}{2017}).
\newblock


\bibitem[Ho(2022)]%
        {Ho2022ClassifierFreeDG}
\bibfield{author}{\bibinfo{person}{Jonathan Ho}.} \bibinfo{year}{2022}\natexlab{}.
\newblock \showarticletitle{Classifier-Free Diffusion Guidance}.
\newblock \bibinfo{journal}{\emph{ArXiv}}  \bibinfo{volume}{abs/2207.12598} (\bibinfo{year}{2022}).
\newblock
\urldef\tempurl%
\url{https://api.semanticscholar.org/CorpusID:249145348}
\showURL{%
\tempurl}


\bibitem[Ho et~al\mbox{.}(2020)]%
        {ho2020denoising}
\bibfield{author}{\bibinfo{person}{Jonathan Ho}, \bibinfo{person}{Ajay Jain}, {and} \bibinfo{person}{Pieter Abbeel}.} \bibinfo{year}{2020}\natexlab{}.
\newblock \showarticletitle{Denoising Diffusion Probabilistic Models}. In \bibinfo{booktitle}{\emph{Proc.~NeurIPS}}.
\newblock


\bibitem[Horwitz et~al\mbox{.}(2024)]%
        {Horwitz2024RecoveringTP}
\bibfield{author}{\bibinfo{person}{Eliahu Horwitz}, \bibinfo{person}{Jonathan Kahana}, {and} \bibinfo{person}{Yedid Hoshen}.} \bibinfo{year}{2024}\natexlab{}.
\newblock \showarticletitle{Recovering the Pre-Fine-Tuning Weights of Generative Models}.
\newblock \bibinfo{journal}{\emph{ArXiv}}  \bibinfo{volume}{abs/2402.10208} (\bibinfo{year}{2024}).
\newblock
\urldef\tempurl%
\url{https://api.semanticscholar.org/CorpusID:267682124}
\showURL{%
\tempurl}


\bibitem[Hou et~al\mbox{.}(2024)]%
        {hou2024easydrag}
\bibfield{author}{\bibinfo{person}{Xingzhong Hou}, \bibinfo{person}{Boxiao Liu}, \bibinfo{person}{Yi Zhang}, \bibinfo{person}{Jihao Liu}, \bibinfo{person}{Yu Liu}, {and} \bibinfo{person}{Haihang You}.} \bibinfo{year}{2024}\natexlab{}.
\newblock \showarticletitle{EasyDrag: Efficient Point-based Manipulation on Diffusion Models}. In \bibinfo{booktitle}{\emph{Proceedings of the IEEE/CVF Conference on Computer Vision and Pattern Recognition}}. \bibinfo{pages}{8404--8413}.
\newblock


\bibitem[Hu et~al\mbox{.}(2021)]%
        {lora}
\bibfield{author}{\bibinfo{person}{Edward~J Hu}, \bibinfo{person}{Phillip Wallis}, \bibinfo{person}{Zeyuan Allen-Zhu}, \bibinfo{person}{Yuanzhi Li}, \bibinfo{person}{Shean Wang}, \bibinfo{person}{Lu Wang}, \bibinfo{person}{Weizhu Chen}, {et~al\mbox{.}}} \bibinfo{year}{2021}\natexlab{}.
\newblock \showarticletitle{{LoRA}: Low-Rank Adaptation of Large Language Models}. In \bibinfo{booktitle}{\emph{International Conference on Learning Representations}}.
\newblock


\bibitem[Huberman-Spiegelglas et~al\mbox{.}(2023)]%
        {huberman2023edit}
\bibfield{author}{\bibinfo{person}{Inbar Huberman-Spiegelglas}, \bibinfo{person}{Vladimir Kulikov}, {and} \bibinfo{person}{Tomer Michaeli}.} \bibinfo{year}{2023}\natexlab{}.
\newblock \showarticletitle{An Edit Friendly DDPM Noise Space: Inversion and Manipulations}.
\newblock \bibinfo{journal}{\emph{arXiv e-prints}} (\bibinfo{year}{2023}), \bibinfo{pages}{arXiv--2304}.
\newblock


\bibitem[Karras et~al\mbox{.}(2021)]%
        {karras2021aliasfree}
\bibfield{author}{\bibinfo{person}{Tero Karras}, \bibinfo{person}{Miika Aittala}, \bibinfo{person}{Samuli Laine}, \bibinfo{person}{Erik Härkönen}, \bibinfo{person}{Janne Hellsten}, \bibinfo{person}{Jaakko Lehtinen}, {and} \bibinfo{person}{Timo Aila}.} \bibinfo{year}{2021}\natexlab{}.
\newblock \bibinfo{title}{Alias-Free Generative Adversarial Networks}.
\newblock
\newblock
\showeprint[arxiv]{2106.12423}~[cs.CV]


\bibitem[Karras et~al\mbox{.}(2019)]%
        {karras2019style}
\bibfield{author}{\bibinfo{person}{Tero Karras}, \bibinfo{person}{Samuli Laine}, {and} \bibinfo{person}{Timo Aila}.} \bibinfo{year}{2019}\natexlab{}.
\newblock \showarticletitle{A style-based generator architecture for generative adversarial networks}. In \bibinfo{booktitle}{\emph{Proceedings of the IEEE conference on computer vision and pattern recognition}}. \bibinfo{pages}{4401--4410}.
\newblock


\bibitem[Karras et~al\mbox{.}(2020)]%
        {karras2020analyzing}
\bibfield{author}{\bibinfo{person}{Tero Karras}, \bibinfo{person}{Samuli Laine}, \bibinfo{person}{Miika Aittala}, \bibinfo{person}{Janne Hellsten}, \bibinfo{person}{Jaakko Lehtinen}, {and} \bibinfo{person}{Timo Aila}.} \bibinfo{year}{2020}\natexlab{}.
\newblock \showarticletitle{Analyzing and improving the image quality of stylegan}. In \bibinfo{booktitle}{\emph{Proceedings of the IEEE/CVF Conference on Computer Vision and Pattern Recognition}}. \bibinfo{pages}{8110--8119}.
\newblock


\bibitem[Kawar et~al\mbox{.}(2023)]%
        {Kawar2022ImagicTR}
\bibfield{author}{\bibinfo{person}{Bahjat Kawar}, \bibinfo{person}{Shiran Zada}, \bibinfo{person}{Oran Lang}, \bibinfo{person}{Omer Tov}, \bibinfo{person}{Huiwen Chang}, \bibinfo{person}{Tali Dekel}, \bibinfo{person}{Inbar Mosseri}, {and} \bibinfo{person}{Michal Irani}.} \bibinfo{year}{2023}\natexlab{}.
\newblock \showarticletitle{Imagic: Text-based real image editing with diffusion models}. In \bibinfo{booktitle}{\emph{Proceedings of the IEEE/CVF Conference on Computer Vision and Pattern Recognition}}. \bibinfo{pages}{6007--6017}.
\newblock


\bibitem[Li et~al\mbox{.}(2023)]%
        {Li2023GLIGENOG}
\bibfield{author}{\bibinfo{person}{Yuheng Li}, \bibinfo{person}{Haotian Liu}, \bibinfo{person}{Qingyang Wu}, \bibinfo{person}{Fangzhou Mu}, \bibinfo{person}{Jianwei Yang}, \bibinfo{person}{Jianfeng Gao}, \bibinfo{person}{Chunyuan Li}, {and} \bibinfo{person}{Yong~Jae Lee}.} \bibinfo{year}{2023}\natexlab{}.
\newblock \showarticletitle{GLIGEN: Open-Set Grounded Text-to-Image Generation}.
\newblock \bibinfo{journal}{\emph{2023 IEEE/CVF Conference on Computer Vision and Pattern Recognition (CVPR)}} (\bibinfo{year}{2023}), \bibinfo{pages}{22511--22521}.
\newblock
\urldef\tempurl%
\url{https://api.semanticscholar.org/CorpusID:255942528}
\showURL{%
\tempurl}


\bibitem[Lin et~al\mbox{.}(2014)]%
        {Lin2014MicrosoftCC}
\bibfield{author}{\bibinfo{person}{Tsung-Yi Lin}, \bibinfo{person}{Michael Maire}, \bibinfo{person}{Serge~J. Belongie}, \bibinfo{person}{James Hays}, \bibinfo{person}{Pietro Perona}, \bibinfo{person}{Deva Ramanan}, \bibinfo{person}{Piotr Doll{\'a}r}, {and} \bibinfo{person}{C.~Lawrence Zitnick}.} \bibinfo{year}{2014}\natexlab{}.
\newblock \showarticletitle{Microsoft COCO: Common Objects in Context}. In \bibinfo{booktitle}{\emph{European Conference on Computer Vision}}.
\newblock


\bibitem[Liu et~al\mbox{.}(2023)]%
        {Liu2023ImprovedBW}
\bibfield{author}{\bibinfo{person}{Haotian Liu}, \bibinfo{person}{Chunyuan Li}, \bibinfo{person}{Yuheng Li}, {and} \bibinfo{person}{Yong~Jae Lee}.} \bibinfo{year}{2023}\natexlab{}.
\newblock \showarticletitle{Improved Baselines with Visual Instruction Tuning}.
\newblock \bibinfo{journal}{\emph{ArXiv}}  \bibinfo{volume}{abs/2310.03744} (\bibinfo{year}{2023}).
\newblock
\urldef\tempurl%
\url{https://api.semanticscholar.org/CorpusID:263672058}
\showURL{%
\tempurl}


\bibitem[Meng et~al\mbox{.}(2021)]%
        {meng2021sdedit}
\bibfield{author}{\bibinfo{person}{Chenlin Meng}, \bibinfo{person}{Yutong He}, \bibinfo{person}{Yang Song}, \bibinfo{person}{Jiaming Song}, \bibinfo{person}{Jiajun Wu}, \bibinfo{person}{Jun-Yan Zhu}, {and} \bibinfo{person}{Stefano Ermon}.} \bibinfo{year}{2021}\natexlab{}.
\newblock \showarticletitle{SDEdit: Guided Image Synthesis and Editing with Stochastic Differential Equations}. In \bibinfo{booktitle}{\emph{International Conference on Learning Representations}}.
\newblock


\bibitem[Mokady et~al\mbox{.}(2023)]%
        {mokady2022null}
\bibfield{author}{\bibinfo{person}{Ron Mokady}, \bibinfo{person}{Amir Hertz}, \bibinfo{person}{Kfir Aberman}, \bibinfo{person}{Yael Pritch}, {and} \bibinfo{person}{Daniel Cohen-Or}.} \bibinfo{year}{2023}\natexlab{}.
\newblock \showarticletitle{Null-text inversion for editing real images using guided diffusion models}. In \bibinfo{booktitle}{\emph{Proceedings of the IEEE/CVF Conference on Computer Vision and Pattern Recognition}}. \bibinfo{pages}{6038--6047}.
\newblock


\bibitem[Molad et~al\mbox{.}(2023)]%
        {Molad2023DreamixVD}
\bibfield{author}{\bibinfo{person}{Eyal Molad}, \bibinfo{person}{Eliahu Horwitz}, \bibinfo{person}{Dani Valevski}, \bibinfo{person}{Alex~Rav Acha}, \bibinfo{person}{Y. Matias}, \bibinfo{person}{Yael Pritch}, \bibinfo{person}{Yaniv Leviathan}, {and} \bibinfo{person}{Yedid Hoshen}.} \bibinfo{year}{2023}\natexlab{}.
\newblock \showarticletitle{Dreamix: Video Diffusion Models are General Video Editors}.
\newblock \bibinfo{journal}{\emph{ArXiv}}  \bibinfo{volume}{abs/2302.01329} (\bibinfo{year}{2023}).
\newblock


\bibitem[Mou et~al\mbox{.}(2023)]%
        {Mou2023DragonDiffusionED}
\bibfield{author}{\bibinfo{person}{Chong Mou}, \bibinfo{person}{Xintao Wang}, \bibinfo{person}{Jie Song}, \bibinfo{person}{Ying Shan}, {and} \bibinfo{person}{Jian Zhang}.} \bibinfo{year}{2023}\natexlab{}.
\newblock \showarticletitle{DragonDiffusion: Enabling Drag-style Manipulation on Diffusion Models}.
\newblock \bibinfo{journal}{\emph{ArXiv}}  \bibinfo{volume}{abs/2307.02421} (\bibinfo{year}{2023}).
\newblock
\urldef\tempurl%
\url{https://api.semanticscholar.org/CorpusID:259342813}
\showURL{%
\tempurl}


\bibitem[Mou et~al\mbox{.}(2024)]%
        {Mou2024DiffEditorBA}
\bibfield{author}{\bibinfo{person}{Chong Mou}, \bibinfo{person}{Xintao Wang}, \bibinfo{person}{Jie Song}, \bibinfo{person}{Ying Shan}, {and} \bibinfo{person}{Jian Zhang}.} \bibinfo{year}{2024}\natexlab{}.
\newblock \showarticletitle{DiffEditor: Boosting Accuracy and Flexibility on Diffusion-based Image Editing}.
\newblock \bibinfo{journal}{\emph{ArXiv}}  \bibinfo{volume}{abs/2402.02583} (\bibinfo{year}{2024}).
\newblock
\urldef\tempurl%
\url{https://api.semanticscholar.org/CorpusID:267499649}
\showURL{%
\tempurl}


\bibitem[Nie et~al\mbox{.}(2024)]%
        {nie2024compositional}
\bibfield{author}{\bibinfo{person}{Weili Nie}, \bibinfo{person}{Sifei Liu}, \bibinfo{person}{Morteza Mardani}, \bibinfo{person}{Chao Liu}, \bibinfo{person}{Benjamin Eckart}, {and} \bibinfo{person}{Arash Vahdat}.} \bibinfo{year}{2024}\natexlab{}.
\newblock \bibinfo{title}{Compositional Text-to-Image Generation with Dense Blob Representations}.
\newblock
\newblock
\showeprint[arxiv]{2405.08246}~[cs.CV]


\bibitem[Oquab et~al\mbox{.}(2023)]%
        {Oquab2023DINOv2LR}
\bibfield{author}{\bibinfo{person}{Maxime Oquab}, \bibinfo{person}{Timoth{\'e}e Darcet}, \bibinfo{person}{Th{\'e}o Moutakanni}, \bibinfo{person}{Huy~Q. Vo}, \bibinfo{person}{Marc Szafraniec}, \bibinfo{person}{Vasil Khalidov}, \bibinfo{person}{Pierre Fernandez}, \bibinfo{person}{Daniel Haziza}, \bibinfo{person}{Francisco Massa}, \bibinfo{person}{Alaaeldin El-Nouby}, \bibinfo{person}{Mahmoud Assran}, \bibinfo{person}{Nicolas Ballas}, \bibinfo{person}{Wojciech Galuba}, \bibinfo{person}{Russ Howes}, \bibinfo{person}{Po-Yao~(Bernie) Huang}, \bibinfo{person}{Shang-Wen Li}, \bibinfo{person}{Ishan Misra}, \bibinfo{person}{Michael~G. Rabbat}, \bibinfo{person}{Vasu Sharma}, \bibinfo{person}{Gabriel Synnaeve}, \bibinfo{person}{Huijiao Xu}, \bibinfo{person}{Herv{\'e} J{\'e}gou}, \bibinfo{person}{Julien Mairal}, \bibinfo{person}{Patrick Labatut}, \bibinfo{person}{Armand Joulin}, {and} \bibinfo{person}{Piotr Bojanowski}.} \bibinfo{year}{2023}\natexlab{}.
\newblock \showarticletitle{{DINOv2}: Learning Robust Visual Features without Supervision}.
\newblock \bibinfo{journal}{\emph{ArXiv}}  \bibinfo{volume}{abs/2304.07193} (\bibinfo{year}{2023}).
\newblock
\urldef\tempurl%
\url{https://api.semanticscholar.org/CorpusID:258170077}
\showURL{%
\tempurl}


\bibitem[Pan et~al\mbox{.}(2023)]%
        {Pan2023DragYG}
\bibfield{author}{\bibinfo{person}{Xingang Pan}, \bibinfo{person}{Ayush~Kumar Tewari}, \bibinfo{person}{Thomas Leimk{\"u}hler}, \bibinfo{person}{Lingjie Liu}, \bibinfo{person}{Abhimitra Meka}, {and} \bibinfo{person}{Christian Theobalt}.} \bibinfo{year}{2023}\natexlab{}.
\newblock \showarticletitle{Drag Your GAN: Interactive Point-based Manipulation on the Generative Image Manifold}.
\newblock \bibinfo{journal}{\emph{ACM SIGGRAPH 2023 Conference Proceedings}} (\bibinfo{year}{2023}).
\newblock
\urldef\tempurl%
\url{https://api.semanticscholar.org/CorpusID:258762550}
\showURL{%
\tempurl}


\bibitem[Pandey et~al\mbox{.}(2023)]%
        {pandey2023diffusion}
\bibfield{author}{\bibinfo{person}{Karran Pandey}, \bibinfo{person}{Paul Guerrero}, \bibinfo{person}{Matheus Gadelha}, \bibinfo{person}{Yannick Hold-Geoffroy}, \bibinfo{person}{Karan Singh}, {and} \bibinfo{person}{Niloy Mitra}.} \bibinfo{year}{2023}\natexlab{}.
\newblock \showarticletitle{Diffusion Handles: Enabling 3D Edits for Diffusion Models by Lifting Activations to 3D}.
\newblock \bibinfo{journal}{\emph{arXiv preprint arXiv:2312.02190}} (\bibinfo{year}{2023}).
\newblock


\bibitem[Patashnik et~al\mbox{.}(2023)]%
        {Patashnik2023LocalizingOS}
\bibfield{author}{\bibinfo{person}{Or Patashnik}, \bibinfo{person}{Daniel Garibi}, \bibinfo{person}{Idan Azuri}, \bibinfo{person}{Hadar Averbuch-Elor}, {and} \bibinfo{person}{Daniel Cohen-Or}.} \bibinfo{year}{2023}\natexlab{}.
\newblock \showarticletitle{Localizing Object-level Shape Variations with Text-to-Image Diffusion Models}.
\newblock \bibinfo{journal}{\emph{2023 IEEE/CVF International Conference on Computer Vision (ICCV)}} (\bibinfo{year}{2023}), \bibinfo{pages}{22994--23004}.
\newblock
\urldef\tempurl%
\url{https://api.semanticscholar.org/CorpusID:257632209}
\showURL{%
\tempurl}


\bibitem[Phung et~al\mbox{.}(2023)]%
        {phung2023grounded}
\bibfield{author}{\bibinfo{person}{Quynh Phung}, \bibinfo{person}{Songwei Ge}, {and} \bibinfo{person}{Jia-Bin Huang}.} \bibinfo{year}{2023}\natexlab{}.
\newblock \showarticletitle{Grounded Text-to-Image Synthesis with Attention Refocusing}.
\newblock \bibinfo{journal}{\emph{arXiv preprint arXiv:2306.05427}} (\bibinfo{year}{2023}).
\newblock


\bibitem[Podell et~al\mbox{.}(2023)]%
        {Podell2023SDXLIL}
\bibfield{author}{\bibinfo{person}{Dustin Podell}, \bibinfo{person}{Zion English}, \bibinfo{person}{Kyle Lacey}, \bibinfo{person}{A. Blattmann}, \bibinfo{person}{Tim Dockhorn}, \bibinfo{person}{Jonas Muller}, \bibinfo{person}{Joe Penna}, {and} \bibinfo{person}{Robin Rombach}.} \bibinfo{year}{2023}\natexlab{}.
\newblock \showarticletitle{{SDXL}: Improving Latent Diffusion Models for High-Resolution Image Synthesis}.
\newblock \bibinfo{journal}{\emph{ArXiv}}  \bibinfo{volume}{abs/2307.01952} (\bibinfo{year}{2023}).
\newblock
\urldef\tempurl%
\url{https://api.semanticscholar.org/CorpusID:259341735}
\showURL{%
\tempurl}


\bibitem[Qi et~al\mbox{.}(2023)]%
        {qi2023fatezero}
\bibfield{author}{\bibinfo{person}{Chenyang Qi}, \bibinfo{person}{Xiaodong Cun}, \bibinfo{person}{Yong Zhang}, \bibinfo{person}{Chenyang Lei}, \bibinfo{person}{Xintao Wang}, \bibinfo{person}{Ying Shan}, {and} \bibinfo{person}{Qifeng Chen}.} \bibinfo{year}{2023}\natexlab{}.
\newblock \showarticletitle{Fatezero: Fusing attentions for zero-shot text-based video editing}. In \bibinfo{booktitle}{\emph{Proceedings of the IEEE/CVF International Conference on Computer Vision}}. \bibinfo{pages}{15932--15942}.
\newblock


\bibitem[Radford et~al\mbox{.}(2021)]%
        {Radford2021LearningTV}
\bibfield{author}{\bibinfo{person}{Alec Radford}, \bibinfo{person}{Jong~Wook Kim}, \bibinfo{person}{Chris Hallacy}, \bibinfo{person}{Aditya Ramesh}, \bibinfo{person}{Gabriel Goh}, \bibinfo{person}{Sandhini Agarwal}, \bibinfo{person}{Girish Sastry}, \bibinfo{person}{Amanda Askell}, \bibinfo{person}{Pamela Mishkin}, \bibinfo{person}{Jack Clark}, \bibinfo{person}{Gretchen Krueger}, {and} \bibinfo{person}{Ilya Sutskever}.} \bibinfo{year}{2021}\natexlab{}.
\newblock \showarticletitle{Learning Transferable Visual Models From Natural Language Supervision}. In \bibinfo{booktitle}{\emph{International Conference on Machine Learning}}.
\newblock


\bibitem[Ramesh et~al\mbox{.}(2022)]%
        {ramesh2022hierarchical}
\bibfield{author}{\bibinfo{person}{Aditya Ramesh}, \bibinfo{person}{Prafulla Dhariwal}, \bibinfo{person}{Alex Nichol}, \bibinfo{person}{Casey Chu}, {and} \bibinfo{person}{Mark Chen}.} \bibinfo{year}{2022}\natexlab{}.
\newblock \showarticletitle{Hierarchical text-conditional image generation with {CLIP} latents}.
\newblock \bibinfo{journal}{\emph{arXiv preprint arXiv:2204.06125}} (\bibinfo{year}{2022}).
\newblock


\bibitem[Richardson et~al\mbox{.}(2023)]%
        {richardson2023conceptlab}
\bibfield{author}{\bibinfo{person}{Elad Richardson}, \bibinfo{person}{Kfir Goldberg}, \bibinfo{person}{Yuval Alaluf}, {and} \bibinfo{person}{Daniel Cohen-Or}.} \bibinfo{year}{2023}\natexlab{}.
\newblock \showarticletitle{ConceptLab: Creative Generation using Diffusion Prior Constraints}.
\newblock \bibinfo{journal}{\emph{arXiv preprint arXiv:2308.02669}} (\bibinfo{year}{2023}).
\newblock


\bibitem[Rombach et~al\mbox{.}(2021)]%
        {Rombach2021HighResolutionIS}
\bibfield{author}{\bibinfo{person}{Robin Rombach}, \bibinfo{person}{A. Blattmann}, \bibinfo{person}{Dominik Lorenz}, \bibinfo{person}{Patrick Esser}, {and} \bibinfo{person}{Bj{\"o}rn Ommer}.} \bibinfo{year}{2021}\natexlab{}.
\newblock \showarticletitle{High-Resolution Image Synthesis with Latent Diffusion Models}.
\newblock \bibinfo{journal}{\emph{2022 IEEE/CVF Conference on Computer Vision and Pattern Recognition (CVPR)}} (\bibinfo{year}{2021}), \bibinfo{pages}{10674--10685}.
\newblock


\bibitem[Ruiz et~al\mbox{.}(2023)]%
        {Ruiz2022DreamBoothFT}
\bibfield{author}{\bibinfo{person}{Nataniel Ruiz}, \bibinfo{person}{Yuanzhen Li}, \bibinfo{person}{Varun Jampani}, \bibinfo{person}{Yael Pritch}, \bibinfo{person}{Michael Rubinstein}, {and} \bibinfo{person}{Kfir Aberman}.} \bibinfo{year}{2023}\natexlab{}.
\newblock \showarticletitle{{DreamBooth}: Fine tuning text-to-image diffusion models for subject-driven generation}. In \bibinfo{booktitle}{\emph{Proceedings of the IEEE/CVF Conference on Computer Vision and Pattern Recognition}}. \bibinfo{pages}{22500--22510}.
\newblock


\bibitem[Sheynin et~al\mbox{.}(2023)]%
        {Sheynin2023EmuEP}
\bibfield{author}{\bibinfo{person}{Shelly Sheynin}, \bibinfo{person}{Adam Polyak}, \bibinfo{person}{Uriel Singer}, \bibinfo{person}{Yuval Kirstain}, \bibinfo{person}{Amit Zohar}, \bibinfo{person}{Oron Ashual}, \bibinfo{person}{Devi Parikh}, {and} \bibinfo{person}{Yaniv Taigman}.} \bibinfo{year}{2023}\natexlab{}.
\newblock \showarticletitle{Emu Edit: Precise Image Editing via Recognition and Generation Tasks}.
\newblock \bibinfo{journal}{\emph{ArXiv}}  \bibinfo{volume}{abs/2311.10089} (\bibinfo{year}{2023}).
\newblock
\urldef\tempurl%
\url{https://api.semanticscholar.org/CorpusID:265221391}
\showURL{%
\tempurl}


\bibitem[Shi et~al\mbox{.}(2023)]%
        {Shi2023DragDiffusionHD}
\bibfield{author}{\bibinfo{person}{Yujun Shi}, \bibinfo{person}{Chuhui Xue}, \bibinfo{person}{Jiachun Pan}, \bibinfo{person}{Wenqing Zhang}, \bibinfo{person}{Vincent Y.~F. Tan}, {and} \bibinfo{person}{Song Bai}.} \bibinfo{year}{2023}\natexlab{}.
\newblock \showarticletitle{DragDiffusion: Harnessing Diffusion Models for Interactive Point-based Image Editing}.
\newblock \bibinfo{journal}{\emph{ArXiv}}  \bibinfo{volume}{abs/2306.14435} (\bibinfo{year}{2023}).
\newblock
\urldef\tempurl%
\url{https://api.semanticscholar.org/CorpusID:259252555}
\showURL{%
\tempurl}


\bibitem[Sohl-Dickstein et~al\mbox{.}(2015)]%
        {sohl2015deep}
\bibfield{author}{\bibinfo{person}{Jascha Sohl-Dickstein}, \bibinfo{person}{Eric Weiss}, \bibinfo{person}{Niru Maheswaranathan}, {and} \bibinfo{person}{Surya Ganguli}.} \bibinfo{year}{2015}\natexlab{}.
\newblock \showarticletitle{Deep unsupervised learning using nonequilibrium thermodynamics}. In \bibinfo{booktitle}{\emph{International Conference on Machine Learning}}. PMLR, \bibinfo{pages}{2256--2265}.
\newblock


\bibitem[Song et~al\mbox{.}(2020)]%
        {song2020denoising}
\bibfield{author}{\bibinfo{person}{Jiaming Song}, \bibinfo{person}{Chenlin Meng}, {and} \bibinfo{person}{Stefano Ermon}.} \bibinfo{year}{2020}\natexlab{}.
\newblock \showarticletitle{Denoising Diffusion Implicit Models}. In \bibinfo{booktitle}{\emph{International Conference on Learning Representations}}.
\newblock


\bibitem[Song and Ermon(2019)]%
        {song2019generative}
\bibfield{author}{\bibinfo{person}{Yang Song} {and} \bibinfo{person}{Stefano Ermon}.} \bibinfo{year}{2019}\natexlab{}.
\newblock \showarticletitle{Generative modeling by estimating gradients of the data distribution}.
\newblock \bibinfo{journal}{\emph{Advances in Neural Information Processing Systems}}  \bibinfo{volume}{32} (\bibinfo{year}{2019}).
\newblock


\bibitem[Tewel et~al\mbox{.}(2024)]%
        {Tewel2024TrainingFreeCT}
\bibfield{author}{\bibinfo{person}{Yoad Tewel}, \bibinfo{person}{Omri Kaduri}, \bibinfo{person}{Rinon Gal}, \bibinfo{person}{Yoni Kasten}, \bibinfo{person}{Lior Wolf}, \bibinfo{person}{Gal Chechik}, {and} \bibinfo{person}{Yuval Atzmon}.} \bibinfo{year}{2024}\natexlab{}.
\newblock \showarticletitle{Training-Free Consistent Text-to-Image Generation}.
\newblock \bibinfo{journal}{\emph{ArXiv}}  \bibinfo{volume}{abs/2402.03286} (\bibinfo{year}{2024}).
\newblock
\urldef\tempurl%
\url{https://api.semanticscholar.org/CorpusID:267412997}
\showURL{%
\tempurl}


\bibitem[Tumanyan et~al\mbox{.}(2023)]%
        {pnpDiffusion2022}
\bibfield{author}{\bibinfo{person}{Narek Tumanyan}, \bibinfo{person}{Michal Geyer}, \bibinfo{person}{Shai Bagon}, {and} \bibinfo{person}{Tali Dekel}.} \bibinfo{year}{2023}\natexlab{}.
\newblock \showarticletitle{Plug-and-play diffusion features for text-driven image-to-image translation}. In \bibinfo{booktitle}{\emph{Proceedings of the IEEE/CVF Conference on Computer Vision and Pattern Recognition}}. \bibinfo{pages}{1921--1930}.
\newblock


\bibitem[von Platen et~al\mbox{.}(2022)]%
        {von-platen-etal-2022-diffusers}
\bibfield{author}{\bibinfo{person}{Patrick von Platen}, \bibinfo{person}{Suraj Patil}, \bibinfo{person}{Anton Lozhkov}, \bibinfo{person}{Pedro Cuenca}, \bibinfo{person}{Nathan Lambert}, \bibinfo{person}{Kashif Rasul}, \bibinfo{person}{Mishig Davaadorj}, {and} \bibinfo{person}{Thomas Wolf}.} \bibinfo{year}{2022}\natexlab{}.
\newblock \bibinfo{title}{Diffusers: State-of-the-art diffusion models}.
\newblock \bibinfo{howpublished}{\url{https://github.com/huggingface/diffusers}}.
\newblock


\bibitem[Voynov et~al\mbox{.}(2023)]%
        {Voynov2023PET}
\bibfield{author}{\bibinfo{person}{Andrey Voynov}, \bibinfo{person}{Q. Chu}, \bibinfo{person}{Daniel Cohen-Or}, {and} \bibinfo{person}{Kfir Aberman}.} \bibinfo{year}{2023}\natexlab{}.
\newblock \showarticletitle{P+: Extended Textual Conditioning in Text-to-Image Generation}.
\newblock \bibinfo{journal}{\emph{ArXiv}}  \bibinfo{volume}{abs/2303.09522} (\bibinfo{year}{2023}).
\newblock


\bibitem[Wang et~al\mbox{.}(2021)]%
        {Wang2021ImprovingGE}
\bibfield{author}{\bibinfo{person}{Jianyuan Wang}, \bibinfo{person}{Ceyuan Yang}, \bibinfo{person}{Yinghao Xu}, \bibinfo{person}{Yujun Shen}, \bibinfo{person}{Hongdong Li}, {and} \bibinfo{person}{Bolei Zhou}.} \bibinfo{year}{2021}\natexlab{}.
\newblock \showarticletitle{Improving GAN Equilibrium by Raising Spatial Awareness}.
\newblock \bibinfo{journal}{\emph{2022 IEEE/CVF Conference on Computer Vision and Pattern Recognition (CVPR)}} (\bibinfo{year}{2021}), \bibinfo{pages}{11275--11283}.
\newblock
\urldef\tempurl%
\url{https://api.semanticscholar.org/CorpusID:244772988}
\showURL{%
\tempurl}


\bibitem[Wang et~al\mbox{.}(2022)]%
        {Wang2022RewritingGR}
\bibfield{author}{\bibinfo{person}{Sheng-Yu Wang}, \bibinfo{person}{David Bau}, {and} \bibinfo{person}{Jun-Yan Zhu}.} \bibinfo{year}{2022}\natexlab{}.
\newblock \showarticletitle{Rewriting geometric rules of a GAN}.
\newblock \bibinfo{journal}{\emph{ACM Transactions on Graphics (TOG)}}  \bibinfo{volume}{41} (\bibinfo{year}{2022}), \bibinfo{pages}{1 -- 16}.
\newblock
\urldef\tempurl%
\url{https://api.semanticscholar.org/CorpusID:250956766}
\showURL{%
\tempurl}


\bibitem[Winter et~al\mbox{.}(2024)]%
        {Winter2024ObjectDropBC}
\bibfield{author}{\bibinfo{person}{Daniel Winter}, \bibinfo{person}{Matan Cohen}, \bibinfo{person}{Shlomi Fruchter}, \bibinfo{person}{Yael Pritch}, \bibinfo{person}{Alex Rav-Acha}, {and} \bibinfo{person}{Yedid Hoshen}.} \bibinfo{year}{2024}\natexlab{}.
\newblock \showarticletitle{ObjectDrop: Bootstrapping Counterfactuals for Photorealistic Object Removal and Insertion}.
\newblock \bibinfo{journal}{\emph{ArXiv}}  \bibinfo{volume}{abs/2403.18818} (\bibinfo{year}{2024}).
\newblock
\urldef\tempurl%
\url{https://api.semanticscholar.org/CorpusID:268724005}
\showURL{%
\tempurl}


\bibitem[Wolf et~al\mbox{.}(2020)]%
        {wolf-etal-2020-transformers}
\bibfield{author}{\bibinfo{person}{Thomas Wolf}, \bibinfo{person}{Lysandre Debut}, \bibinfo{person}{Victor Sanh}, \bibinfo{person}{Julien Chaumond}, \bibinfo{person}{Clement Delangue}, \bibinfo{person}{Anthony Moi}, \bibinfo{person}{Pierric Cistac}, \bibinfo{person}{Tim Rault}, \bibinfo{person}{Rémi Louf}, \bibinfo{person}{Morgan Funtowicz}, \bibinfo{person}{Joe Davison}, \bibinfo{person}{Sam Shleifer}, \bibinfo{person}{Patrick von Platen}, \bibinfo{person}{Clara Ma}, \bibinfo{person}{Yacine Jernite}, \bibinfo{person}{Julien Plu}, \bibinfo{person}{Canwen Xu}, \bibinfo{person}{Teven~Le Scao}, \bibinfo{person}{Sylvain Gugger}, \bibinfo{person}{Mariama Drame}, \bibinfo{person}{Quentin Lhoest}, {and} \bibinfo{person}{Alexander~M. Rush}.} \bibinfo{year}{2020}\natexlab{}.
\newblock \showarticletitle{Transformers: State-of-the-Art Natural Language Processing}. In \bibinfo{booktitle}{\emph{Proceedings of the 2020 Conference on Empirical Methods in Natural Language Processing: System Demonstrations}}. \bibinfo{publisher}{Association for Computational Linguistics}, \bibinfo{address}{Online}, \bibinfo{pages}{38--45}.
\newblock
\urldef\tempurl%
\url{https://www.aclweb.org/anthology/2020.emnlp-demos.6}
\showURL{%
\tempurl}


\bibitem[Wu et~al\mbox{.}(2022)]%
        {Wu2022TuneAVideoOT}
\bibfield{author}{\bibinfo{person}{Jay~Zhangjie Wu}, \bibinfo{person}{Yixiao Ge}, \bibinfo{person}{Xintao Wang}, \bibinfo{person}{Weixian Lei}, \bibinfo{person}{Yuchao Gu}, \bibinfo{person}{Wynne Hsu}, \bibinfo{person}{Ying Shan}, \bibinfo{person}{Xiaohu Qie}, {and} \bibinfo{person}{Mike~Zheng Shou}.} \bibinfo{year}{2022}\natexlab{}.
\newblock \showarticletitle{Tune-A-Video: One-Shot Tuning of Image Diffusion Models for Text-to-Video Generation}.
\newblock \bibinfo{journal}{\emph{2023 IEEE/CVF International Conference on Computer Vision (ICCV)}} (\bibinfo{year}{2022}), \bibinfo{pages}{7589--7599}.
\newblock
\urldef\tempurl%
\url{https://api.semanticscholar.org/CorpusID:254974187}
\showURL{%
\tempurl}


\bibitem[Xu et~al\mbox{.}(2023)]%
        {Xu2023OpenVocabularyPS}
\bibfield{author}{\bibinfo{person}{Jiarui Xu}, \bibinfo{person}{Sifei Liu}, \bibinfo{person}{Arash Vahdat}, \bibinfo{person}{Wonmin Byeon}, \bibinfo{person}{Xiaolong Wang}, {and} \bibinfo{person}{Shalini~De Mello}.} \bibinfo{year}{2023}\natexlab{}.
\newblock \showarticletitle{Open-Vocabulary Panoptic Segmentation with Text-to-Image Diffusion Models}.
\newblock \bibinfo{journal}{\emph{2023 IEEE/CVF Conference on Computer Vision and Pattern Recognition (CVPR)}} (\bibinfo{year}{2023}), \bibinfo{pages}{2955--2966}.
\newblock
\urldef\tempurl%
\url{https://api.semanticscholar.org/CorpusID:257405338}
\showURL{%
\tempurl}


\bibitem[Yang et~al\mbox{.}(2022)]%
        {Yang2022PaintBE}
\bibfield{author}{\bibinfo{person}{Binxin Yang}, \bibinfo{person}{Shuyang Gu}, \bibinfo{person}{Bo Zhang}, \bibinfo{person}{Ting Zhang}, \bibinfo{person}{Xuejin Chen}, \bibinfo{person}{Xiaoyan Sun}, \bibinfo{person}{Dong Chen}, {and} \bibinfo{person}{Fang Wen}.} \bibinfo{year}{2022}\natexlab{}.
\newblock \showarticletitle{Paint by Example: Exemplar-based Image Editing with Diffusion Models}.
\newblock \bibinfo{journal}{\emph{2023 IEEE/CVF Conference on Computer Vision and Pattern Recognition (CVPR)}} (\bibinfo{year}{2022}), \bibinfo{pages}{18381--18391}.
\newblock
\urldef\tempurl%
\url{https://api.semanticscholar.org/CorpusID:253802085}
\showURL{%
\tempurl}


\bibitem[Yang et~al\mbox{.}(2023)]%
        {yang2023reco}
\bibfield{author}{\bibinfo{person}{Zhengyuan Yang}, \bibinfo{person}{Jianfeng Wang}, \bibinfo{person}{Zhe Gan}, \bibinfo{person}{Linjie Li}, \bibinfo{person}{Kevin Lin}, \bibinfo{person}{Chenfei Wu}, \bibinfo{person}{Nan Duan}, \bibinfo{person}{Zicheng Liu}, \bibinfo{person}{Ce Liu}, \bibinfo{person}{Michael Zeng}, {et~al\mbox{.}}} \bibinfo{year}{2023}\natexlab{}.
\newblock \showarticletitle{Reco: Region-controlled text-to-image generation}. In \bibinfo{booktitle}{\emph{Proceedings of the IEEE/CVF Conference on Computer Vision and Pattern Recognition}}. \bibinfo{pages}{14246--14255}.
\newblock


\bibitem[Yu et~al\mbox{.}(2015)]%
        {Yu2015LSUNCO}
\bibfield{author}{\bibinfo{person}{Fisher Yu}, \bibinfo{person}{Yinda Zhang}, \bibinfo{person}{Shuran Song}, \bibinfo{person}{Ari Seff}, {and} \bibinfo{person}{Jianxiong Xiao}.} \bibinfo{year}{2015}\natexlab{}.
\newblock \showarticletitle{LSUN: Construction of a Large-scale Image Dataset using Deep Learning with Humans in the Loop}.
\newblock \bibinfo{journal}{\emph{ArXiv}}  \bibinfo{volume}{abs/1506.03365} (\bibinfo{year}{2015}).
\newblock
\urldef\tempurl%
\url{https://api.semanticscholar.org/CorpusID:8317437}
\showURL{%
\tempurl}


\bibitem[Yu et~al\mbox{.}(2022)]%
        {yu2022scaling}
\bibfield{author}{\bibinfo{person}{Jiahui Yu}, \bibinfo{person}{Yuanzhong Xu}, \bibinfo{person}{Jing~Yu Koh}, \bibinfo{person}{Thang Luong}, \bibinfo{person}{Gunjan Baid}, \bibinfo{person}{Zirui Wang}, \bibinfo{person}{Vijay Vasudevan}, \bibinfo{person}{Alexander Ku}, \bibinfo{person}{Yinfei Yang}, \bibinfo{person}{Burcu~Karagol Ayan}, {et~al\mbox{.}}} \bibinfo{year}{2022}\natexlab{}.
\newblock \showarticletitle{Scaling Autoregressive Models for Content-Rich Text-to-Image Generation}.
\newblock \bibinfo{journal}{\emph{arXiv preprint arXiv:2206.10789}} (\bibinfo{year}{2022}).
\newblock


\bibitem[Zhang et~al\mbox{.}(2023)]%
        {zhang2023controlnet}
\bibfield{author}{\bibinfo{person}{Lvmin Zhang}, \bibinfo{person}{Anyi Rao}, {and} \bibinfo{person}{Maneesh Agrawala}.} \bibinfo{year}{2023}\natexlab{}.
\newblock \showarticletitle{Adding Conditional Control to Text-to-Image Diffusion Models}. In \bibinfo{booktitle}{\emph{Proceedings of the IEEE/CVF International Conference on Computer Vision (ICCV)}}. \bibinfo{pages}{3836--3847}.
\newblock


\bibitem[Zheng et~al\mbox{.}(2023)]%
        {zheng2023layoutdiffusion}
\bibfield{author}{\bibinfo{person}{Guangcong Zheng}, \bibinfo{person}{Xianpan Zhou}, \bibinfo{person}{Xuewei Li}, \bibinfo{person}{Zhongang Qi}, \bibinfo{person}{Ying Shan}, {and} \bibinfo{person}{Xi Li}.} \bibinfo{year}{2023}\natexlab{}.
\newblock \showarticletitle{LayoutDiffusion: Controllable Diffusion Model for Layout-to-image Generation}. In \bibinfo{booktitle}{\emph{Proceedings of the IEEE/CVF Conference on Computer Vision and Pattern Recognition}}. \bibinfo{pages}{22490--22499}.
\newblock


\end{thebibliography}

\clearpage
\appendix
\section{Implementation Details}
\label{sec:implementation_details}

In \Cref{sec:method_implementation_details} we start by providing the implementation details of our method. Next, in \Cref{sec:baselines_implementation_details} we provide the baselines' implementation details. Later, in \Cref{sec:autoamtic_metrics_implementation_details} we provide the implementation details of the automatic metrics we used. Finally, in \Cref{sec:user_study_implementation_details} we provide the detail of the user study we conducted.

\subsection{Implementation Details of Our Method}
\label{sec:method_implementation_details}

Below, we provide the full implementation details of our method: in \Cref{sec:gsa_details} we start by providing the implementation details for the gated self-attention visualization we used, next, in \Cref{sec:sa_anchoring_details} we provide the implementation details of the soft self-attention anchoring we used, and finally, in \Cref{sec:bld_integration_details} we explain about the Blended Latent Diffusion integration.

\subsubsection{Gated Self-Attention Visualization Implementation Details}
\label{sec:gsa_details}

\begin{figure*}[tp]
    \centering
    \includegraphics[width=1\linewidth]{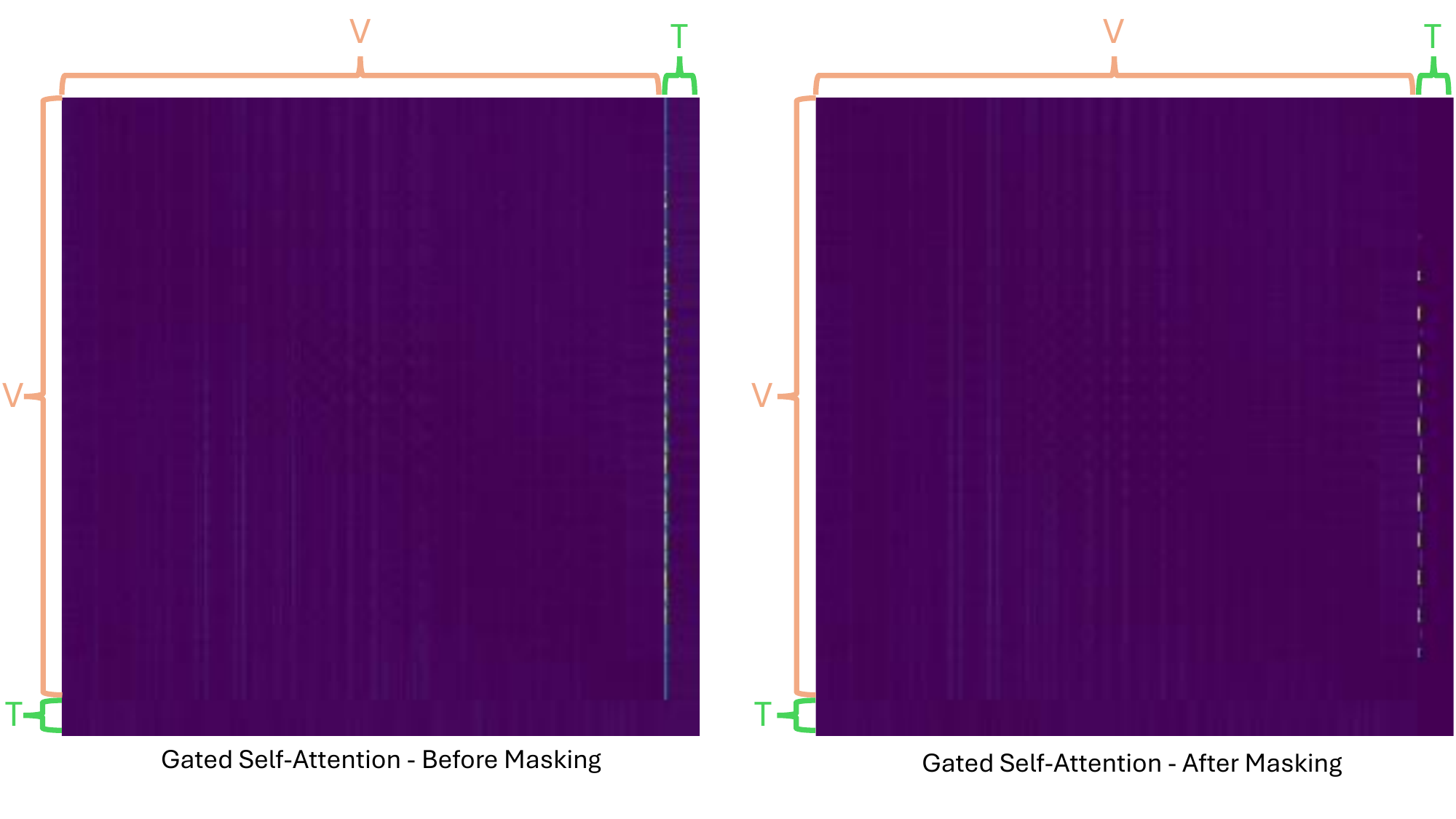}
    
    \caption{\textbf{Full Gated Self-Attention Visualization.} We provide our visualization for the gated self-attention layers, before our inference time masking (left) and after it (right). As can be seen, even though there is no constraint on the attention distribution (as in any other self-attention layer), we found empirically that the vast majority of the attention is being formed between the used pooled textual tokens (the first part of $T$) and the visual tokens $V$. And not between the sets themselves. This behavior suggests that this kind of layers behaves de facto as cross attention layers. Our inference-time masking mechanism (right) constrain the attention between the textual tokens and only their corresponding visual tokens within their blobs.}
    \label{fig:full_gsa_visualization}
\end{figure*}

As explained in
\Cref{sec:blobgen_entanglement}
we offer a method to visualize the gated self-attention layer of GLIGEN~\cite{Li2023GLIGENOG} and BlobGEN~\cite{nie2024compositional}. In gated self-attention, a projection layer first converts the CLIP \cite{Radford2021LearningTV} text embeddings of the text description $S_i$ to the text tokens $T = \{t_1, ... t_n\}$. They are then merged with the visual tokens $V = \{v_1, ... v_k\}$ into a unified set $V \cup T = \{v_1, ... v_k, t_1, ... t_n\}$, which altogether are used to calculate the self-attention features, using the standard self-attention mechanism (plus a gated skip connection).

We examine the attention between the projected text token $t_i$ and all the visual tokens $V$. This is a $K$-dimensional vector, which we first reshape into two dimensions $\sqrt{K}\times \sqrt{K}$, and then resize to a canonical size. We term these maps ``reshaped self-attention'', which are averaged over all the denoising steps. We average these maps across all the diffusion steps and all the layers, by resizing them to a canonical size.

In \Cref{fig:full_gsa_visualization} we provide a full visualization of these maps. As can be seen, even though there is no constraint on the attention distribution (as in any other self-attention layer), we found empirically that the vast majority of the attention is being formed between the used pooled textual tokens (the first part of $T$) and the visual tokens $V$. There is little interaction within the individual sets themselves. This behavior suggests that this kind of self-attention layers behaves as de facto cross attention layers. Our inference-time masking mechanism (right) further constrains each textual token to only attend to its corresponding visual token within the blob region.

Please note that these kind of visualizations and the masking manipulations are different from the those common in the text-to-image diffusion-based models literature \cite{hertz2022prompt,Avrahami2023BreakASceneEM,Chefer2023AttendandExciteAS} as we do not manipulate the traditional \emph{cross-attention} layers but the gated \emph{self-attention} layers.

\subsubsection{Soft Self-Attention Anchoring Implementation Details}
\label{sec:sa_anchoring_details}

As explained in
\Cref{sec:object_moving_generated},
we propose the soft self-attention anchoring to fuse the spatial information from the localized model and the appearance information from the input source image. Specifically, in the first $\rho = \frac{T}{2}$ steps of the denoising process, we perform an adaptive, \emph{soft blending} of the attention features of the generated target image with the features of the source image. The interpolation coefficient is time-dependent: 
we take more visual appearance from the source image in the beginning but more spatial information from the target image in the later steps. 

Next, during the last $T - \rho$ steps of the denoising process, we use the soft blending result $O_a$ as anchor points for the target object. In each denoising step $t\in[T-\rho, ..., 2, 1]$ and each self-attention layer, we perform the \emph{nearest-neighbor copying}: each entry from the anchor attention features $O_a$ within the target blob $B_d$ is replaced by its nearest-neighbor entry from the source attention features $O_s$ within the source blob $B_s$. To calculate the nearest-neighbor, we normalize each self-attention entry of $O$ and calculate its cosine similarity with each entry of $O_s$. Please note that the nearest-neighbor operation is only within the source blobs $B_s$ and destination blob $B_d$ boundaries. To calculate the these blobs, we reshape them to the corresponding self-attention size of each layer.

\subsubsection{Blended Latent Diffusion Integration}
\label{sec:bld_integration_details}
Blended Latent Diffusion~\cite{blended_2022_CVPR,avrahami2023blendedlatent} is a method designed for localized image editing using text-to-image diffusion models. the input image is fused into the diffusion process along with an input mask to preserve it background, while encouraging the generated content (in the unmasked area) to be consistent to the background. We also use this method in our pipeline of editing real images, as introduced in
\Cref{sec:object_moving_real}.
Given the source blob $B_s$ and the destination blob $B_d$ provided by the user, we take the union blob that contains both of them $B_u = B_s \cup B_d$, and morphologically dilate it with a kernel of a size of $50 \times 50$. We treat this dilated blob as the editable area, which we provide to the Blended Latent Diffusion method to edit real images during the entire diffusion process (\ie~ the hyperparameter of noising diffusion steps $k=T$, where $T$ is the total number of diffusion steps).

\subsection{Implementation Details of Baselines}
\label{sec:baselines_implementation_details}

\begin{table}
    \centering
    \caption{\textbf{Inference Time Comparison.} We report the inference time of the baselines and our method of editing a single $512 \times 512$ image. All the reported running times we calculated using a single NVIDIA A100 GPU.}
        \begin{tabular}{>{\columncolor[gray]{0.95}}lc}
            \toprule

            \textbf{Method} & 
            Inference time $(sec)$ 
            \\
            
            \midrule

            PBE \cite{Yang2022PaintBE} &
            9 sec
            \\

            Diffusion SG \cite{self_guidance} &
            14 sec
            \\

            Anydoor \cite{Chen2023AnyDoorZO} &
            9 sec
            \\

            DragDiffusion \cite{Shi2023DragDiffusionHD} &
            148 sec
            \\

            DragonDiffusion \cite{Mou2023DragonDiffusionED} &
            13 sec
            \\

            DiffEditor \cite{Mou2024DiffEditorBA} &
            13 sec
            \\

            \midrule

            DiffUHaul (ours) &
            13 sec
            \\
            
            \bottomrule
        \end{tabular}
    \label{tab:inference_times_comparison}
\end{table}

As described in
\Cref{sec:comparisons},
we compare our method against the available, most relevant object dragging baselines: Paint-By-Example \cite{Yang2022PaintBE}, AnyDoor \cite{Chen2023AnyDoorZO}, Diffusion Self-Guidance \cite{self_guidance}, DragDiffusion \cite{Shi2023DragDiffusionHD}, DragonDiffusion \cite{Mou2023DragonDiffusionED} and DiffEditor \cite{Mou2024DiffEditorBA}. Out of these, Diffusion Self-Guidance, DragonDiffusion and DiffEditor directly support the task of object dragging. The rest of these baselines need some adaptations to our problem, as described below.

Paint-By-Example and AnyDoor present a way to add an object to an image. Hence, in order to convert it to our problem setting we constructed a designated pipline: we started taking the source image $I$ and inpaint the source area blob $B_s$ using Stable Diffusion Inpaint \cite{von-platen-etal-2022-diffusers}, to get an inpainted version $\hat{I}$. Then, we used Paint-By-Example/Anydoor to inpaint the new image $\hat{I}$ again, in the target blob area $B_d$ by providing the original object in the original image $I$ as a reference.

DragDiffusion is originally designed to tackle the problem of keypoint-based dragging. Thus, in order to adapt it to our method, we take the centroid of the source blob $B_s$ as well as other points sampled inside the source blob region, and then translate them to the target blob $B_d$.

We used the official baseline implementations with a comparable backbone of Stable Diffusion v1 \cite{Rombach2021HighResolutionIS} using 50 DDIM diffusion steps, except Diffusion Self-Guidance \cite{Epstein2022BlobGANSD}, of which the only available implementation is based upon SDXL \cite{Podell2023SDXLIL}.

In \Cref{tab:inference_times_comparison} we report the inference time of our method and the baselines for a single image editing using an NVIDIA A100 GPU. All the methods takes around 10 seconds, except DragDiffusion that is using an extensive LoRA~\cite{lora, Horwitz2024RecoveringTP} training and latent optimization, which increases the inference time significantly.

We used the following third-party packages in this research:
\begin{itemize}
    \item Official GLIGEN~\cite{Li2023GLIGENOG} implementation at \url{https://github.com/gligen/GLIGEN}.
    \item Official Paint-By-Example~\cite{Yang2022PaintBE} Diffusers~\cite{von-platen-etal-2022-diffusers} implementation.
    \item Official Diffusion Self-Guidance~\cite{self_guidance} SDXL implementation at \url{https://colab.research.google.com/drive/1SEM1R9mI9cF-aFpqg3NqHP8gN8irHuJi}.
    \item Official AnyDoor~\cite{Chen2023AnyDoorZO} implementation at \url{https://github.com/ali-vilab/AnyDoor}.
    \item Official DragDiffusion~\cite{Shi2023DragDiffusionHD} implementation at \url{https://github.com/Yujun-Shi/DragDiffusion}.
    \item Official DragonDiffusion~\cite{Mou2023DragonDiffusionED} and DiffEditor \cite{Mou2024DiffEditorBA} at \url{https://github.com/MC-E/DragonDiffusion}.
    \item DINOv2~\cite{Oquab2023DINOv2LR} ViT-g/14 implementation by HuggingFace Transformers~\cite{wolf-etal-2020-transformers} at \url{https://github.com/huggingface/transformers}.
\end{itemize}

\subsection{Implementation Details of Automatic Metrics }
\label{sec:autoamtic_metrics_implementation_details}

As described in
\Cref{sec:comparisons},
in order to automatically compare our method against baselines quantitatively, we utilized COCO~\cite{Lin2014MicrosoftCC} validation dataset. We filtered it to contain only images with a main ``thing'' class object (the number of ``stuff'' object is unbounded) that occupies at least 5\% of the image size, but not more than 25\% of the image size. Then, we utilize ODISE~\cite{Xu2023OpenVocabularyPS} to get instance segmentation maps, then we use an ellipse fitting optimization with the goal of maximizing the Intersection Over Union (IOU) between the ellipse and the generated mask. Next, we crop a local region around each blob and use  LLaVA-1.5~\cite{Liu2023ImprovedBW} for the local captioning. Finally, we choose a random location for the target blob $B_d$ that is at least 64 pixels. This resulted with 672 filtered images and 8 target blob locations per image, which is a total of 6,048 evaluated samples per baseline.

Next, we propose using three metrics: foreground similarity, object traces and realism. Foreground similarity quantifies whether the source object is indeed dragged to the target location. To this end, we crop a tight square area around the source blob $B_s$ in the source image $I$, and the target blob $B_d$ in the target image $T_t$. Next, we align the crops to a canonical position and mask the background in these crops by aligning the object to the left side of the image (in order to avoid translation artifacts). Finally, we utilize DINOv2 \cite{Oquab2023DINOv2LR} to measure the perceptual similarity between these crops. We strive to \emph{maximize} this metric.

Similarly, in order to measure the object duplication phenomenon, we crop a tight square area around the source blob $B_s$ in the source image $I$, and around the source blob $B_s$ in the target image $I'$. Next, we mask the target blob $B_d$ area in the target image $I'$. Finally, we utilize again DINOv2 \cite{Oquab2023DINOv2LR} to measure the perceptual similarity between these crops. We strive to \emph{minimize} this metric. Lastly, in order to measure the realism of the image, we compute the KID score \cite{Binkowski2018DemystifyingMG} using 672 real and generated images. The reason of using KID instead of the FID score~\citep{heusel2017gans} is that it better aligns with human perception of image generation quality when the provided real and fake sets are small. 

\begin{figure}[th]
    \includegraphics[width=1\linewidth]{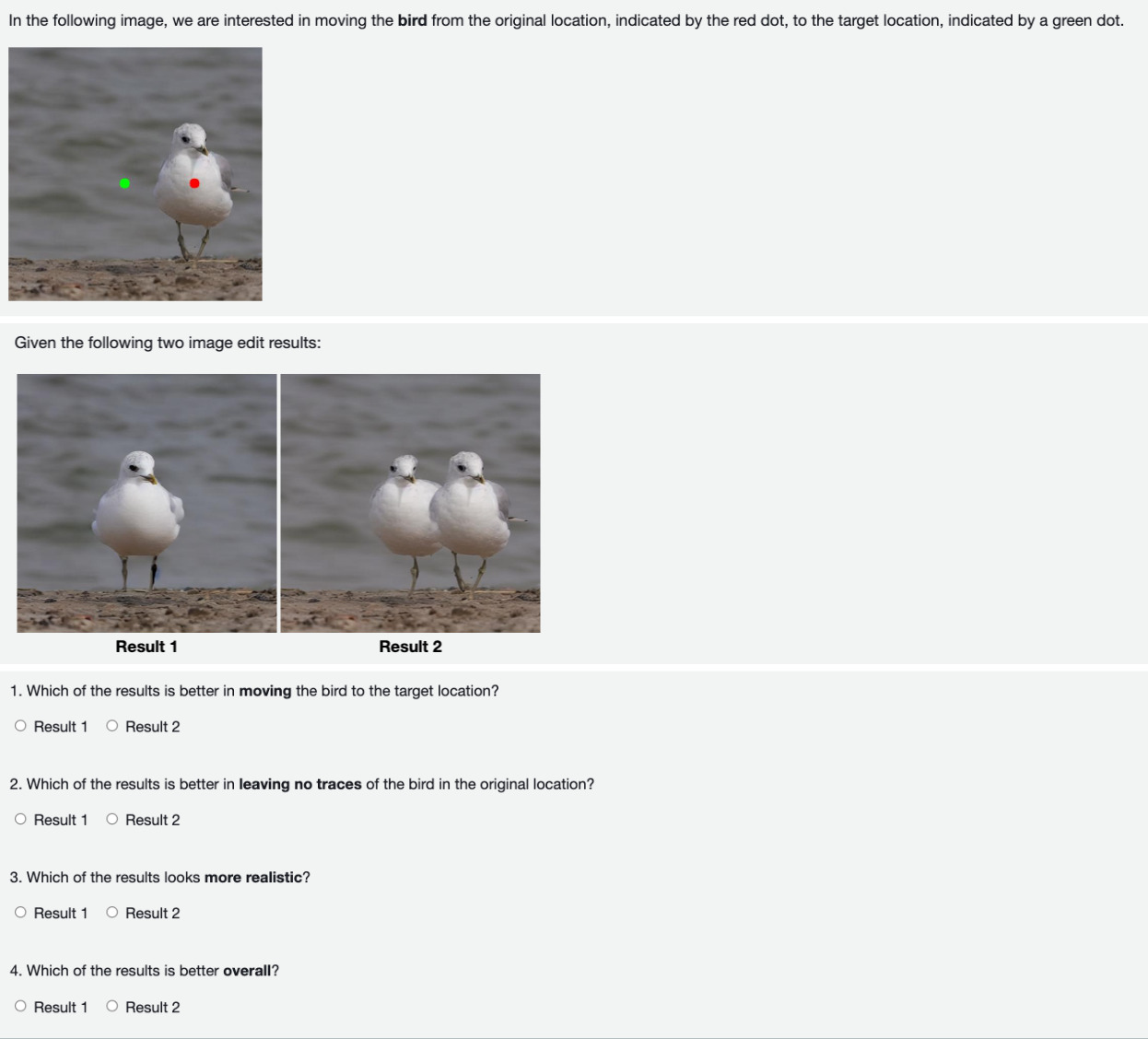}
    
    \caption{\textbf{User Study Trial.} We provide an example of one trial task in the user study we conducted using Amazon Mechanical Turk (AMT)~\cite{amt}. The users were asked four questions of a two-alternative forced-choice format. The full instructions can be seen in \Cref{fig:user_study_instructions}.}
    \label{fig:user_study_trial}
\end{figure}

\begin{figure}[th]
    \includegraphics[width=1\linewidth]{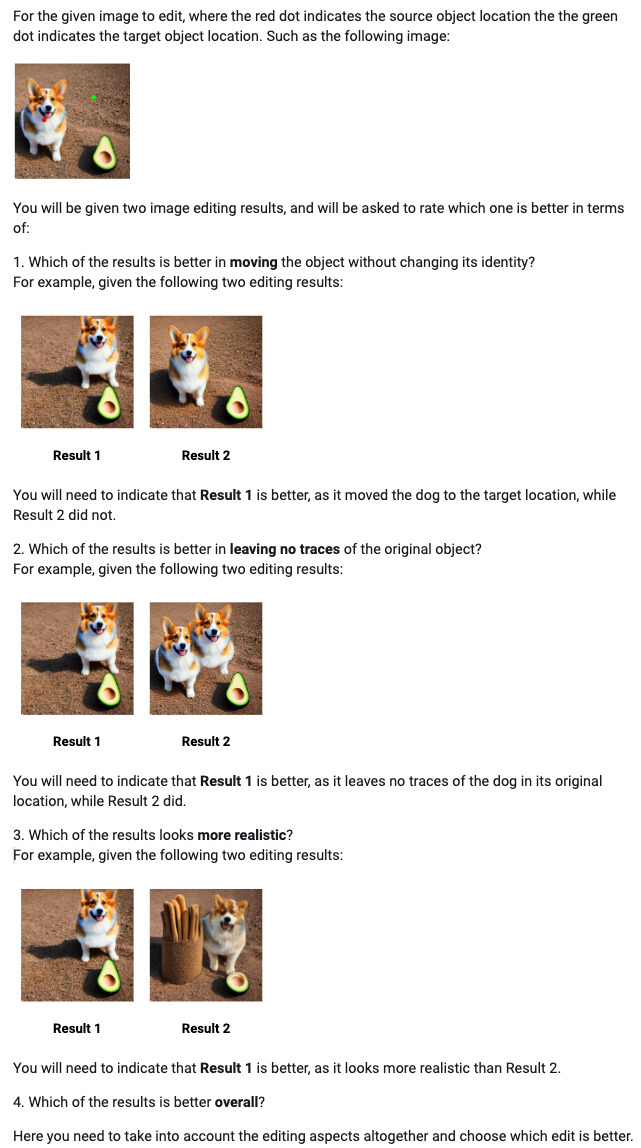}
    
    \caption{\textbf{User Study Instructions.} We provide the full instructions for the user study we conducted using Amazon Mechanical Turk (AMT)~\cite{amt}, to compare our method with each baseline.}
    \label{fig:user_study_instructions}
\end{figure}

\subsection{User Study Implementation Details}
\label{sec:user_study_implementation_details}

\begin{table}[th]
    \centering
    \caption{\textbf{User Study Statistical Significance.} A binomial statistical test of the user study results suggests that our results are statistically significant (p-value < 5\%)}
    \begin{adjustbox}{width=1\columnwidth}
        \begin{tabular}{>{\columncolor[gray]{0.95}}lcccc}
            \toprule

            \textbf{Ours vs} & 
            Dragging $(\uparrow)$ &
            No traces $(\uparrow)$ &
            Realism $(\uparrow)$ &
            Overall $(\uparrow)$
            \\

            &
            p-value &
            p-value &
            p-value &
            p-value
            \\

            \midrule

            PBE \cite{Yang2022PaintBE} &
            < 1e-8 &
            < 1e-8 &
            < 1e-8 &
            < 1e-8
            \\

            DiffusionSG \cite{self_guidance} &
            < 1e-8 &
            < 1e-8 &
            < 1e-8 &
            < 1e-8
            \\

            Anydoor \cite{Chen2023AnyDoorZO} &
            < 1e-8 &
            < 1e-8 &
            < 1e-8 &
            < 1e-8
            \\

            DragDiffusion \cite{Shi2023DragDiffusionHD} &
            < 1e-8 &
            < 2e-5 &
            < 6e-6 &
            < 5e-7
            \\

            DragonDiffusion \cite{Mou2023DragonDiffusionED} &
            < 1e-8 &
            < 1e-8 &
            < 1e-8 &
            < 1e-8
            \\

            DiffEditor \cite{Mou2024DiffEditorBA} &
            < 1e-8 &
            < 1e-8 &
            < 1e-8 &
            < 1e-8
            \\
            
            \bottomrule
        \end{tabular}
    \end{adjustbox}
    \label{tab:user_study_statistical_significance}
\end{table}

As explained in
\Cref{sec:user_study}
We conduct an extensive user study using the Amazon Mechanical Turk (AMT) platform \cite{amt}. We use the automatically extracted dataset as explained in 
Section 5.1 in the main paper.
We compare all the baselines using the standard two-alternative forced-choice format. Users are instructed the following ``In the following image, we are interested in moving the \{CATEGORY\} from the original location, indicated by the red dot, to the target location, indicated by a green dot.'' where \{CATEGORY\} is the ~\cite{Lin2014MicrosoftCC} object class category. Then, the users are given two editing results: our method and one of the baselines, and are asked: (1) ``Which of the results is better in moving the \{CATEGORY\} to the target location?'', (2) ``Which of the results is better in leaving no traces of the \{CATEGORY\} in the original location?'', (3) ``Which of the results looks more realistic?'' and (4) ``Which of the results is better overall?''. The users are also given detailed instructions with examples. An example of one trail can be seen in \Cref{fig:user_study_trial}, and the full instructions can be seen in \Cref{fig:user_study_instructions}.

We gather 7 ratings per sample, resulting 448 ratings per baseline, totaling 2,688 responses. The time
allotted per task is one hour, to allow the raters to properly
evaluate the results without time pressure. A binomial statistical test of the user study results, as presented in \Cref{tab:user_study_statistical_significance}, suggesting that our results are statistically significant (p-value < 5\%).

\section{Additional Results}
\label{sec:additional_results}

\begin{figure*}[t]
    \centering
    \setlength{\tabcolsep}{5pt}
    \renewcommand{\arraystretch}{0.9}
    \setlength{\ww}{0.28\columnwidth}

    \begin{tabular}{cccccc}
        \rotatebox[origin=c]{90}{{Input}}
        {\includegraphics[valign=c, width=\ww]{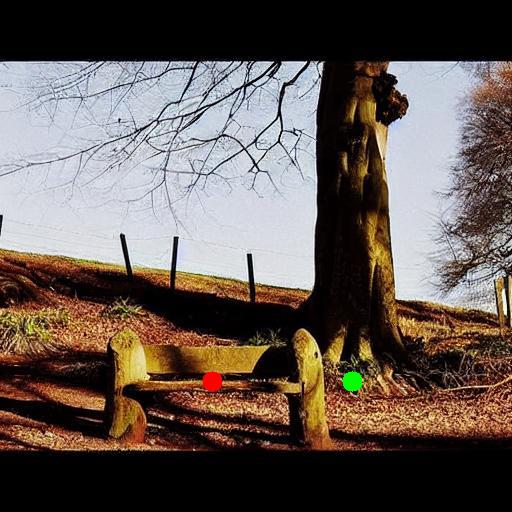}} &
        {\includegraphics[valign=c, width=\ww]{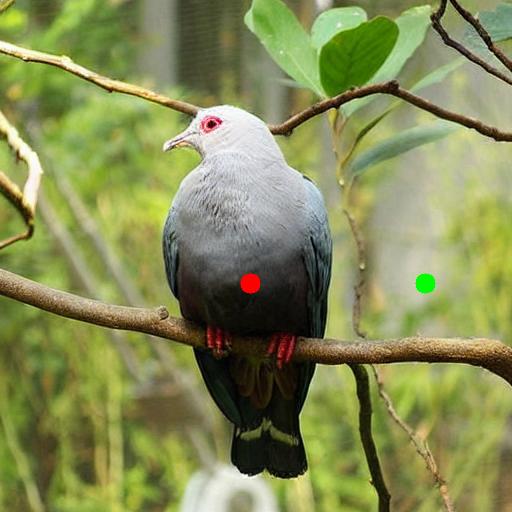}} &
        {\includegraphics[valign=c, width=\ww]{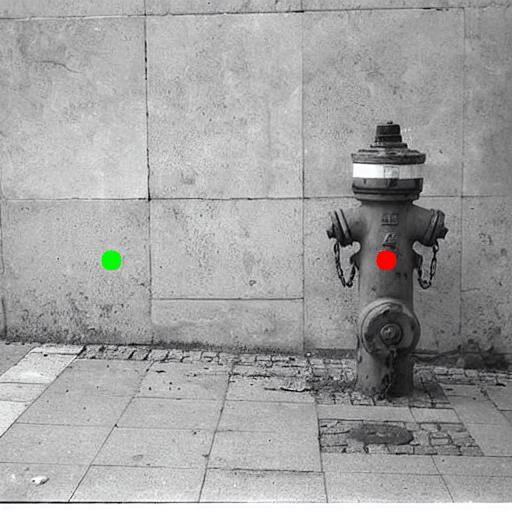}} &
        {\includegraphics[valign=c, width=\ww]{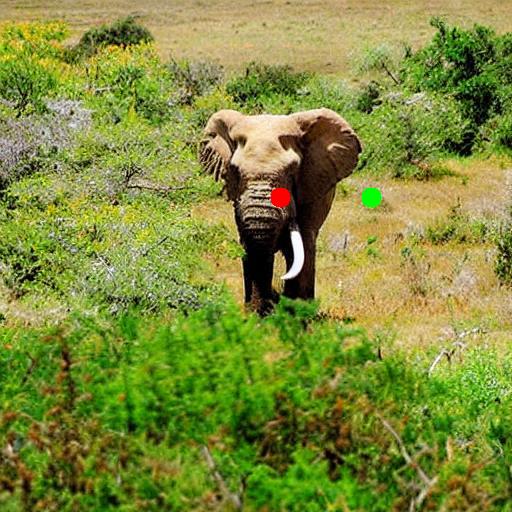}} &
        {\includegraphics[valign=c, width=\ww]{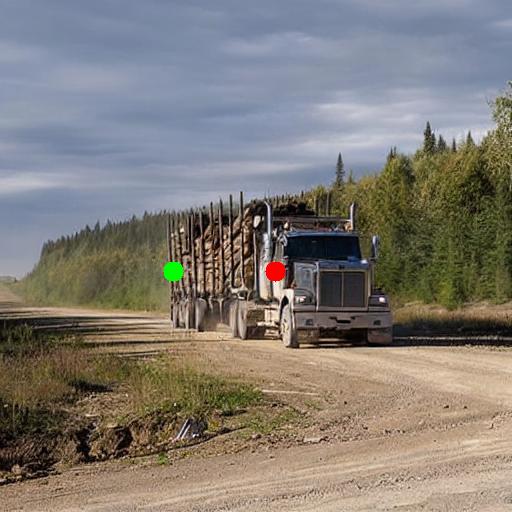}} &
        {\includegraphics[valign=c, width=\ww]{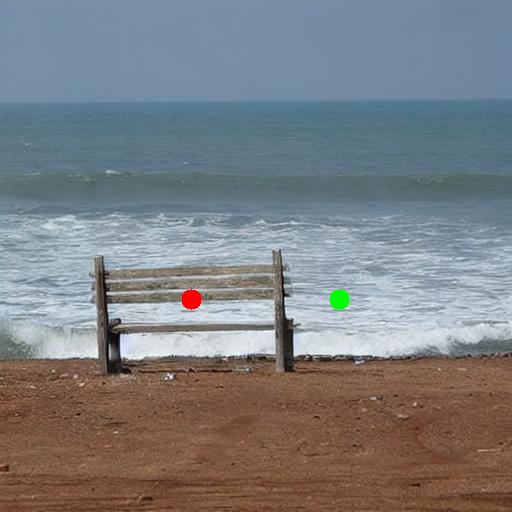}}
        \vspace{1px}
        \\

        \midrule

        \rotatebox[origin=c]{90}{{PBE}}
        {\includegraphics[valign=c, width=\ww]{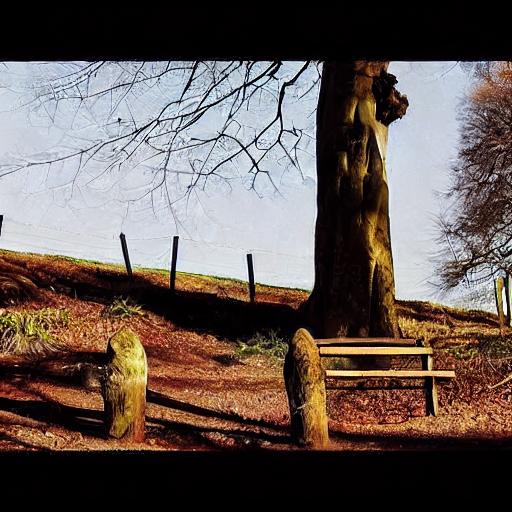}} &
        {\includegraphics[valign=c, width=\ww]{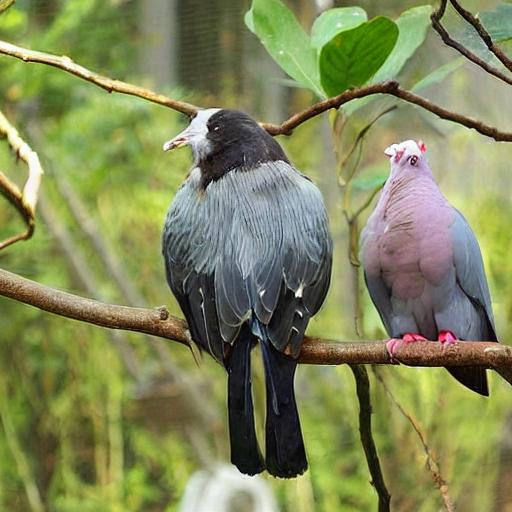}} &
        {\includegraphics[valign=c, width=\ww]{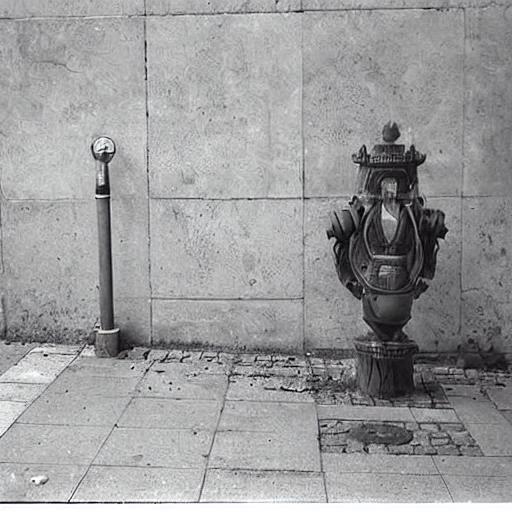}} &
        {\includegraphics[valign=c, width=\ww]{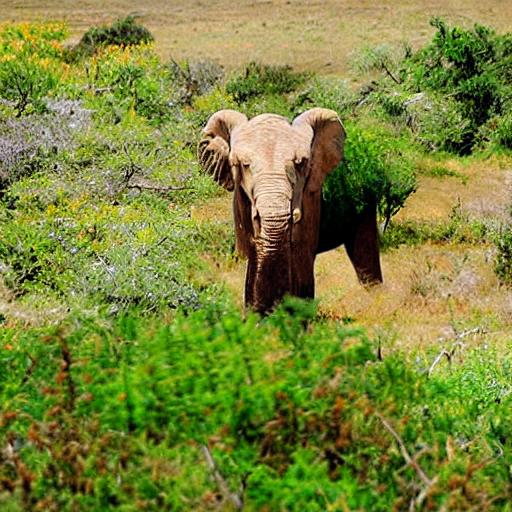}} &
        {\includegraphics[valign=c, width=\ww]{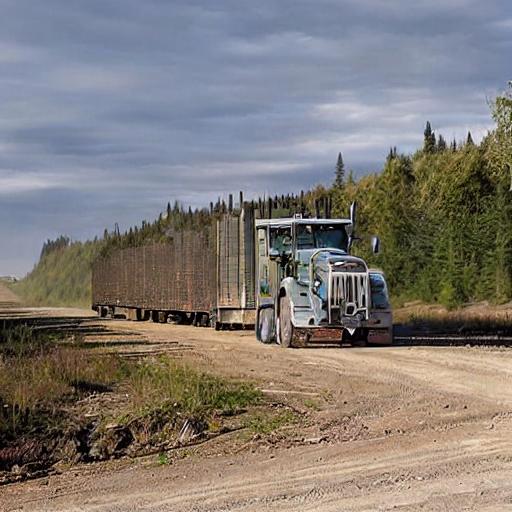}} &
        {\includegraphics[valign=c, width=\ww]{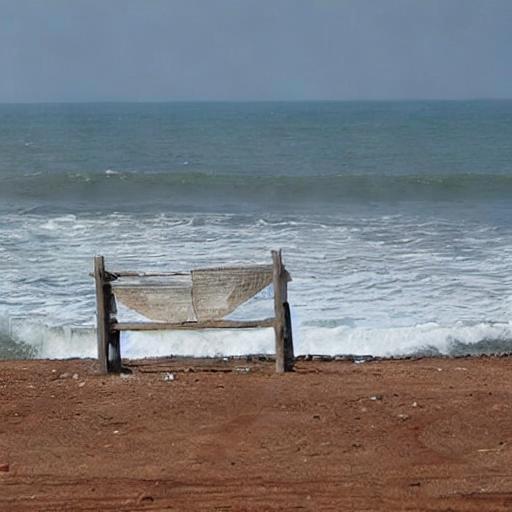}}
        \vspace{1px}
        \\

        \rotatebox[origin=c]{90}{{Diffusion SG}}
        {\includegraphics[valign=c, width=\ww]{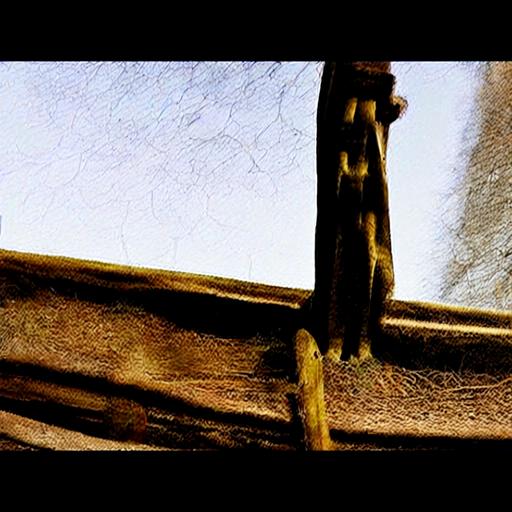}} &
        {\includegraphics[valign=c, width=\ww]{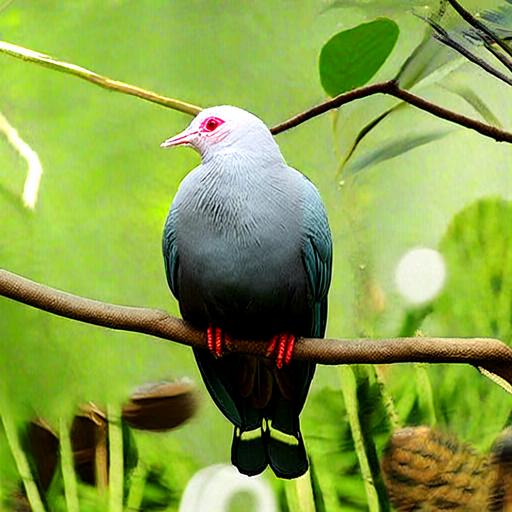}} &
        {\includegraphics[valign=c, width=\ww]{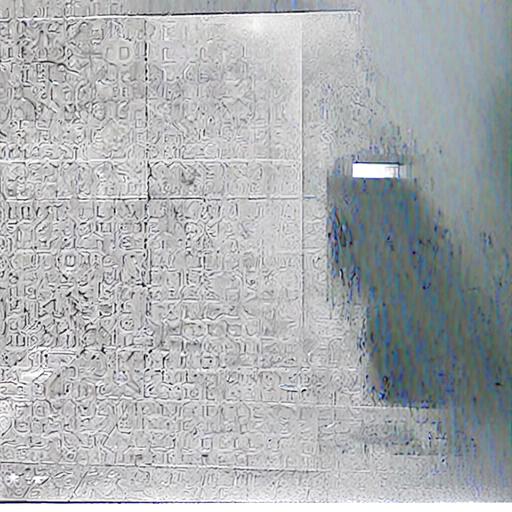}} &
        {\includegraphics[valign=c, width=\ww]{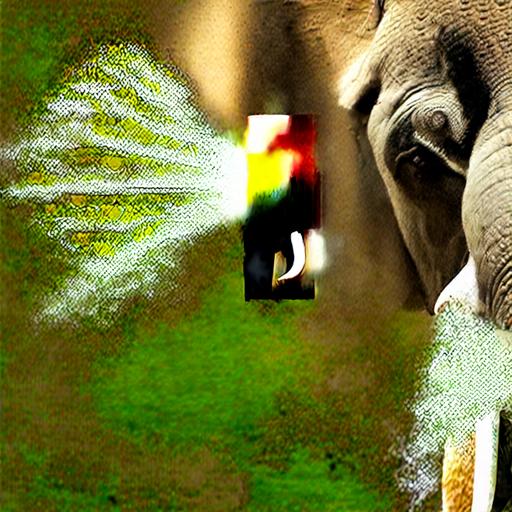}} &
        {\includegraphics[valign=c, width=\ww]{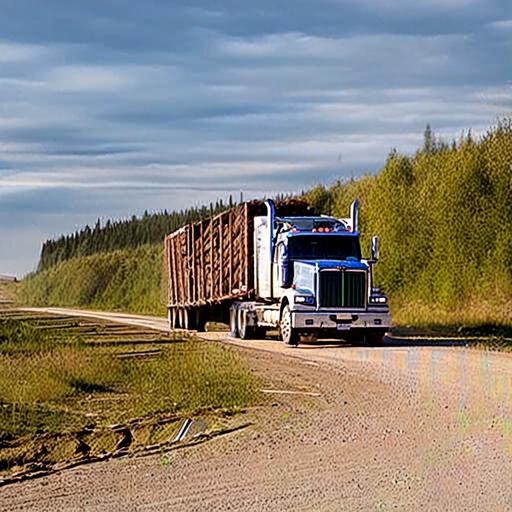}} &
        {\includegraphics[valign=c, width=\ww]{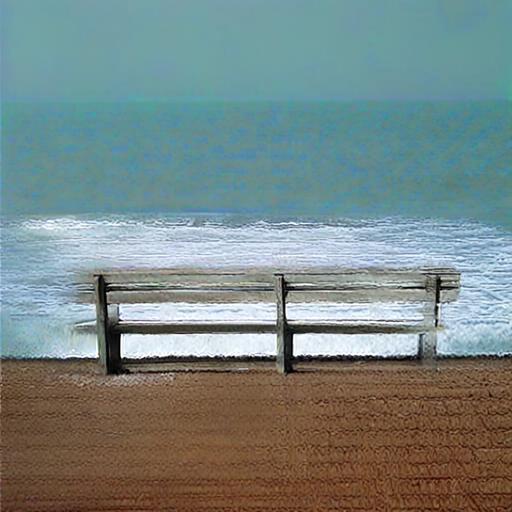}}
        \vspace{1px}
        \\

        \rotatebox[origin=c]{90}{{AnyDoor}}
        {\includegraphics[valign=c, width=\ww]{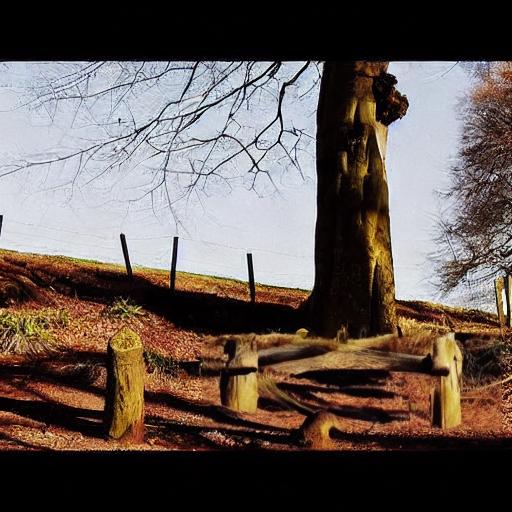}} &
        {\includegraphics[valign=c, width=\ww]{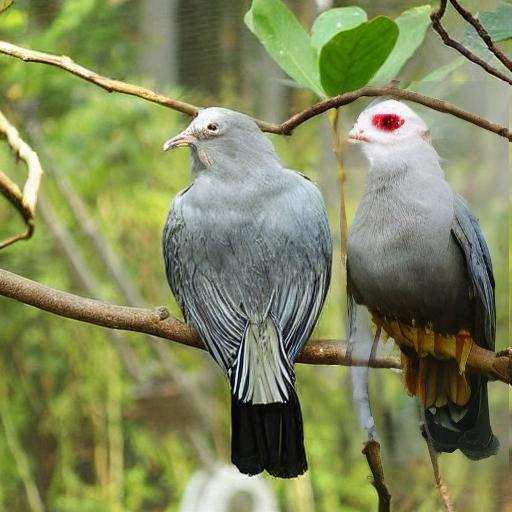}} &
        {\includegraphics[valign=c, width=\ww]{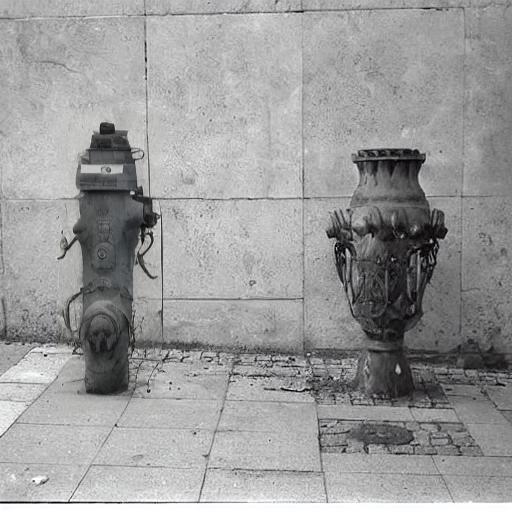}} &
        {\includegraphics[valign=c, width=\ww]{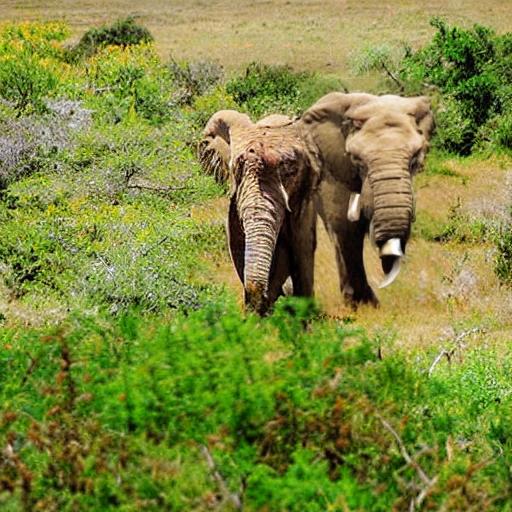}} &
        {\includegraphics[valign=c, width=\ww]{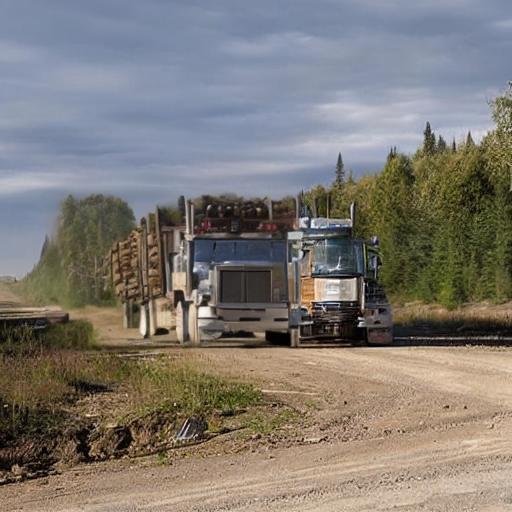}} &
        {\includegraphics[valign=c, width=\ww]{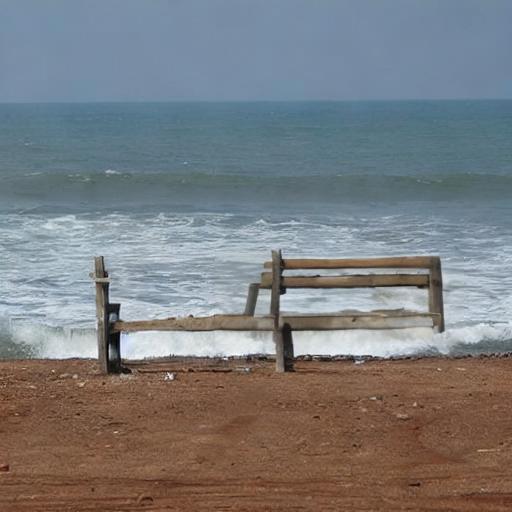}}
        \vspace{1px}
        \\

        \rotatebox[origin=c]{90}{{DragDiffusion}}
        {\includegraphics[valign=c, width=\ww]{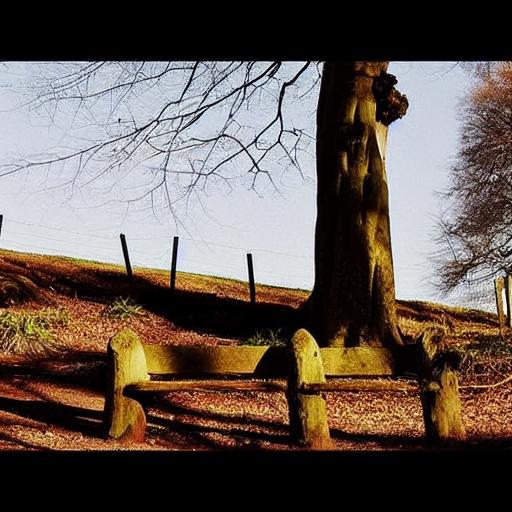}} &
        {\includegraphics[valign=c, width=\ww]{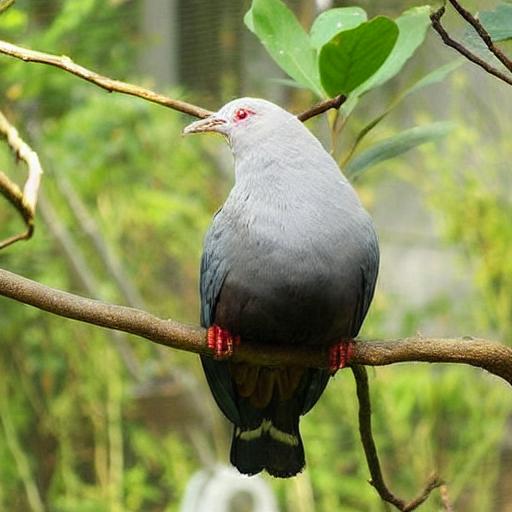}} &
        {\includegraphics[valign=c, width=\ww]{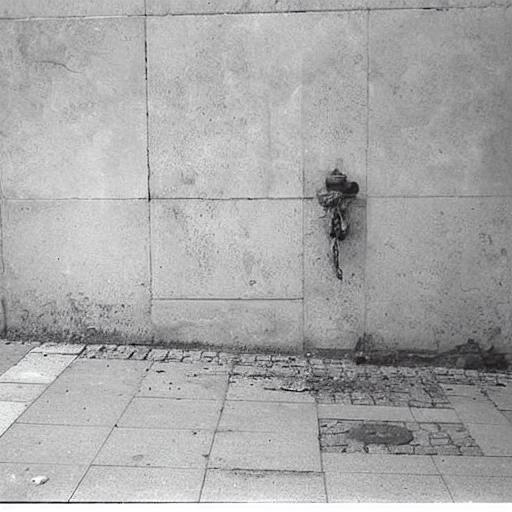}} &
        {\includegraphics[valign=c, width=\ww]{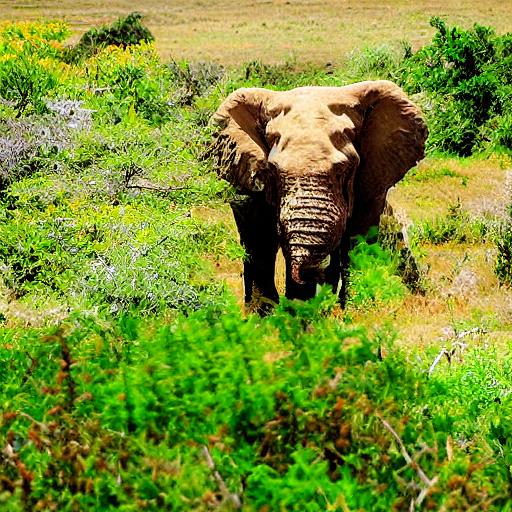}} &
        {\includegraphics[valign=c, width=\ww]{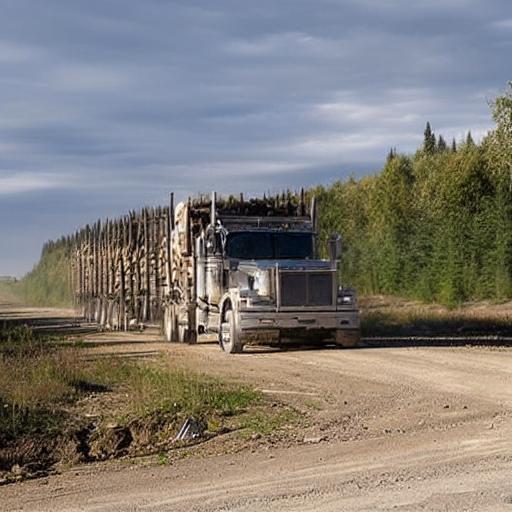}} &
        {\includegraphics[valign=c, width=\ww]{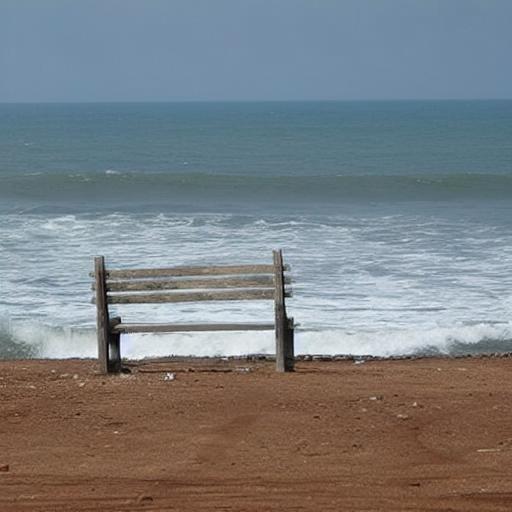}}
        \vspace{1px}
        \\

        \rotatebox[origin=c]{90}{{DragonDiffusion}}
        {\includegraphics[valign=c, width=\ww]{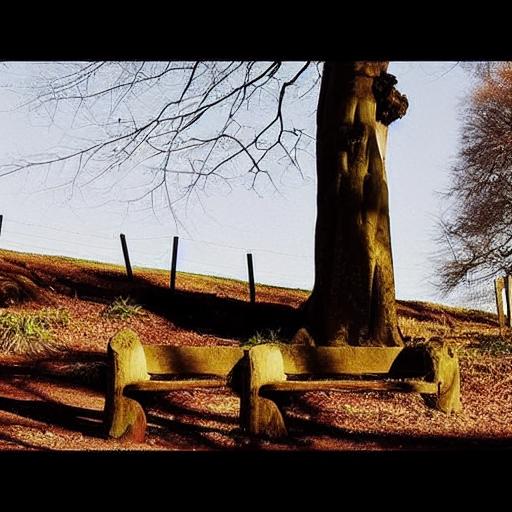}} &
        {\includegraphics[valign=c, width=\ww]{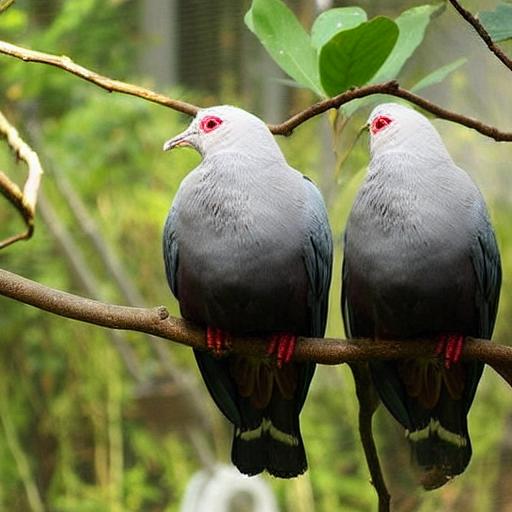}} &
        {\includegraphics[valign=c, width=\ww]{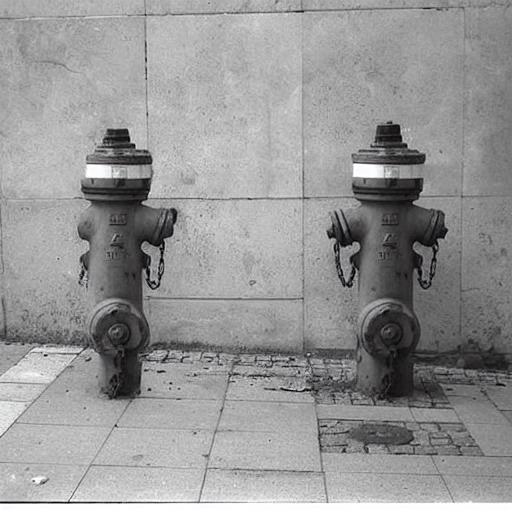}} &
        {\includegraphics[valign=c, width=\ww]{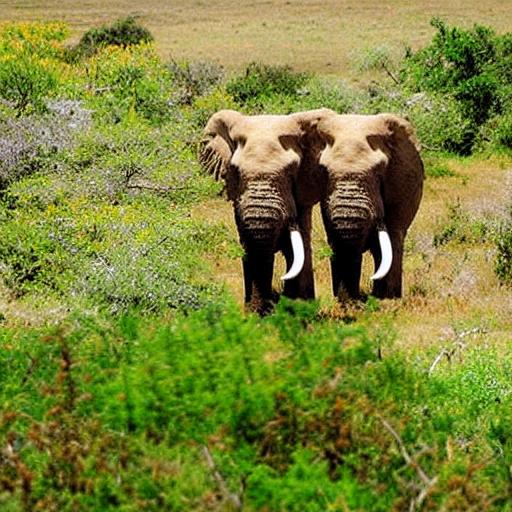}} &
        {\includegraphics[valign=c, width=\ww]{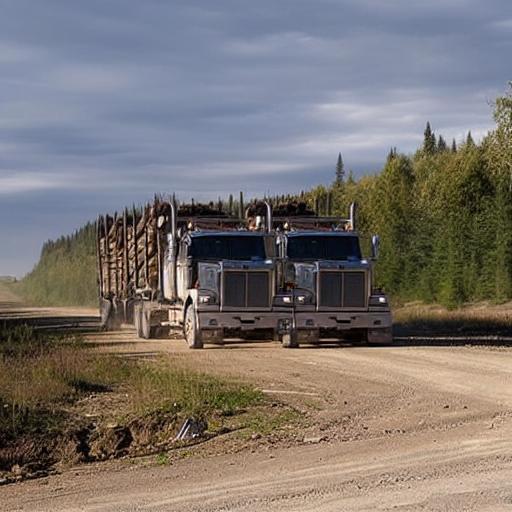}} &
        {\includegraphics[valign=c, width=\ww]{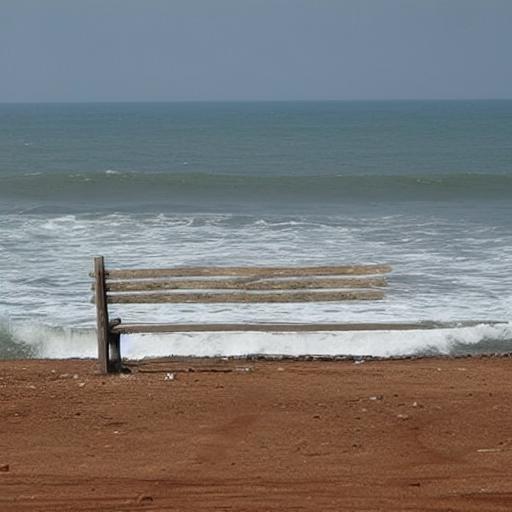}}
        \vspace{1px}
        \\

        \rotatebox[origin=c]{90}{{DiffEditor}}
        {\includegraphics[valign=c, width=\ww]{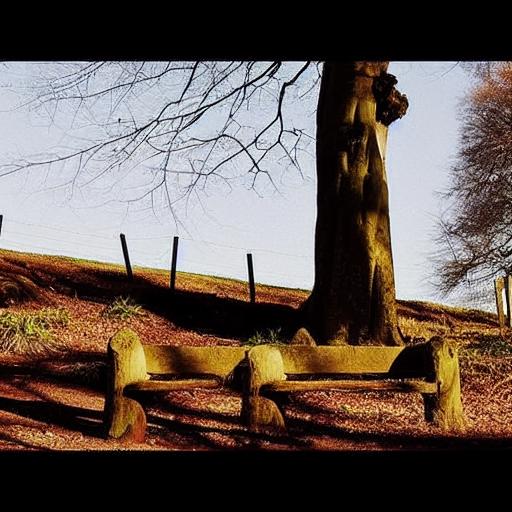}} &
        {\includegraphics[valign=c, width=\ww]{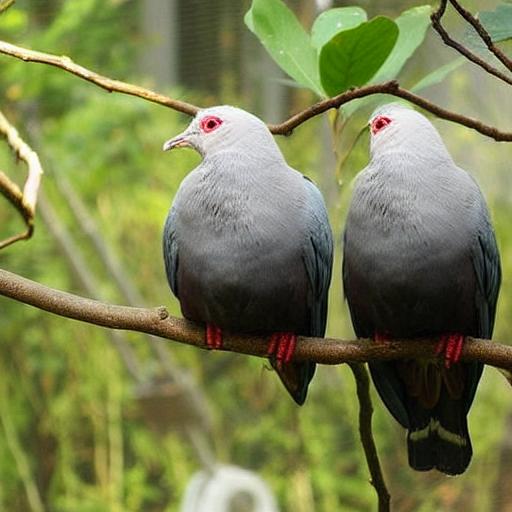}} &
        {\includegraphics[valign=c, width=\ww]{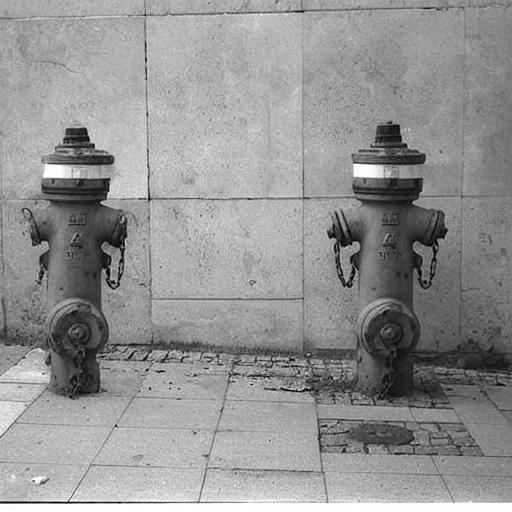}} &
        {\includegraphics[valign=c, width=\ww]{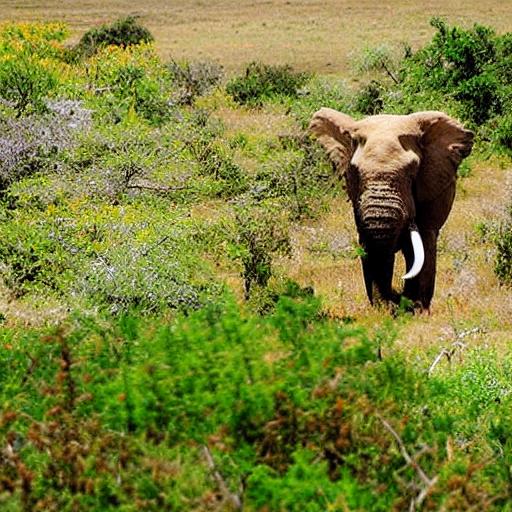}} &
        {\includegraphics[valign=c, width=\ww]{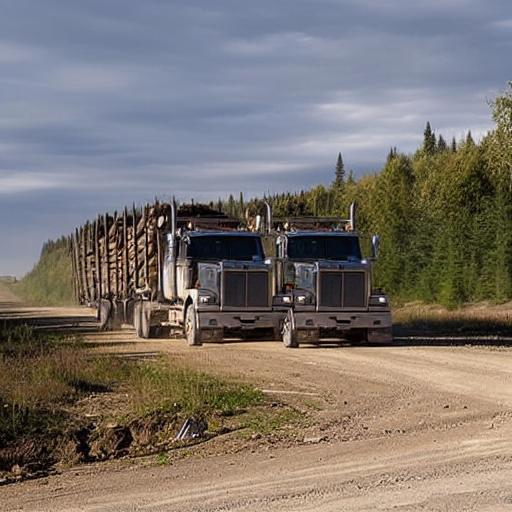}} &
        {\includegraphics[valign=c, width=\ww]{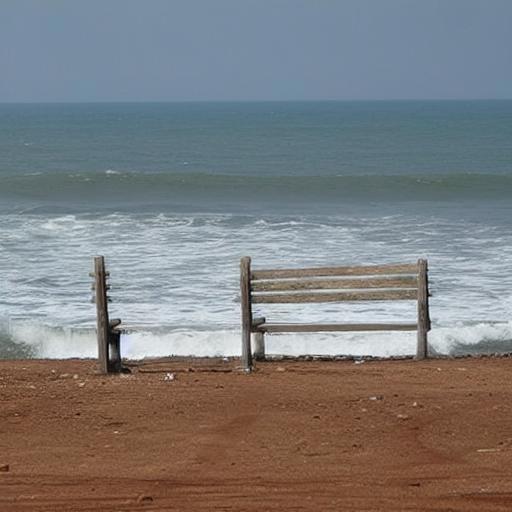}}
        \vspace{1px}
        \\

        \rotatebox[origin=c]{90}{{Ours}}
        {\includegraphics[valign=c, width=\ww]{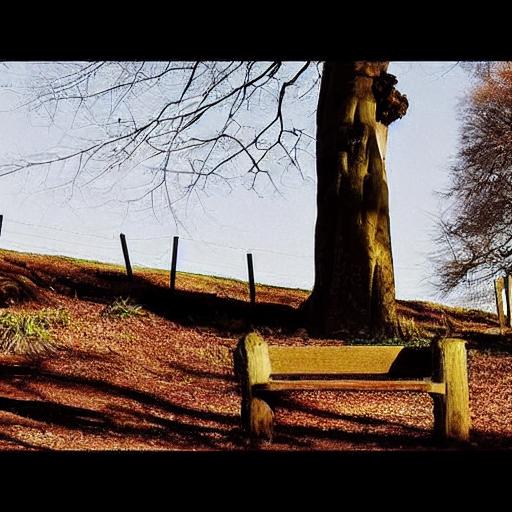}} &
        {\includegraphics[valign=c, width=\ww]{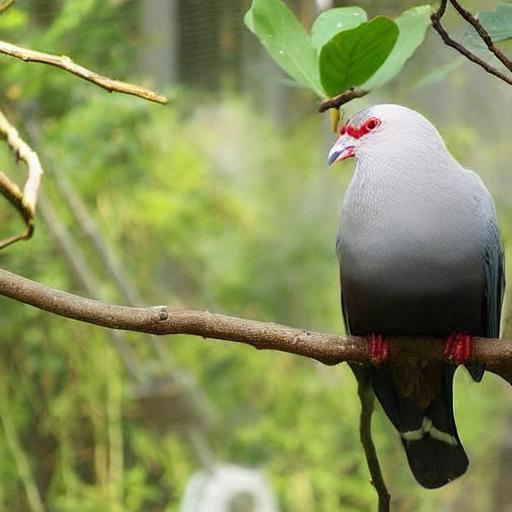}} &
        {\includegraphics[valign=c, width=\ww]{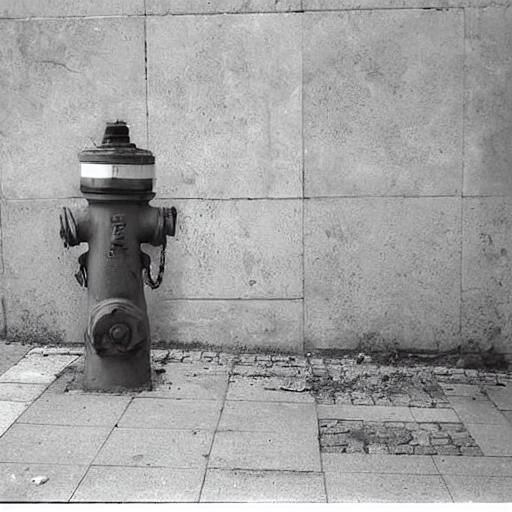}} &
        {\includegraphics[valign=c, width=\ww]{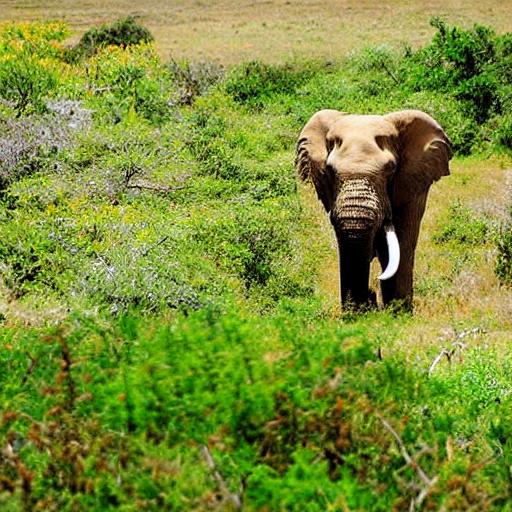}} &
        {\includegraphics[valign=c, width=\ww]{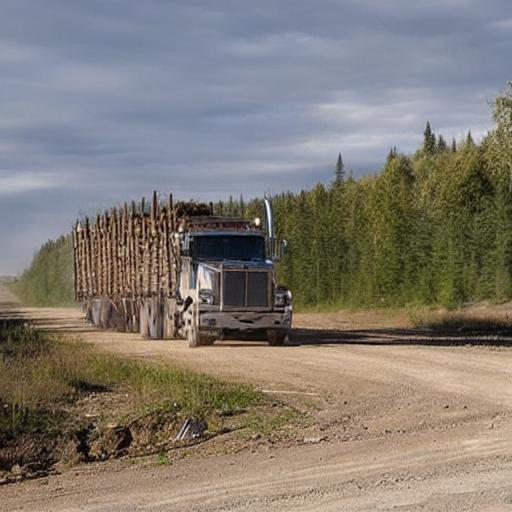}} &
        {\includegraphics[valign=c, width=\ww]{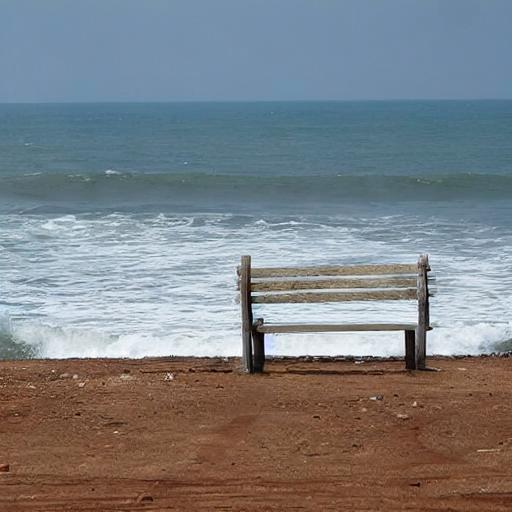}}
        \\

    \end{tabular}
    
    \caption{\textbf{Additional Qualitative Automatic Comparison.} As explained in 
    \Cref{sec:comparisons}, 
    we used a filtered version of COCO validation set \cite{Lin2014MicrosoftCC}. The source and target locations are denoted by red and green points, respectively. As can be seen, PBE \cite{Yang2022PaintBE}, DiffusionSG \cite{self_guidance} and Anydoor \cite{Chen2023AnyDoorZO} mainly suffer from a bad preservation of the foreground object. DragDiffusion \cite{Shi2023DragDiffusionHD} struggles with dragging the object, while DragonDiffusion \cite{Mou2023DragonDiffusionED} and DiffEditor \cite{Mou2024DiffEditorBA} suffers from object traces. Our method, on the other hand, strikes the balance between dragging the object and preserving its identity.}
    \label{fig:additional_qualitative_automatic_comparison}
\end{figure*}

\begin{figure*}[t]
    \centering
    \setlength{\tabcolsep}{5pt}
    \renewcommand{\arraystretch}{0.9}
    \setlength{\ww}{0.28\columnwidth}

    \begin{tabular}{cccccc}
        \rotatebox[origin=c]{90}{{Input}}
        {\includegraphics[valign=c, width=\ww]{figures/automatic_qualitative/assets/bear/inp_vis.jpg}} &
        {\includegraphics[valign=c, width=\ww]{figures/automatic_qualitative/assets/bench/inp_vis.jpg}} &
        {\includegraphics[valign=c, width=\ww]{figures/automatic_qualitative/assets/pipe/inp_vis.jpg}} &
        {\includegraphics[valign=c, width=\ww]{figures/automatic_qualitative/assets/zebra/inp_vis.jpg}} &
        {\includegraphics[valign=c, width=\ww]{figures/automatic_qualitative/assets/cow/inp_vis.jpg}} &
        {\includegraphics[valign=c, width=\ww]{figures/automatic_qualitative/assets/bird/inp_vis.jpg}}
        \vspace{1px}
        \\

        \midrule

        \rotatebox[origin=c]{90}{{w/o GSA masking}}
        {\includegraphics[valign=c, width=\ww]{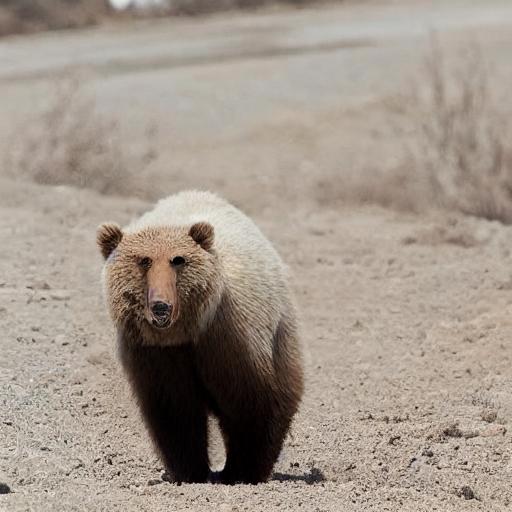}} &
        {\includegraphics[valign=c, width=\ww]{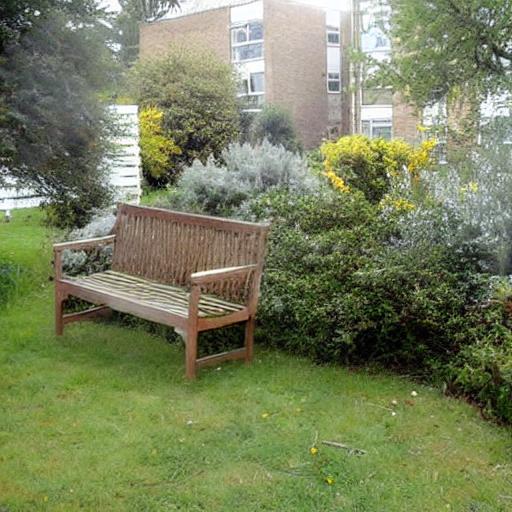}} &
        {\includegraphics[valign=c, width=\ww]{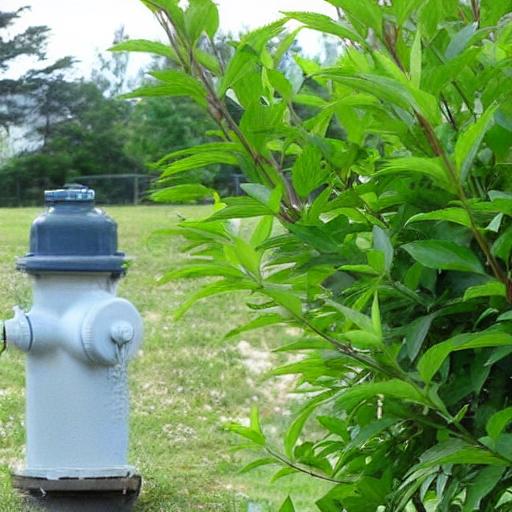}} &
        {\includegraphics[valign=c, width=\ww]{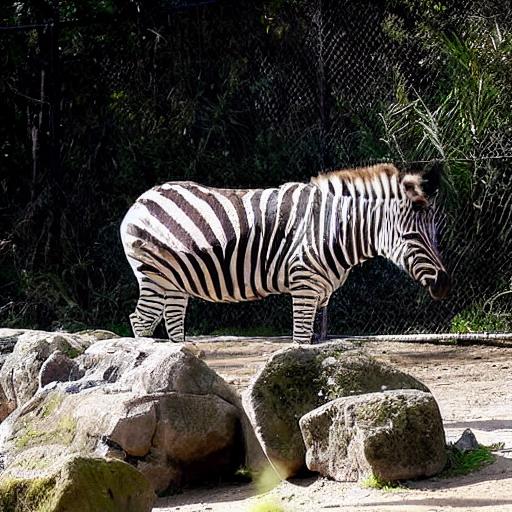}} &
        {\includegraphics[valign=c, width=\ww]{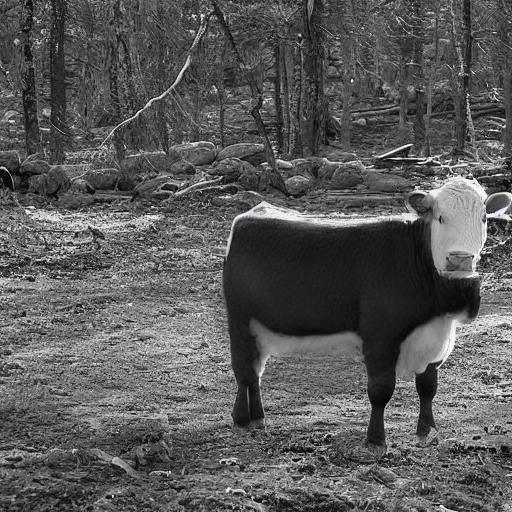}} &
        {\includegraphics[valign=c, width=\ww]{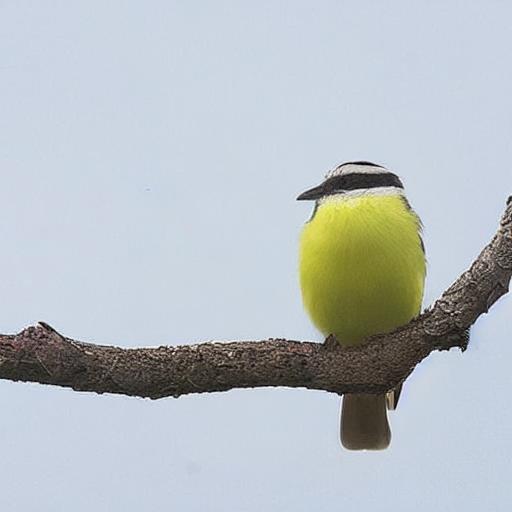}}
        \vspace{1px}
        \\

        \rotatebox[origin=c]{90}{{w/o SA sharing}}
        {\includegraphics[valign=c, width=\ww]{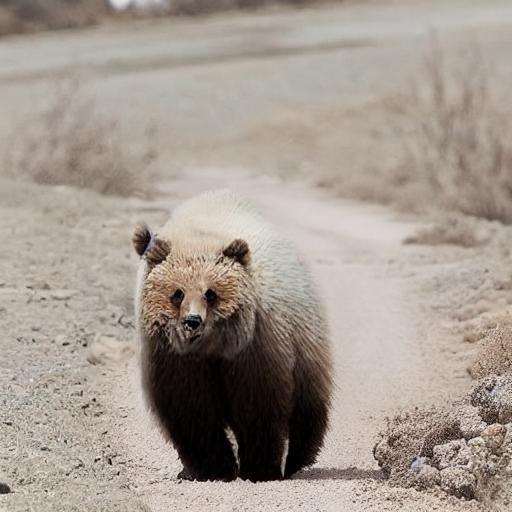}} &
        {\includegraphics[valign=c, width=\ww]{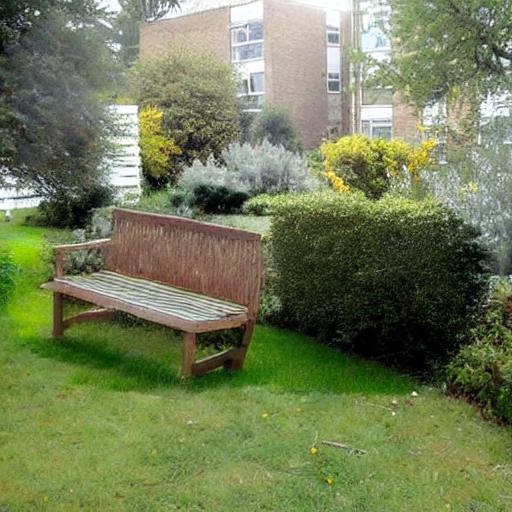}} &
        {\includegraphics[valign=c, width=\ww]{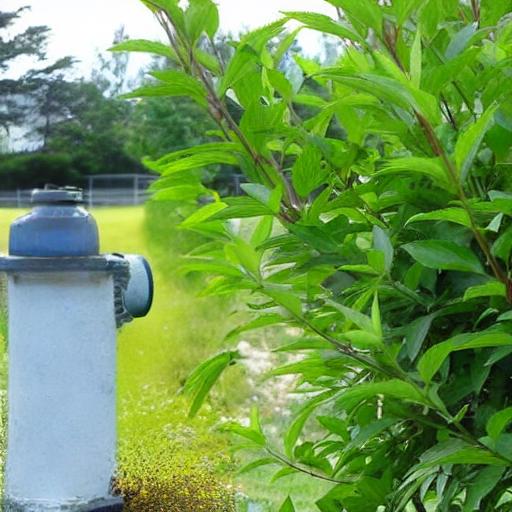}} &
        {\includegraphics[valign=c, width=\ww]{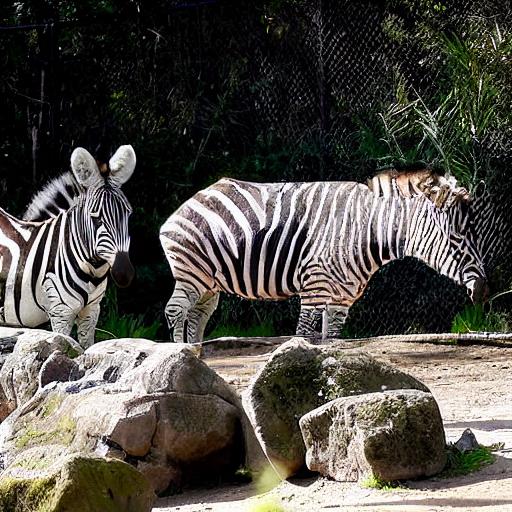}} &
        {\includegraphics[valign=c, width=\ww]{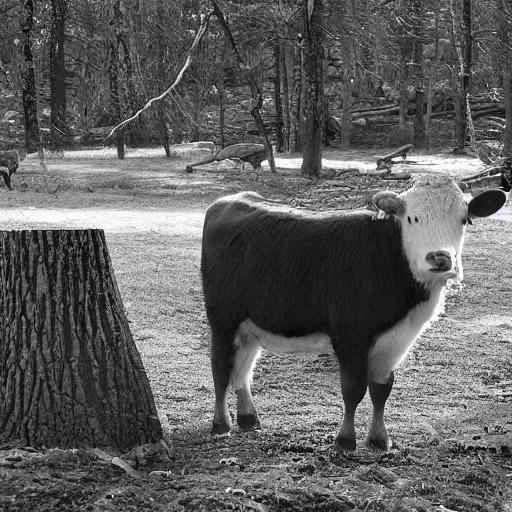}} &
        {\includegraphics[valign=c, width=\ww]{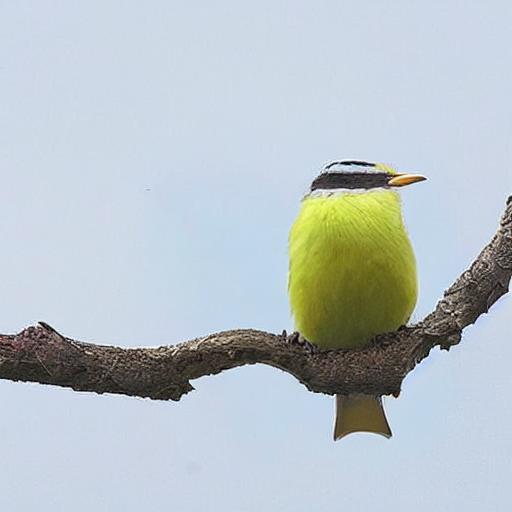}}
        \vspace{1px}
        \\

        \rotatebox[origin=c]{90}{{w/o anchoring}}
        {\includegraphics[valign=c, width=\ww]{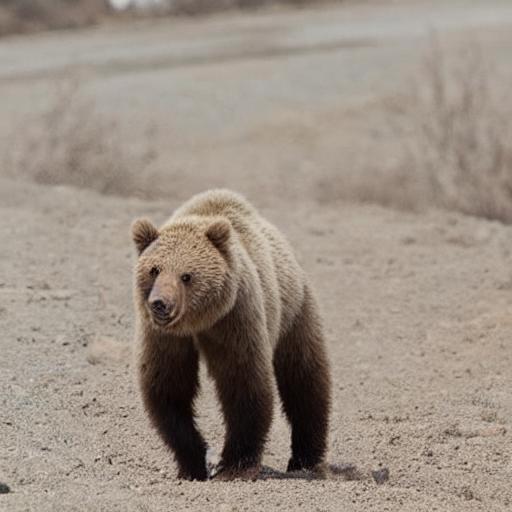}} &
        {\includegraphics[valign=c, width=\ww]{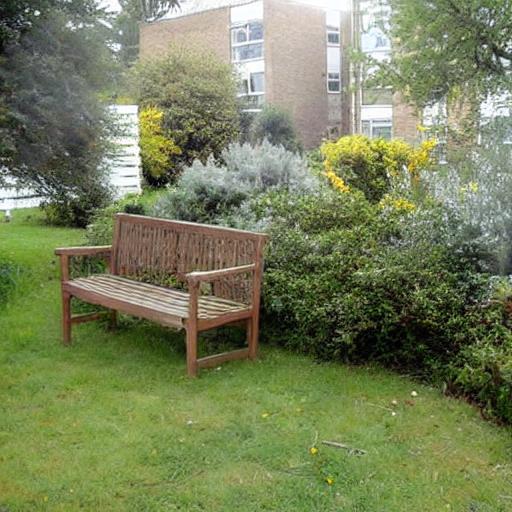}} &
        {\includegraphics[valign=c, width=\ww]{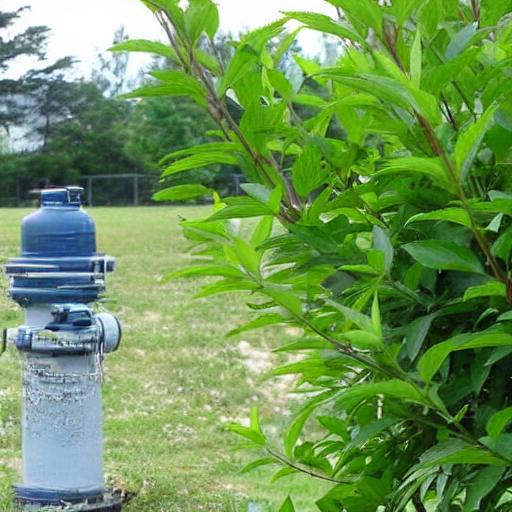}} &
        {\includegraphics[valign=c, width=\ww]{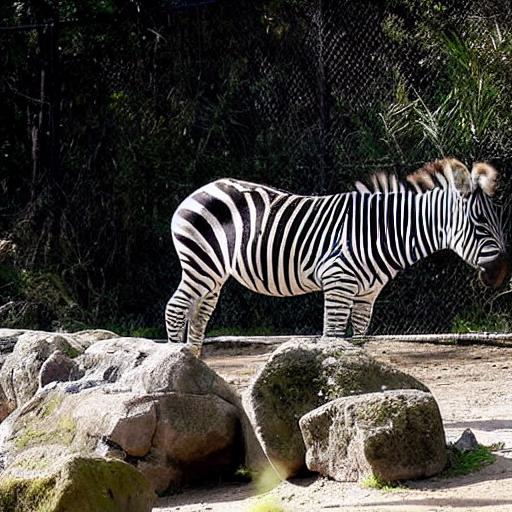}} &
        {\includegraphics[valign=c, width=\ww]{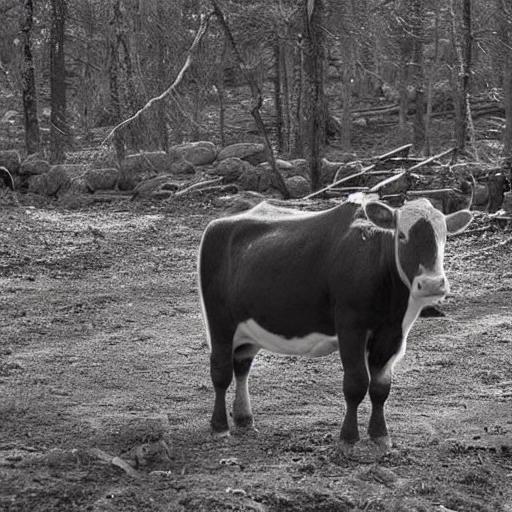}} &
        {\includegraphics[valign=c, width=\ww]{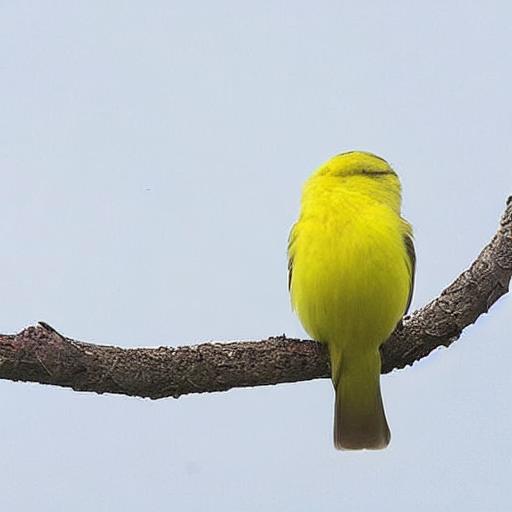}}
        \vspace{1px}
        \\

        \rotatebox[origin=c]{90}{{w/o DDPM}}
        {\includegraphics[valign=c, width=\ww]{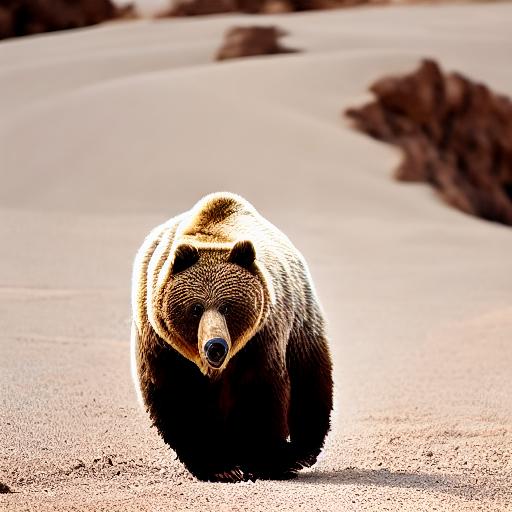}} &
        {\includegraphics[valign=c, width=\ww]{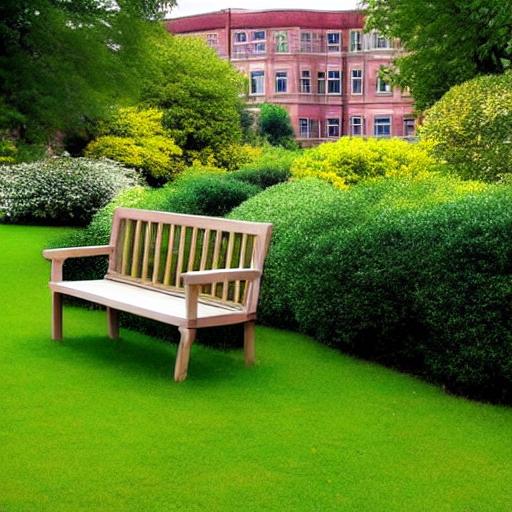}} &
        {\includegraphics[valign=c, width=\ww]{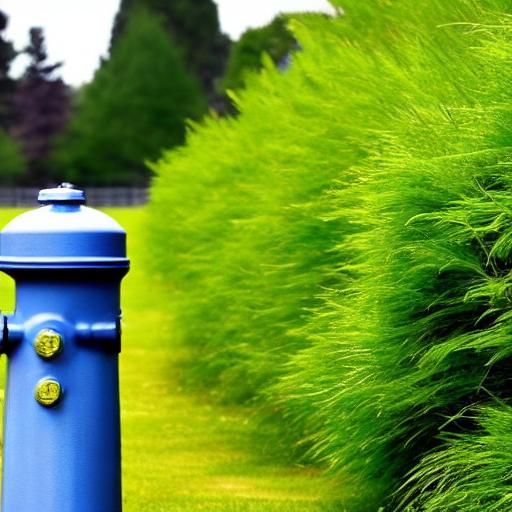}} &
        {\includegraphics[valign=c, width=\ww]{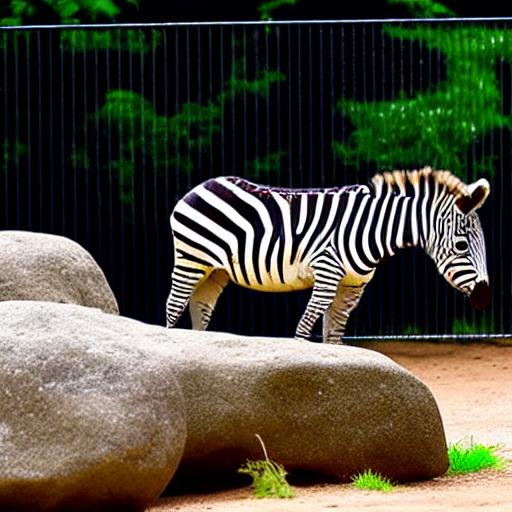}} &
        {\includegraphics[valign=c, width=\ww]{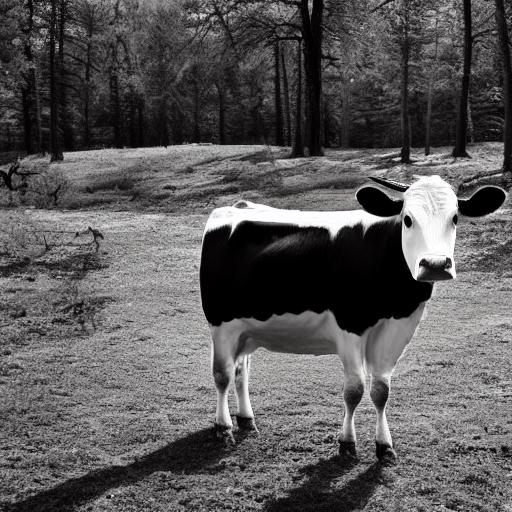}} &
        {\includegraphics[valign=c, width=\ww]{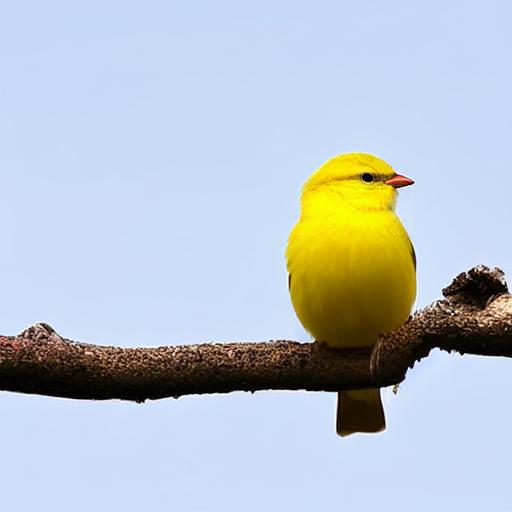}}
        \vspace{1px}
        \\

        \rotatebox[origin=c]{90}{{Ours}}
        {\includegraphics[valign=c, width=\ww]{figures/automatic_qualitative/assets/bear/ours.jpg}} &
        {\includegraphics[valign=c, width=\ww]{figures/automatic_qualitative/assets/bench/ours.jpg}} &
        {\includegraphics[valign=c, width=\ww]{figures/automatic_qualitative/assets/pipe/ours.jpg}} &
        {\includegraphics[valign=c, width=\ww]{figures/automatic_qualitative/assets/zebra/ours.jpg}} &
        {\includegraphics[valign=c, width=\ww]{figures/automatic_qualitative/assets/cow/ours.jpg}} &
        {\includegraphics[valign=c, width=\ww]{figures/automatic_qualitative/assets/bird/ours.jpg}}
        \\

    \end{tabular}
    
    \caption{\textbf{Qualitative Results in Ablation Study.} As explained in
    \Cref{sec:ablation_study},
    we ablate four key components of our method: (a) w/o GSA masking, (b) w/o SA sharing, (c) w/o soft attention anchoring and (d) w/o DDPM SA attention. As can be seen, all these components improve the foreground object consistency. For example, see the distorted face of the zebra or the distorted shape of the water pipe in the ablated cases.}
    \label{fig:qualitative_ablation_comparison}
\end{figure*}

In \Cref{fig:additional_qualitative_automatic_comparison} we provide an additional qualitative comparison of our method against the baselines on the automatically extracted dataset (as explained in \Cref{sec:autoamtic_metrics_implementation_details}). As can be seen, PBE \cite{Yang2022PaintBE}, DiffusionSG \cite{self_guidance} and Anydoor \cite{Chen2023AnyDoorZO} mainly suffer from a bad preservation of the foreground object. DragDiffusion \cite{Shi2023DragDiffusionHD} struggles with dragging the object, while DragonDiffusion \cite{Mou2023DragonDiffusionED} and DiffEditor \cite{Mou2024DiffEditorBA} suffers from object traces. Our method, on the other hand, strikes the balance between dragging the object and preserving its identity.

In addition, in \Cref{fig:qualitative_ablation_comparison} we provide a qualitative visualization of the ablation study we conducted. As can be seen, all these components improve the foreground object consistency. Please notice, the contribution of the GSA masking, as reflected by our automatic metric in 
\Cref{tab:ablation_study}
in the main paper, is limited in comparison to the other components that we added. However, the GSA masking mitigates the GSA leakage, which affects changes in the identity of the moved objects. This effect is visualized in \Cref{fig:qualitative_ablation_comparison}: see the distorted face of the zebra or the distorted shape of the water pipe in the second row, which ablates GSA masking.

\section{Societal Impact}
\label{sec:societal_impact}

We believe that the advent of technology enabling seamless object dragging within images holds tremendous promise for a wide array of creative and practical uses. It may democratize content manipulation for individuals lacking expertise and artistic skills. Furthermore, we believe that this method may present an invaluable tool for professional artists by expediting their creative processes without compromising quality.

Conversely, akin to other generative AI technologies, this method is susceptible to misuse, potentially giving rise to the creation of deceptive and misleading visual content. The ease and accessibility afforded by this technology could amplify concerns regarding the authenticity and trustworthiness of visual media in various contexts, including journalism, advertising, and social media which may erode the public trust in such content. Therefore, while recognizing its transformative potential, it is imperative to remain vigilant and implement appropriate safeguards to mitigate the proliferation of misinformation and uphold ethical standards in content creation and dissemination.

\end{document}